\documentclass{article} %

\usepackage[preprint]{neurips2026}

\usepackage{float} %
\usepackage[ruled,vlined,linesnumbered]{algorithm2e}

\usepackage{amsmath,amssymb}
\usepackage{xcolor}
\usepackage{colortbl}
\usepackage{tikz}
\usetikzlibrary{tikzmark}

\definecolor{llm}{RGB}{235,242,255}
\definecolor{REPLCOLOR}{RGB}{236,255,240}
\definecolor{statecolor}{RGB}{60,60,60}
\definecolor{codecolor}{RGB}{120,60,160}
\definecolor{subrlmcolor}{RGB}{0,140,140}

\SetAlgoVlined
\DontPrintSemicolon

\usepackage{microtype}
\usepackage{graphicx}
\usepackage{subcaption}
\usepackage{booktabs}

\usepackage{hyperref}

\usepackage[textsize=tiny]{todonotes}

\usepackage[T1]{fontenc}          %

\usepackage[most]{tcolorbox}      %

\definecolor{CustomRed}{HTML}{E30A17}
\definecolor{DarkCyan}{HTML}{0E9594}
\definecolor{OrangeWeb}{HTML}{FFA400}
\definecolor{Avocado}{HTML}{688E26}
\definecolor{TyrianPurple}{HTML}{550527}
\definecolor{BurntUmber}{HTML}{772E25}
\definecolor{FireEngineRed}{HTML}{C33626}
\definecolor{Carmine}{HTML}{990011}
\definecolor{CaribbeanCurrent}{HTML}{187077}
\definecolor{LightGreen}{HTML}{D2f4d3}
\definecolor{PigmentGreen}{HTML}{08A045}
\definecolor{VanillaPink}{HTML}{E57a81}
\definecolor{EarthYellow}{HTML}{E0A458}

\definecolor{PastelBlue}{HTML}{1E70EB}
\definecolor{PastelGreen}{HTML}{238636}
\definecolor{PastelPurple}{HTML}{8957E5}
\definecolor{PastelYellow}{HTML}{9f6B01}
\definecolor{PastelRed}{HTML}{DA3532}

\definecolor{PastelOrange}{HTML}{F66A0A}    %
\definecolor{PastelPink}{HTML}{EA4AAA}      %
\definecolor{PastelTeal}{HTML}{1B7C83}      %
\definecolor{PastelCyan}{HTML}{00B4D8}      %
\definecolor{PastelIndigo}{HTML}{5A32A3}    %
\definecolor{PastelBrown}{HTML}{7F4E1E}     %
\definecolor{PastelOlive}{HTML}{6C7E00}     %
\definecolor{PastelGray}{HTML}{586069}      %
\definecolor{PastelTurquoise}{HTML}{30CFCF} %
\definecolor{PastelCoral}{HTML}{FF6F61}     %

\definecolor{TEST1}{HTML}{5FCB8C}
\colorlet{PrimaryColor}{CaribbeanCurrent}
\colorlet{SecondaryColor}{EarthYellow}

\colorlet{LinkColor}{PrimaryColor}
\colorlet{llm}{PastelCyan!30}
\colorlet{REPLCOLOR}{PastelOrange!30}

\definecolor{statecolor}{RGB}{0,102,204}      %
\definecolor{reasoncolor}{RGB}{0,153,153}     %
\definecolor{codecolor}{RGB}{230,120,20}      %
\definecolor{turnscolor}{RGB}{128,0,128}      %
\definecolor{toolscolor}{RGB}{150,75,0}       %
\definecolor{subrlmcolor}{RGB}{200,0,200}     %

\usepackage{graphicx}
\usepackage{hyperref}
\usepackage{url}
\usepackage{amsmath}
\usepackage{amssymb}
\usepackage{mathtools}
\usepackage{amsfonts}  %
\usepackage{xcolor}  %
\usepackage{booktabs}
\usepackage{multirow}
\usepackage{array}
\usepackage{caption}
\usepackage{siunitx}
\usepackage{enumitem}
\usepackage{listings} 
\usepackage{xcolor}

\usepackage[capitalize,noabbrev]{cleveref}

\newcommand{\score}[3]{%
    #1 {\tiny\textcolor{gray}{(\$#2 $\pm$ \$#3)}}%
} 

\newcommand{\scoreNA}[1]{%
    #1 {\tiny\textcolor{gray}{(N/A) $\pm$ (N/A)}}%
}

\lstdefinestyle{customstyle}{
  backgroundcolor=\color{gray!10},   %
  basicstyle=\ttfamily\tiny, 
  breaklines=true,                    %
  frame=single,                       %
  xleftmargin=10pt,                   %
  xrightmargin=10pt,                  %
  aboveskip=10pt,                     %
  belowskip=10pt                      %
}

\newcommand{\RLM}{RLM}

\begin{document}

\title{Recursive Language Models}

\author{%
  Alex L. Zhang\thanks{Correspondence to Alex L. Zhang, Omar Khattab <\url{altzhang@mit.edu}, \url{okhattab@mit.edu}>.} \\
  MIT CSAIL \\
  \texttt{altzhang@mit.edu} \\
  \And
  Tim Kraska \\
  MIT CSAIL \\
  \texttt{kraska@mit.edu} \\
  \And
  Omar Khattab \\
  MIT CSAIL \\
  \texttt{okhattab@mit.edu} \\
}

\maketitle

\begin{abstract}
We study allowing large language models (LLMs) to process arbitrarily long prompts through the lens of inference-time scaling. We propose \textbf{Recursive Language Models} (\textbf{RLM}s), a general inference paradigm that treats long prompts as part of an external \textit{environment} and allows the LLM to \textit{programmatically} examine, decompose, and \textit{recursively call itself over} snippets of the prompt. 
We find that \RLM{}s can successfully process inputs more than an order of magnitude beyond model context window limits and, even for shorter prompts, dramatically outperform the quality of vanilla frontier LLMs and common long-context and coding scaffolds (e.g., on GPT-5 by a median across the evaluated benchmarks of $26\%$ against compaction, $130\%$ against CodeAct with sub-calls, and $13\%$ against Claude Code) across four diverse long-context tasks while having comparable cost.
At a small scale, we post-train the first model around the RLM. Our model, \textbf{RLM-Qwen3-8B}, outperforms the underlying Qwen3-8B model by a median of $28\%$ and even approaches the quality of vanilla GPT-5 on three long-context tasks. Code is available at \url{https://github.com/alexzhang13/rlm}.
\end{abstract}

\section{Introduction} \label{sec1:introduction}
\begin{figure}[hbt!]
    \centering
    \includegraphics[width=0.95\textwidth]{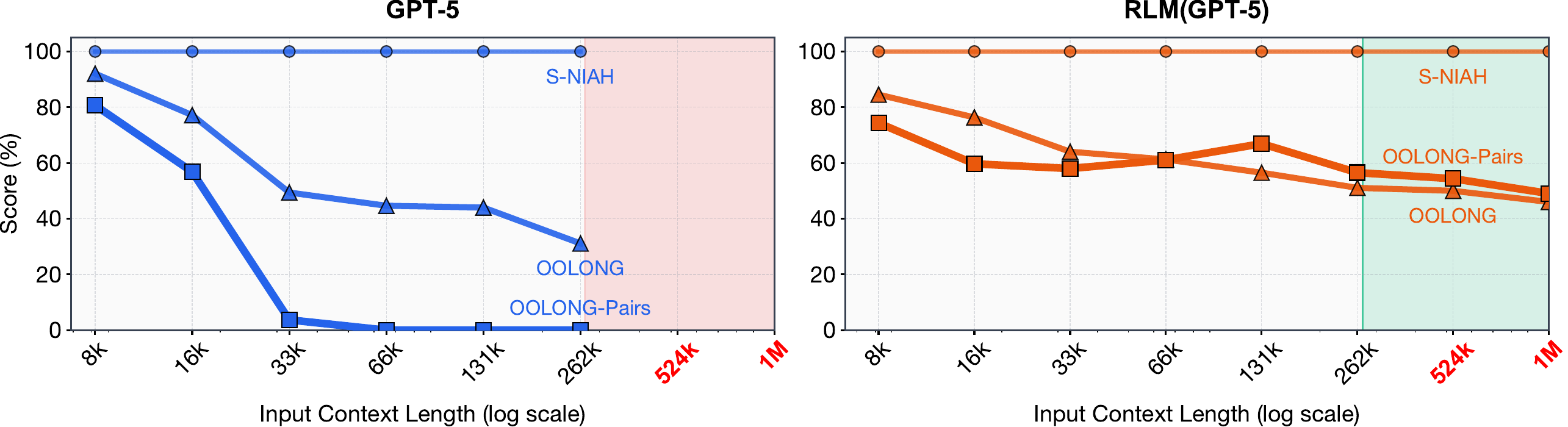}
    \caption{A comparison of GPT-5 and a corresponding \RLM{}(recursion depth=1) using GPT-5 on three long-context tasks of increasing complexity: \textbf{S-NIAH}, \textbf{OOLONG}, and \textbf{OOLONG-Pairs}. For each task, we scale the input length from $2^{13}$ to $2^{20}$. GPT-5 performance degrades significantly as a function of both input length and task complexity, while the \RLM{} maintains strong performance.
    Inputs beyond the red region do not fit in GPT-5's context window of 272K tokens, but the \RLM{} handles them effectively. Additional experiments across other models and benchmarks are in \S\ref{sec4:long-input}.
    }
    \vspace{-3mm}
    \label{fig:rlm-scaling}
\end{figure}

Frontier reasoning models have limited context windows and, even within their limits, tend to exhibit \textit{context rot}~\citep{hong2025contextrot}, a phenomenon illustrated in Figure~\ref{fig:rlm-scaling} where quality degrades steeply as prompts get longer. Though we expect context lengths to steadily rise through improvements to training, architecture, and infrastructure, we are interested in \textit{whether it is possible to scale the context size of general-purpose LLMs by orders of magnitude}. This is increasingly urgent as LLMs begin to be widely adopted for long-horizon tasks, in which they must routinely process tens if not hundreds of millions of tokens.

We study this question through the lens of scaling inference-time compute. We are inspired by the way that \textit{reasoning models}, another inference strategy, have become the fundamental interface to LLMs, resulting not only in empirical gains but also additional theoretical expressive power~\citep{merrillexpressive} compared to vanilla Transformers. %
Though most inference-time methods for dealing with long context are task-specific~\citep{wu2021recursivelysummarizingbookshuman,chang2024booookscore}, the most popular general approach is \textit{context condensation} or \textit{compaction} \citep{khattab2021baleen,smith2025openhands_context_condensensation,openai_codex_cli,wu2025resumunlockinglonghorizonsearch}, where context from user requests or agent trajectories is repeatedly summarized once it exceeds a length threshold. Unfortunately, compaction is rarely expressive enough for tasks that require dense access throughout the prompt. It presumes that \textit{some} details that appear early in the prompt can safely be forgotten to make room for new content.

\begin{figure*}[htb!]
    \centering
    \includegraphics[width=0.75\textwidth]{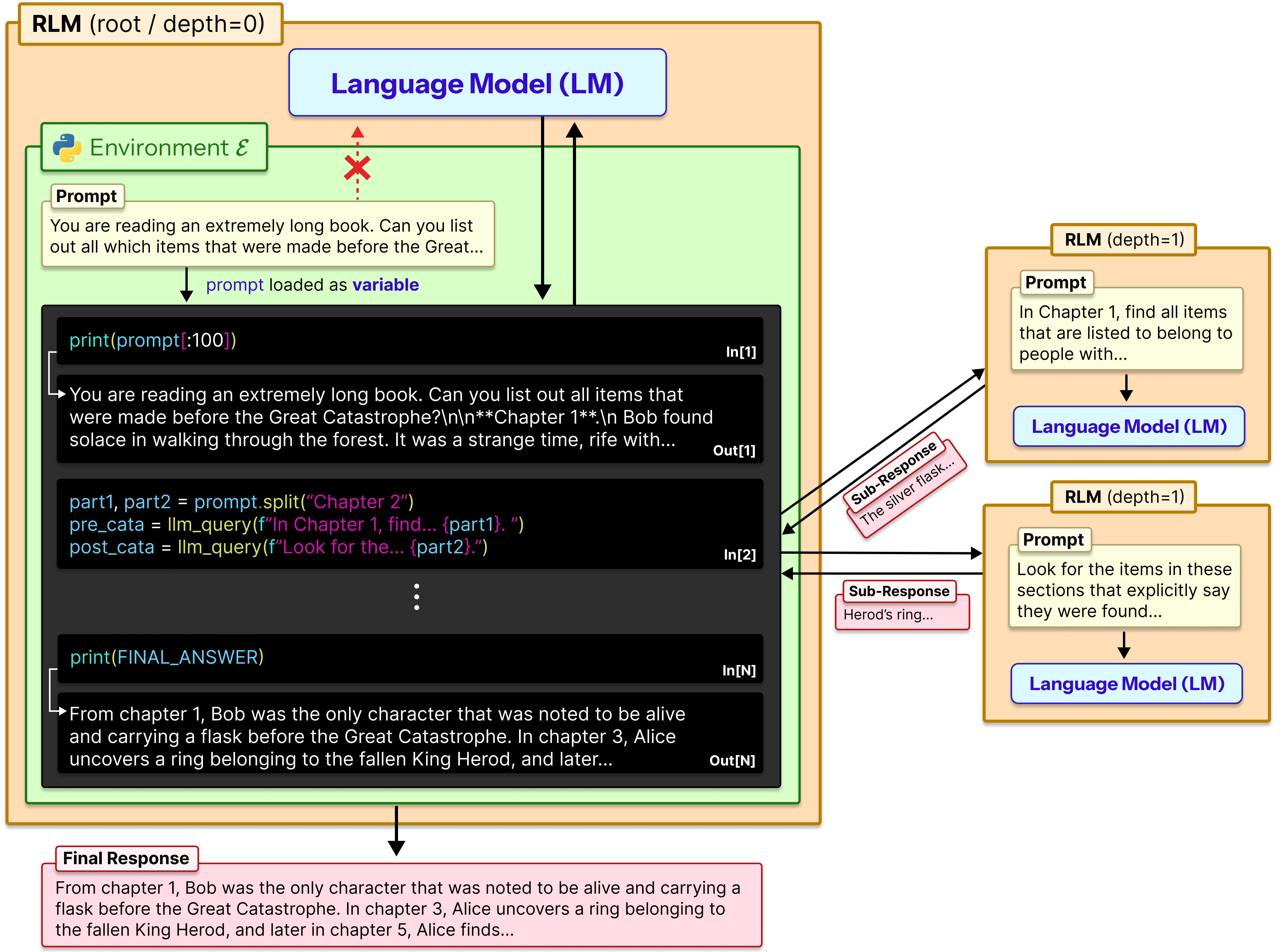}
    \caption{A Recursive Language Model (\RLM{}) treats prompts as part of the environment. It loads the input prompt as a variable inside a REPL environment $\mathcal{E}$ and writes code to peek into, decompose, and invoke itself recursively over programmatic snippets of the variable.}
    \label{fig:rlm-repl}
    \vspace{-1em}
\end{figure*}

We introduce \textbf{Recursive Language Models} (\textbf{\RLM{}}s), a general-purpose inference paradigm for dramatically scaling the effective input and output lengths of LLMs. The key insight is that arbitrarily long user prompts should not be fed into the neural network (e.g., Transformer) directly but should instead be treated as \textit{part of the environment that the LLM is tasked to \textbf{symbolically and recursively} interact with}. \textcolor{black}{This system serves as an abstracted ``language model'' without context limitations.}

As Figure~\ref{fig:rlm-repl} shows, an RLM exposes the same external interface as an LLM or a reasoning model: it accepts a string prompt of arbitrary structure and produces a string response. Given a prompt $P$, the RLM initializes a Read-Eval-Print Loop (REPL) programming environment in which $P$ is set as the value of a variable. It then offers the LLM general context about the REPL environment (e.g., the length of the string $P$), and permits it to write code that peeks into and decomposes $P$, and to iteratively observe any side effects from execution. Crucially, RLMs encourage the LLM to understand, transform, and execute the input prompt by \textit{writing symbolic programs that invoke the LLM itself} on as many slices of the input as necessary.

By treating the prompt itself as an external object and enabling symbolic recursion, RLMs tackle limitations of expressive power in recent work on coding agents, retrieval agents, and sub-agent delegation. In particular, prior coding agents and retrieval agents treat some designated external data source (e.g., a filesystem or a corpus of search documents) as an environment for fetching snippets. However, they can only fill up the underlying LLM's context window with snippets before facing compaction. Similarly, prior self-delegation approaches~\citep{anthropic_claude_code_subagents,sentient2025roma,schroeder2025threadthinkingdeeperrecursive, sun2025scalinglonghorizonllmagent} allow LLMs to invoke themselves as sub-agents. However, they are handicapped by the underlying LLM's limited output lengths because they are designed to verbalize sub-calls autoregressively rather than producing them programmatically. %

We evaluate \RLM{}s using a frontier closed model (GPT-5;~\citealt{singh2025openaigpt5card}) and a frontier open model (Qwen3-Coder-480B-A35B;~\citealt{Qwen3-Coder-480B-A35B}) across four tasks with varying levels of complexity: deep research~\citep{chen2025browsecompplusfairtransparentevaluation}, information aggregation~\citep{bertsch2025oolongevaluatinglongcontext}, code repository understanding~\citep{bai2025longbenchv2deeperunderstanding}, and a synthetic pairwise reasoning task where even frontier models fail catastrophically. We compare RLMs against direct LLM calls as well as context compaction, retrieval tool-use agents, and code-generation agents with and without sub-calls.

We find that RLMs demonstrate extremely strong performance even at the 10M+ token scale, and substantially outperform other approaches at long-context processing, in many cases by double-digit percentage gains while maintaining comparable cost. In particular, as demonstrated in Figure~\ref{fig:rlm-scaling}, \RLM{}s exhibit far less severe degradation for longer contexts and more sophisticated tasks. 

Finally, at a small scale, we post-train the first natively recursive language model, demonstrating that \RLM{}s can be improved quickly with little additional training. While a small open model (Qwen3-8B;~\citealt{yang2025qwen3technicalreport}) struggles to solve long context tasks even in an RLM scaffold, our simple general-purpose training recipe uses only 1,000 samples from unrelated domains to improve its performance by a median of $28.3\%$ across the four evaluation tasks.

\section{Recursive Language Models} \label{sec3:rlm}
Given a base neural language model $\mathcal{M}$ with maximum context size $K$, a Recursive Language Model (\RLM{}) is an inference-time scaffold around $\mathcal{M}$ that treats the user prompt as part of the environment without giving up the ability to densely process its content through different calls to $\mathcal{M}$.
Given an arbitrary-length prompt string $P\in\Sigma^\star$, an \RLM{} interacts with a persistent external environment $\mathcal{E}$ and returns a response string $Y\in\Sigma^\star$ (Figure~\ref{fig:rlm-repl}).
We would like effectively \emph{unbounded input tokens} ($|P|\gg K$), \emph{unbounded output tokens}, and an \emph{unbounded semantic horizon}, e.g. the ability to do $\Omega(|P|)$ or $\Omega(|P|^2)$ semantic work.

Algorithm~\ref{alg:rlm-call} describes how an RLM achieves this. Given a prompt $P$, the RLM initializes a persistent REPL programming environment with a variable containing the user prompt as a string and a function for invoking a sub-RLM with a new prompt. Then, it starts the RLM loop. In the first iteration, the algorithm invokes the \textit{root} neural model $\mathcal{M}$ with only (constant-size) metadata about the user prompt, like its length, a short prefix, and how to access parts of it.

The root is instructed via prompting (Appendix~\ref{appx3:methods}) and/or fine-tuning (Appendix~\ref{appx5:training}) to operate like an RLM: that is, to \textit{generate code that helps it understand and transform parts of its prompt $P$}, and to build up intermediate values and the final response into new variables, potentially by \textit{invoking the sub-RLM within loops}. In Section~\ref{sec4.3-results}, we find that existing LLMs can be prompted to do this and that training an 8B model to be natively recursive is promising.

Each iteration of the RLM loop executes code in the REPL, updates REPL state (intermediate variables), and collects in \texttt{stdout} any printed text. Only (constant-size) metadata about \texttt{stdout}, like a short prefix and length, is appended to $\mathcal{M}$'s history for the next iteration.\footnote{This is key: it forces $\mathcal{M}$ to rely on variables and sub-calls to manage long strings instead of polluting its window. In principle, if we trim each turn to $c$ tokens, we will have at most $K/c$ root iterations, each of which can launch arbitrarily many sub-calls. This is not a fundamental limitation, e.g. one could move the root horizon itself into a variable, but we typically want to limit the iterations at any level of recursion irrespective.} Once the RLM sets the variable \texttt{Final} inside the REPL, iteration stops and the value in \texttt{Final} is returned as the response.

RLMs make three simple design choices that are missing from many existing scaffolds. To highlight these, we include Algorithm~\ref{alg:bad-agent} to illustrate a deceptively ``similar'' algorithm that is far less expressive. Both algorithms support some notion of sub-calls, external objects, and code execution, but they differ in terms of where the prompt and intermediate values live and where recursion occurs.

\begin{table}[htbp]
    \centering
    \begin{minipage}[t]{0.48\textwidth}
        \begin{algorithm}[H]
\LinesNotNumbered
\footnotesize
\caption{A recursive language model, around LLM $\mathcal{M}$, which itself acts as a ``language model''.}
\label{alg:rlm-call}
\KwIn{prompt $P$}
\KwOut{response $Y$}

\texttt{state} $\gets$ \texttt{InitREPL(prompt=P)}\;

\texttt{state} $\gets$ \texttt{AddFunction(}\texttt{state},\, \textcolor{subrlmcolor}{\texttt{sub\_RLM}}\ensuremath{_{\mathcal{M}}}\texttt{)}\;

\texttt{hist} $\gets [\texttt{Metadata(state)}]$\;

\While{True}{
    \texttt{code} $\gets$ \colorbox{llm}{\texttt{LLM}\ensuremath{_{\mathcal{M}}}\texttt{(hist)}}\;

    (\texttt{state}, \texttt{stdout}) $\gets$ \colorbox{REPLCOLOR}{\texttt{REPL(state, code)}}\;

    \texttt{hist} $\gets$ \texttt{hist} $\,\Vert\,$ \texttt{code} $\,\Vert\,$ \texttt{Metadata(stdout)}\;

    \If{\texttt{state[Final]} is set}{
        \Return \texttt{state[Final]}\;
    }
}
\end{algorithm}

    \end{minipage}
    \hfill
    \begin{minipage}[t]{0.48\textwidth}
        \begin{algorithm}[H]
\LinesNotNumbered
\footnotesize
\caption{Alternate scaffold with standard (poor) design choices.} %
\label{alg:bad-agent}
\KwIn{prompt $P$}
\KwOut{response $Y$}
\texttt{actions} $\gets \{\texttt{Finish},\, \texttt{Exec},\, \texttt{Search},\, \textcolor{subrlmcolor}{\texttt{sub\_LLM}}\ensuremath{_{\mathcal{M}}}\}$\;
\texttt{hist} $\gets [\texttt{Metadata(actions)},\, P]$\hfill{\scriptsize\color{red}\textsf{Flaw \#1}}\;
\While{True}{
    (\texttt{action}, \texttt{val}) $\gets$ \colorbox{llm}{\texttt{LLM}\ensuremath{_{\mathcal{M}}}\texttt{(hist)}}\;
    \If{\texttt{action} is \texttt{Finish}}{
        \Return \texttt{val}\hfill{\scriptsize\color{red}\textsf{Flaw \#2}}\;
    }
    \texttt{out} $\gets$ \colorbox{REPLCOLOR}{\texttt{RUN(action, val)}}\hfill{\scriptsize\color{red}\textsf{Flaw \#3}}\;
    \texttt{hist} $\gets$ \texttt{hist} $\Vert$ (\texttt{action}, \texttt{val}, \texttt{out})\;
    \If{\texttt{Tok(hist)} > K}{
        \texttt{hist} $\gets$ \texttt{Compact(hist)}\;
    }
}
\end{algorithm}
\vspace{-1em}

    \end{minipage}
\end{table}

First, an \RLM{} must give the underlying LLM $\mathcal{M}$ a \emph{symbolic handle} to the user prompt $P$, so the model can manipulate it without copying text into the root context window. Instead, ineffective Algorithm~\ref{alg:bad-agent} starts by putting the user prompt $P$ into the LLM context window (\texttt{hist}), inheriting the window limitations of $\mathcal{M}$ and falling back to heuristics like context compaction. Even though the scaffold can access external data with, say, a \texttt{Search} action, it is bounded with respect to user input.

Second, ineffective Algorithm~\ref{alg:bad-agent} asks $\mathcal{M}$ to generate the output directly, via a \texttt{Finish} action. This may seem innocuous, but it means outputs cannot be longer than the context window of $\mathcal{M}$.

Third, and perhaps most importantly, an RLM requires \emph{symbolic recursion}. That is, code running \emph{inside} $\mathcal{E}$ must be able to invoke $\mathcal{M}$ on programmatically constructed transformations of $P$ (e.g., inside arbitrarily large loops), storing intermediate results symbolically. Though Algorithm~\ref{alg:bad-agent} includes both a code execution action and a ``sub-LLM'' action separately, it is not able to invoke the sub-LLM programmatically and hence can only delegate a few \textit{explicitly verbalized tasks} rather than writing short programs that can, say, loop over slices of the prompt and launch $\Omega(|P|)$ or even $\Omega(|P|^2)$ processes to understand or transform all parts of $P$.

\textcolor{black}{We implement our \RLM{} definition in Algorithm~\ref{alg:rlm-call} as follows: we equip an LLM with a Python REPL, where all tools, including sub-LM or sub-RLM calls, are available as modules. The initial prompt is stored as a variable in the REPL. The LLM interacts in a loop until it provides a final answer, which can be from either a variable in the REPL, or from the LLM itself. The LLM can also print from the REPL, but it is truncated to prevent overflowing the context too quickly.}

\section{Scaling Long Context Tasks} \label{sec4:long-input}
We hypothesize that the effective context window~\citep{hsieh2024rulerwhatsrealcontext, goldman2025reallylongcontextneed, hong2025contextrot} of an LLM cannot be understood independently of the \textit{specific task}. That is, more ``complex'' problems will exhibit degradation at even \textit{shorter} lengths than simpler ones. Because of this, we must characterize tasks in terms of how their complexity \textit{scales with prompt length}.

For example, needle-in-a-haystack (NIAH) problems generally keep `needles' constant as prompt length is scaled. As a result, frontier models can now reliably solve these tasks in RULER~\citep{hsieh2024rulerwhatsrealcontext} in the 1M+ token settings but struggle at far shorter lengths on OOLONG~\citep{bertsch2025oolongevaluatinglongcontext}, a task where the answer depends explicitly on almost every line in the prompt.\footnote{This helps explain the patterns seen in Figure~\ref{fig:rlm-scaling} earlier: \mbox{GPT-5} scales effectively on the S-NIAH task, where the needle size is constant despite longer prompts, but shows faster degradation at increasingly \textit{shorter} context lengths on the \textit{linear}-complexity OOLONG and the \textit{quadratic}-complexity OOLONG-Pairs.}

\subsection{Tasks}
\label{sec4.1-experimental-setup}
We design our evaluation around tasks where we can vary the lengths of the prompts, so we can consider problems whose difficulties scale differently with context length.

\textbf{S-NIAH}. Following the single needle-in-the-haystack task in RULER~\citep{hsieh2024rulerwhatsrealcontext}, we consider a set of 50 single tasks that require finding a specific phrase or number in a large set of unrelated text. Here, the information being sought scales as $O(1)$ with respect to input length.

\textbf{BrowseComp-Plus (1K documents)}~\citep{chen2025browsecompplusfairtransparentevaluation}. A multi-hop question-answering benchmark for DeepResearch~\citep{OpenAI_DeepResearch_2025} questions that requires reasoning over multiple different documents in an offline corpus. Following~\citet{sun2025scalinglonghorizonllmagent}, we use 150 randomly sampled instances as our evaluation set; we provide $1000$ randomly chosen documents as input, in which the gold and evidence documents are guaranteed to exist. We report the percentage of correct answers. The answer to each task requires piecing together information from several documents, making this harder than \textbf{S-NIAH} despite also requiring a constant number of documents. %

\textbf{OOLONG}~\citep{bertsch2025oolongevaluatinglongcontext}. A long reasoning benchmark that requires semantically labeling and aggregating these labels to form a final answer. We focus specifically on the \texttt{trec\_coarse} split, a set of $50$ tasks over a dataset of questions with semantic labels. Each task requires using nearly all dataset questions, and therefore scales linearly in processing complexity relative to the input length. %

\textbf{OOLONG-Pairs}. A modified variant of the \texttt{trec\_coarse} split of OOLONG with $20$ queries that specifically require aggregating \textit{pairs} of chunks to construct the final answer. We report F1 scores over the answer, which is a list of entries. Each task requires using nearly all \textit{pairs} of entries of the dataset, and therefore requires processing quadratically-many items relative to the input length. In Appendix~\ref{appx:oolong-pairs}, we list all queries in this benchmark.

\textbf{LongBench-v2 CodeQA}~\citep{bai2025longbenchv2deeperunderstanding}. A multi-choice code repository understanding split from LongBench-v2 that is challenging for modern frontier models. Each instance requires reasoning over a fixed number of files in a codebase to find the right answer.

\subsection{Methods and Baselines} \label{sec4.2-methods}
We compare \RLM{}s against commonly used task-agnostic inference methods, using two modern LMs, GPT-5 with medium reasoning~\citep{singh2025openaigpt5card} and default sampling parameters, and Qwen3-Coder-480B-A35B~\citep{yang2025qwen3technicalreport} using the sampling parameters described in~\citet{Qwen3-Coder-480B-A35B}. For Qwen3-Coder-480B-A35B, we compute costs based on the compute provider Fireworks~\citep{Qwen3-Coder-Fireworks}. In addition to evaluating the base model on all tasks, we also evaluate the following methods and baselines:

\textbf{CodeAct.} We compare directly to a  CodeAct~\citep{wang2024executablecodeactionselicit} agent that can execute code inside of a ReAct~\citep{yao2023reactsynergizingreasoningacting} loop. Unlike an \RLM, CodeAct does not offload the user prompt to the code environment, and instead provides it directly to the LM. We consider two variants: (1) a version following~\citet{jimenez2024swebenchlanguagemodelsresolve, chen2025browsecompplusfairtransparentevaluation} where we equip this agent with a BM25~\citep{10.1561/1500000019} retriever; (2) a version with a sub-call tool inside of the REPL. Compared to \RLM{}s, this method loads the context directly into the model.

\textbf{Compaction agent.} Following~\citet{sun2025scalinglonghorizonllmagent,wu2025resumunlockinglonghorizonsearch,yu2025memagentreshapinglongcontextllm}, we consider an iterative agent that compacts the context as it is filled. For example, given a corpus of documents, it will iteratively accumulate the documents and summarize when full. In cases where a single document exceeds the model window, the agent will chunk the document and iteratively compact it. For the GPT-5 experiments, due to the extremely high cost of applying this strategy to millions of tokens, we use GPT-5-nano for compaction and GPT-5 to provide the final answer.

\textbf{Coding agents.} \textcolor{black}{We compare against commonly used coding agents like OpenCode~\citep{anomalyco_opencode} and Claude Code~\citep{anthropic_claude_code_subagents}. We consider two variants, one where the context is offloaded to a file, and another where it is directly used as the initial prompt. Closed source agents like Claude Code are designed around a corresponding model, so we use Claude Opus 4.1 with Claude Code v2.0.0 (released around the same time as the GPT-5 model we use in our main results) for this baseline.}

\textbf{\RLM{}}. We implement an \RLM{} with a Python REPL environment, which loads a module for querying a sub-LM and uses a system prompt presented in Appendix~\ref{appx3:methods}. For the GPT-5 experiments, we use GPT-5-mini for the recursive LMs and GPT-5 for the root LM, as we found this choice to strike a good balance between the capabilities of RLMs and the cost of the recursive calls. We also evaluate several different max recursion depths allowable to the RLM, from 0-3. Max recursion depth 0 is an RLM without sub-calling capabilities. Max recursion depth 1 allows sub-calling LLMs, while max depth >1 allows sub-calling RLMs. We notate a \RLM{} with max recursion depth $N$ using a model as \RLM(model, depth=$N$), e.g. \RLM(GPT-5, depth=2), and assume depth=1 if not stated otherwise.

{\textbf{Fine-tuning.} To create \textbf{RLM-Qwen3-8B}, we fine-tune Qwen3-8B on 1,000 filtered trajectories of Qwen3-Coder-480B-A35B as an \RLM{} with Qwen3-8B sub-calls on LongBenchPro~\citep{chen2026longbenchprorealisticcomprehensive} tasks. We use sampling parameters described in~\citet{Qwen3-8B}, and evaluate the fine-tuned RLM-Qwen3-8B as an \RLM{}. The key insight for training is that being an effective sub-call model is roughly similar to being a general purpose reasoning model, so we can make the training much more tractable at small scale by focusing on improving the root model's ability to manipulate the REPL and to launch recursive calls. We provide more training details in Appendix~\ref{appx5:training}.}

\section{Results and Discussion} \label{sec4.3-results}
Table~\ref{tab:main} reports our main evaluation results. We additionally explore how vanilla frontier model and \RLM{} performance degrade as input contexts grow in Figure~\ref{fig:rlm-scaling}.

\begin{table*}[ht]
\centering
\caption{Performance comparison of different methods across long-context benchmarks of varying complexity. In \textcolor{gray}{gray} is the average API cost $\pm$ the standard deviation of each method on each task. $^{*}$ indicates runs where a method (sometimes) ran into input context limits. Provider costs were computed under OpenAI for GPT-5, under Fireworks for Qwen3 models, and under Anthropic for Claude Opus 4.1. All non-zero scores are rounded to at least $0.1$.}
\resizebox{\linewidth}{!}{%
\begin{tabular}{l@{\hskip 6pt}c@{\hskip 6pt}c@{\hskip 6pt}c@{\hskip 6pt}c}
\toprule
\textbf{Model} & \textbf{CodeQA} & \textbf{BrowseComp+ (1K)} & \textbf{OOLONG} & \textbf{OOLONG-Pairs} \\
\midrule
\textbf{Task Length $N$ (tokens)} & 23K-4.2M & 6M-11M & 131K & 32K \\
\midrule
\multicolumn{5}{l}{\textbf{GPT-5} {\tiny (with RLM sub-calls to GPT-5-mini)}} \\
\midrule
Base Model     
& \score{24.0$^{*}$}{0.13}{0.07}
& \scoreNA{0.0$^{*}$}
& \score{44.0}{0.14}{0.02}
& \score{0.1}{0.16}{0.10} \\

CodeAct (+ BM25)      
& \score{22.0$^{*}$}{0.06}{0.08}
& \score{51.0}{0.71}{1.20}
& \score{38.0}{0.61}{1.06}
& \score{24.7}{0.75}{0.43} \\

CodeAct (+ sub-calls)   
& \score{24.0$^{*}$}{0.06}{0.08}
& \scoreNA{0.0$^{*}$}
& \score{40.0}{0.85}{1.27}
& \score{28.4}{1.11}{0.62} \\

Compaction agent     
& \score{58.0}{1.31}{1.46}
& \score{70.5}{0.57}{0.10}
& \score{46.0}{0.13}{0.01}
& \score{0.1}{0.13}{0.09} \\

OpenCode    
& \scoreNA{18.0$^{*}$}
& \scoreNA{0.0$^{*}$}
& \scoreNA{32.0}
& \scoreNA{3.1} \\

OpenCode (+ context offloading)  
& \scoreNA{64.0}
& \scoreNA{\textbf{94.0}}
& \scoreNA{52.0}
& \scoreNA{4.8} \\

\rowcolor{gray!15}
RLM (recursion depth=0)   
& \score{58.0}{0.18}{0.56}
& \score{88.0}{0.44}{0.90}
& \score{36.0}{0.37}{0.42}
& \score{43.9}{0.69}{1.16} \\

\rowcolor{gray!15}
RLM (recursion depth=1)        
& \score{62.0}{0.11}{0.10}
& \score{91.3}{0.99}{1.22}
& \score{56.0}{0.43}{0.85}
& \score{58.0}{0.33}{0.20} \\

\rowcolor{gray!15}
RLM (recursion depth=2)        
& \score{\textbf{66.0}}{0.15}{0.30}
& \score{92.0}{0.55}{0.69}
& \score{56.5}{1.10}{3.25}
& \score{65.5}{0.33}{0.44} \\

\rowcolor{gray!15}
RLM (recursion depth=3)        
& \score{58.0}{0.15}{0.27}
& \score{92.0}{0.51}{0.54}
& \score{\textbf{58.0}}{0.51}{0.54}
& \score{\textbf{76.0}}{0.39}{0.32} \\

\midrule
\multicolumn{5}{l}{\textbf{Qwen3-Coder-480B-A35B}} \\
\midrule
Base Model     
& \score{20.0$^{*}$}{0.13}{0.08}
& \scoreNA{0.0$^{*}$}
& \score{36.0}{0.06}{0.00}
& \score{0.1}{0.05}{0.01} \\

CodeAct (+ BM25)      
& \score{24.0$^{*}$}{0.17}{0.08}
& \score{12.7}{0.39}{0.50}
& \score{38.0}{1.51}{1.09}
& \score{0.3}{1.54}{0.35} \\

CodeAct (+ sub-calls)
& \score{26.0$^{*}$}{0.28}{0.30}
& \scoreNA{0.0$^{*}$}
& \score{32.0}{1.83}{1.14}
& \score{0.1}{1.49}{0.46} \\

Compaction agent     
& \score{50.0}{1.26}{1.50}
& \score{38.0}{8.98}{2.12}
& \score{44.1}{0.15}{0.01}
& \score{0.31}{0.05}{0.00} \\

OpenCode    
& \scoreNA{12.0$^{*}$}
& \scoreNA{0.0$^{*}$}
& \scoreNA{36.0}
& \scoreNA{0.0} \\

OpenCode (+ context offloading)  
& \scoreNA{40.0}
& \scoreNA{58.0}
& \scoreNA{24.0}
& \scoreNA{2.1} \\

\rowcolor{gray!15}
RLM (recursion depth=0)      
& \score{\textbf{66.0}}{0.18}{0.58}
& \score{46.0}{0.82}{0.69}
& \score{43.5}{0.32}{0.13}
& \score{17.3}{1.77}{1.23} \\

\rowcolor{gray!15}
RLM (recursion depth=1)        
& \score{56.0}{0.92}{1.23}
& \score{44.7}{0.84}{0.63}
& \score{\textbf{48.0}}{0.61}{0.49}
& \score{\textbf{23.1}}{1.02}{0.52} \\

\rowcolor{gray!15}
RLM (recursion depth=2)        
& \score{54.0}{1.88}{3.30}
& \score{68.0}{1.05}{0.67}
& \score{26.0}{1.03}{1.65}
& \score{19.0}{1.61}{0.99} \\

\rowcolor{gray!15}
RLM (recursion depth=3)        
& \score{44.0}{1.65}{1.63}
& \score{\textbf{68.7}}{1.10}{0.80}
& \score{32.0}{0.80}{1.03}
& \score{21.1}{1.67}{1.21} \\

\midrule
\multicolumn{5}{l}{\textbf{Claude Opus 4.1}} \\
\midrule
Claude Code
& \score{12.0$^{*}$}{2.03}{0.57}
& \scoreNA{0.0$^{*}$}
& \score{40.2}{3.43}{1.60}
& \score{0.1}{6.75}{3.57} \\
Claude Code (+ context offloading)
& \score{62.0}{1.25}{0.54}
& \score{84.0}{2.03}{1.49}
& \score{48.0}{0.98}{0.55}
& \score{6.5}{2.99}{1.16} \\

\bottomrule
\end{tabular}}
\label{tab:main}
\end{table*}

\textbf{Observation 1: \RLM{}s can scale to the 10M+ token regime and can outperform base LMs and existing task-agnostic agent scaffolds on long context tasks}. Across all tasks, \RLM{}s demonstrate strong performance on prompts well beyond the effective context window of a frontier LM, outperforming base models and common long-context scaffolds by up to $2\times$ the performance while maintaining comparable or cheaper average token costs. Notably, \RLM{}s scale well beyond the base models' context window. For instance, on BrowseComp-Plus (1K), a linearly extrapolated cost for GPT-5-mini ingesting 6-11M input tokens is $\$1.50 - \$2.75$, while \RLM(GPT-5, depth=1) has an average cost of $\$0.99$ and outperforms both the compaction and retrieval baselines by over $29\%$. 

Furthermore, on tasks where processing costs scale with the input context, \RLM{}s make significant improvements over the base model, even on tasks within the model's context window. On OOLONG, the \RLM{}(depth=1) with GPT-5 and Qwen3-Coder outperform the base model by $28.4\%$ and $33.3\%$ respectively. On OOLONG-Pairs, both GPT-5 and Qwen3-Coder make little progress with F1 scores of $\leq0.1\%$, while the \RLM{}(depth=1) using these models achieve F1 scores of $58.0\%$ and $23.1\%$ respectively, highlighting the capability of \RLM{}s to handle extremely information-dense tasks.

\textbf{Observation 2: The REPL is necessary for handling long inputs, while the recursive sub-calling of \RLM{}s provides strong benefits on information-dense inputs.} A key characteristic of \RLM{}s is offloading the context as a variable in an environment $\mathcal{E}$ that the model can interact with. In particular, \RLM{}(depth=0) and coding agents like Claude Code and OpenCode are able to scale beyond the context limit of the model and outperform other task-agnostic baselines on most long context settings. On CodeQA in particular with Qwen3-Coder-480B-A35B, the no-sub-calling \RLM(depth=0) is able to outperform all sub-calling variants of the \RLM.

On information-dense tasks like OOLONG or OOLONG-Pairs, we observed several cases where programmatic recursive LM sub-calling is necessary. In \S\ref{sec4.4-qualitative}, we see \RLM{}(Qwen3-Coder) perform the necessary semantic transformation line-by-line through recursive sub-calls, while the ablation without sub-calls is forced to use keyword heuristics to solve these tasks. On OOLONG-Pairs in particular, the higher recursive depth variants of the \RLM{} for GPT-5 outperform all other methods including Claude Code and OpenCode by a large margin.

\textbf{Observation 3: LM performance degrades as a function of input length and problem complexity, while \RLM{} performance scales better.} The benchmarks {S-NIAH}, {OOLONG}, and {OOLONG-Pairs} contain a fixed number of tasks over contexts with lengths ranging from $2^{13}$ to $2^{20}$. Each benchmark can be categorized by different processing complexity of the input context with respect to length (roughly constant, linear, and quadratic respectively). In Figure~\ref{fig:rlm-scaling}, we directly compare an \RLM{}(GPT-5, depth=1) to base GPT-5, and find that GPT-5 performance degrades significantly faster for more complex tasks, which aligns with the findings of \citet{goldman2025reallylongcontextneed}, while \RLM{} performance degrades at a slower rate. For context lengths beyond $2^{14}$, the \RLM{} consistently outperforms GPT-5.

Furthermore, \RLM{} costs scale proportionally to the complexity of the task, while still remaining in the same order of magnitude of cost as GPT-5 (see Figure~\ref{fig:cost-scaling} in Appendix~\ref{appx1:runtime-cost}). In \S\ref{sec4.4-qualitative}, we explore the choices that the \RLM{} makes that cause these differences in cost. %

\textbf{Observation 4: The inference cost of \RLM{}s remains comparable to other methods, and in some cases base LM calls.} On average, we find in Table~\ref{tab:main} that the inference cost of \RLM{}s is cheaper or comparable to most other baselines, including standard coding agents. Furthermore, in Figure~\ref{fig:quartiles} in Appendix~\ref{appx1:runtime-cost}, we find that the median \RLM{} run is cheaper than the median base model run, but more expensive on average due to outlier trajectories where the \RLM{} struggles to find an answer.

We additionally report runtime numbers of each method in Figures~\ref{fig:runtime-gpt-5},~\ref{fig:runtime-qwen3} in Appendix~\ref{appx1:runtime-cost}, but we note several important caveats. Unlike API costs, these numbers are heavily dependent on implementation details such as the machine used, API request latency, and the asynchrony of LM calls. In our implementation of the baselines and \RLM{}s, all LM calls are blocking / sequential. Nevertheless, similar to costs, we observe a wide range of runtimes, especially for \RLM{}s.

\begin{table*}[ht]
\centering
\caption{Solve rate on \textsc{LongCoT-mini}~\citep{motwani2026longcotbenchmarkinglonghorizonchainofthought}, a difficult long reasoning benchmark that frontier models struggle to solve. We select the best performing model from the paper (GPT-5.2) and compare to an \RLM{} with and without decomposition hints (prompt provided in Appendix~\ref{appx:longcot}).}
\resizebox{\linewidth}{!}{%
\begin{tabular}{l@{\hskip 6pt}c@{\hskip 6pt}c@{\hskip 6pt}c@{\hskip 6pt}c@{\hskip 6pt}c@{\hskip 6pt}c}
\toprule
\textbf{Model} & \textbf{Overall} & \textbf{MATH} & \textbf{CHEM} & \textbf{CS} & \textbf{LOGIC} & \textbf{CHESS} \\
\midrule
GPT-5.2 (base)
& 38.7
& 26.0
& 37.0
& 40.4
& 53.6
& 36.6 \\

RLM (GPT-5.2, recursion depth=1)
& 50.6
& 5.6
& 50.0
& 11.0
& 86.7
& 93.0 \\

RLM (GPT-5.2, recursion depth=1) + decomposition hints
& \textbf{65.6}
& \textbf{32.0}
& \textbf{52.0}
& \textbf{46.0}
& \textbf{99.0}
& \textbf{99.0} \\

\bottomrule
\end{tabular}}
\label{tab:longcot}
\end{table*}

\textbf{Observation 5: Beyond long-context, \RLM{}s enable longer reasoning capabilities.} In Table~\ref{tab:longcot}, we report \RLM{} performance on LongCoT-mini~\citep{motwani2026longcotbenchmarkinglonghorizonchainofthought}, a challenging long reasoning benchmark where frontier models solve compositional problems containing interdependent subproblems. We compare with the best model reported in the paper, GPT-5.2, and find that \RLM{}(GPT-5.2, depth=1) uses the REPL to outperform the base model. Furthermore, when providing explicit hints on how to decompose tasks, we find the \RLM{} is able to reliably generate a graph of the problem, solving each node using sub-calls as it programmatically traverses the reasoning graph. It outperforms the base model on all domains and by a $69.5\%$ performance increase overall.

\begin{figure}[!hbt]
    \centering
    \includegraphics[width=1\linewidth]{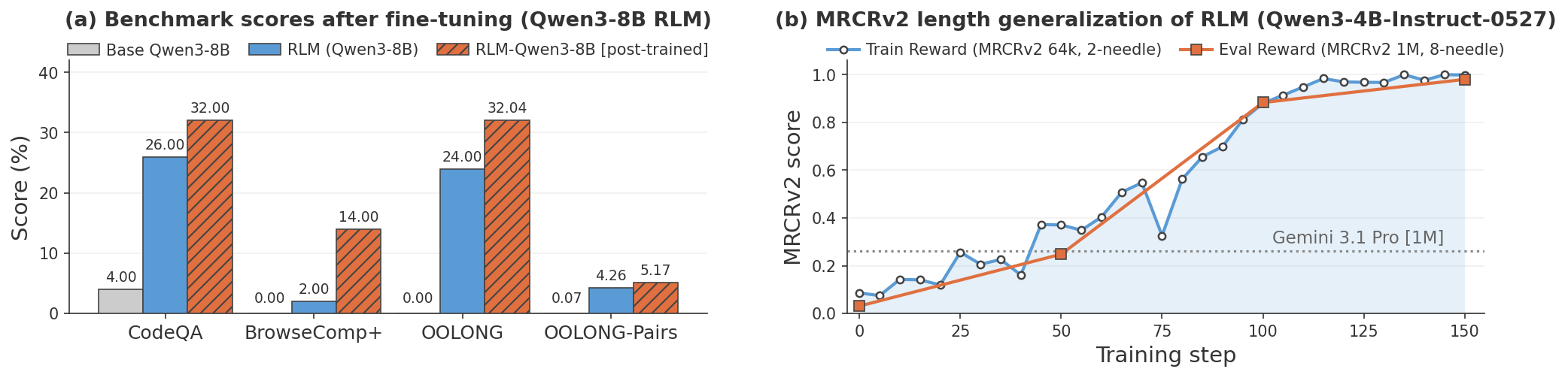}
    \caption{(a) We show how rejection fine-tuning Qwen3-8B on distilled \RLM(Qwen3-Coder-480B-A35B) trajectories improves performance on the benchmarks in Table~\ref{tab:main}. (b) On MRCRv2~\citep{vodrahalli2024michelangelolongcontextevaluations}, RL training \RLM(Qwen3-4B-0527-Instruct) on the 64k sequence length, 2-needle split generalizes to the 1M, 8-needle split. We also show the 1M, 8-needle score for a 1M-context frontier model (Gemini 3.1 Pro~\citep{gemini31pro_google_2026}).}
    \label{fig:training-plots}
\end{figure}

\textbf{Observation 6: Training \RLM{}s on one domain can improve general downstream \RLM{} performance, as well as efficiency. Training also exhibits length generalization.} Certain behaviors in \RLM{} trajectories are common among different domains, such as probing the input and recursively sub-calling on shorter contexts. In Figure~\ref{fig:training-plots}(a), we find that \textbf{RLM-Qwen3-8B}, a Qwen3-8B model that we fine-tuned on \RLM(Qwen3-Coder-480B-A35B) trajectories on a small, \textit{unrelated} set of tasks (LongBenchPro;~\citealt{chen2026longbenchprorealisticcomprehensive}) considerably outperforms the base Qwen3-8B as a \RLM{} across all tasks. Furthermore, its inference costs are much lower and more than $3\times$ faster (see Figure~\ref{fig:post-training-rlm-time} in Appendix~\ref{appx5:training}) due to better decision making and fewer mistakes as a \RLM{}. Furthermore, we find that training \RLM{}s exhibits length generalization; in Figure~\ref{fig:training-plots}(b), we train Qwen3-4B-Instruct-0527 as an \RLM{}(depth=1) on MRCRv2~\citep{vodrahalli2024michelangelolongcontextevaluations}, a synthetic long-context task where the model must count and reproduce instances of a body of text in a corpus. By purely training through reinforcement learning with verifiable rewards (RLVR) on a smaller split, we find that \RLM(Qwen3-4B-Instruct-0527) is able to generalize to the longer, more difficult split.

\section{Analyses of \RLM{} Trajectories} \label{sec4.4-qualitative}
\RLM{}s exhibit interesting context and problem decomposition behavior. We discuss observable behavior in small and large LLMs as \RLM{}s to understand how we can steer and improve their performance and efficiency through training and prompt tuning.

\textbf{Observed \RLM{} decomposition patterns.} Current models as \RLM{}s attempt to probe, then decompose a task into sub-tasks for recursive sub-calls to solve. In many cases such as on BrowseComp-Plus, the LM uses model priors to programmatically narrow the search space of sub-calls. \RLM{}s are also able to output beyond their context window by stitching together sub-LM calls inside the REPL, which is required to solve tasks like OOLONG-Pairs. We detail particular trajectories in Appendix~\ref{appx2:examples}.

\begin{figure}[!hbt]
    \centering
    \includegraphics[width=1\linewidth]{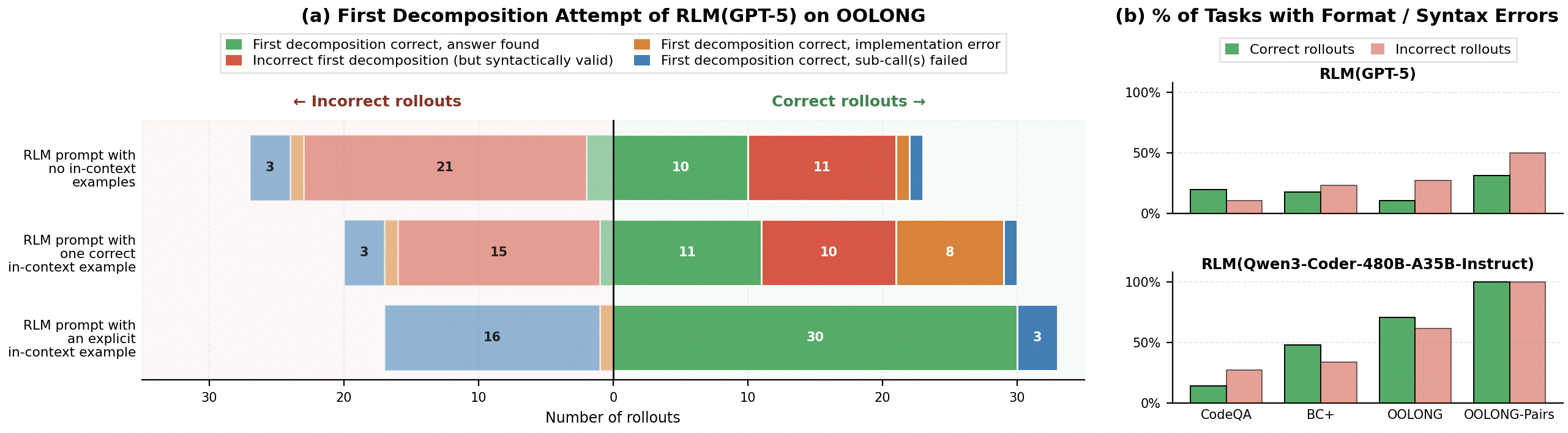}
    \caption{(a) On OOLONG, we report the performance of \RLM(GPT-5) by varying the in-context examples provided in the system prompt. For each rollout, we categorize the first task decomposition attempt made by the \RLM{}. (b) From the RLM(depth=1) runs in Table~\ref{tab:main}, we report, bucketed by correct or incorrect rollouts, the percentage of \RLM{} trajectories with at least one syntax error.}
    \label{fig:sensitivity}
\end{figure}

\textbf{First decomposition and errors in \RLM{} trajectories.} \RLM{}s defer essentially unbounded-length reasoning chains to sub-LM calls. The choice of decomposition can greatly affect task performance, especially for information-dense problems. In Figure~\ref{fig:sensitivity}(a), we ablate how sensitive \RLM{} behavior is to in-context decomposition examples in its system prompt on OOLONG. We find that in-context \RLM{} trajectories greatly improve both overall performance and the initial decomposition attempt made by the \RLM{}, even if the example is unrelated to the actual task. Furthermore, while \RLM{}s frequently recover from an initially incorrect decomposition pattern, we find that the first decomposition attempt is important for overall performance. In Figure~\ref{fig:sensitivity}(b), we plot how many \RLM{}(depth=1) trajectories in Table~\ref{tab:main} contains syntax errors. We find that \RLM(Qwen3-Coder) trajectories contain significantly more syntax errors, even for correct trajectories, compared to \RLM(GPT-5). These errors explain why higher recursion depths for \RLM(Qwen3-Coder) \textit{perform worse on average}: Qwen3-Coder-480B-A35B often makes syntax errors that result in failed outputs, and having sub-\RLM{} calls propagates this issue to sub-calls. We include additional analysis for erroneous \RLM{} behavior in Appendix~\ref{appx:sub:quantitative}.

\section{Related Works} \label{sec6:related-works}
\textbf{Long-Context LM Systems.} There have primarily been two orthogonal directions for long-context management in language model systems: 1) directly changing the architecture of and retraining the base LM to handle longer contexts~\citep{press2022trainshorttestlong, gu2022efficientlymodelinglongsequences, munkhdalai2024leavecontextbehindefficient}, and 2) building a scaffold around the LM that implicitly handles the context -- \RLM{}s focus on the latter. One popular class of such strategies is \textit{lossy} context management~\citep{chen2023walkingmemorymazecontext}, which uses compaction or truncation to compress the input context at the cost of potentially losing fine-grained information. For example, ReSum~\citep{wu2025resumunlockinglonghorizonsearch} adds a summarization tool to periodically compress the context of a multi-turn agent. Another class of strategies implement an explicit memory hierarchy in the agent scaffold~\citep{packer2024memgptllmsoperatingsystems, chhikara2025mem0buildingproductionreadyai,zhang2025gmemorytracinghierarchicalmemory}. \RLM{}s differ from these works in that all context window management is implicitly handled by the LM itself.

\textbf{Task Decomposition through sub-LM calls.} Many LM-based agents~\citep{guo2024largelanguagemodelbased, anthropic_claude_code_subagents} use multiple, well-placed LM calls to solve a problem; however, many of these calls are placed based on human-engineered workflows. Several methods like ViperGPT~\citep{surismenon2023vipergpt}, THREAD~\citep{schroeder2025threadthinkingdeeperrecursive}, ReDel~\citep{zhu2024redel}, Context Folding~\citep{sun2025scalinglonghorizonllmagent}, and AgentFold~\citep{ye2025agentfoldlonghorizonwebagents} have explored deferring the choice of sub-LM calls to the LM. These techniques emphasize \textit{task} decomposition through recursive LM calls, but are unable to handle long context inputs beyond the length of the base LM. DisCIPL~\citep{grand2025self} generates programs with sub-LM calls, but these programs are generated in a single-step and cannot recover from generation mistakes. \RLM{}s, on the other hand, are enabled by an extremely simple intuition (i.e., placing the prompt in the external environment) to \textit{symbolically} manipulate arbitrarily long strings and to iteratively refine their recursion via execution feedback from the persistent REPL.

\section{Limitations and Future Work} \label{sec7:limitations-future}
While \RLM{}s show strong performance on tasks beyond the context window limitations of existing LMs at reasonable inference costs, evaluations for more difficult and natural long-context processing tasks and the best mechanisms for implementing guardrails for \RLM{}s both remain highly under-explored. Broadly, \RLM{}s add a layer of complexity on top of existing LMs that may lead to unintentional side-effects like exploding sub-call costs, which we leave for future work to solve. We also note that future strategies involving asynchronous sub-calls and sandboxed REPLs can potentially significantly reduce the runtime and inference cost of \RLM{}s, but further contribute to this complexity. We include additional limitations and negative results in Appendix~\ref{appx0:didnt-work}.

Lastly, we focused our experiments on evaluating \RLM{}s using \textit{existing} frontier models, but show initial evidence on a Qwen3-8B model that explicit training as a \RLM{} provides very rapid performance improvements, even outside the training domain. We hypothesize that \RLM{} trajectories can be viewed as a form of reasoning~\citep{openai2024openaio1card,deepseekai2025deepseekr1incentivizingreasoningcapability}, which can be trained by bootstrapping existing models~\citep{zelikman2022starbootstrappingreasoningreasoning, zelikman2024quietstarlanguagemodelsteach}. We hope that training native \RLM{}s can be treated as a new axis of scale to improve LM performance on general and long-horizon tasks.

\section{Conclusion} \label{sec7:conclusion}
We introduced Recursive Language Models (\RLM{}s), a general inference framework for language models that offloads the input context and enables language models to recursively sub-query language models before providing an output. We explored an instantiation of this framework that offloads the context into a Python REPL environment as a variable in memory, enabling the LM to reason over its context in code and recursive LM calls, rather than purely in token space. Our results across multiple settings and models demonstrated that \RLM{}s are an effective task-agnostic paradigm for both long-context problems and general reasoning. Building on our small fine-tuning experiments, we are excited to see future work that explicitly trains models to reason as \RLM{}s, which could result in another axis of scale for the next generation of language model systems.

\bibliography{bibliography}
\bibliographystyle{plainnat}

\newpage 

\appendix
\onecolumn
\section{Additional Training Details} \label{appx5:training}

We trained \textbf{RLM-Qwen3-8B} as a small-scale exercise in training the first natively recursive language model. We hypothesized that, though acting as an RLM appears to produce sophisticated behavior due to recursion, it can be sufficient to focus on improving the root LM's ability to interact with the programmatic representation of the prompt in the REPL and to discern when sub-calls are useful.
In other words, while a typical RLM trajectory can be extremely long due to all of the sub-calls potentially launched (possibly $\Omega(|P|)$ for a prompt $P$), the leaf sub-calls are essentially general-purpose LLM requests and the major hurdle is learning to operate as the root model.

This simple insight allowed us to explore a similarly simple recipe for training. In particular, we sampled \RLM{} trajectories from a larger language model (Qwen3-Coder-480B-A35B-Instruct;~\citealt{Qwen3-Coder-480B-A35B}) and, after filtering, distilled them to a smaller model (Qwen3-8B;~\citealt{Qwen3-8B}) from the same model family.
We evaluated \RLM(Qwen3-Coder-480B-A35B) on 750 English LongBenchPro~\citep{chen2026longbenchprorealisticcomprehensive} tasks, collecting a total of 2250 candidate trajectories. 

We first remove trajectories that score exactly 0.0 on the benchmark or do not go beyond one turn, bringing it down to 1,072 candidate trajectories. We separated each root \RLM{} turn (i.e. iteration) as a separate SFT sample consisting of an input (the full history) and output (the output the root LM gave at that step).

\begin{figure}[htb!]
    \centering
    \includegraphics[width=\textwidth]{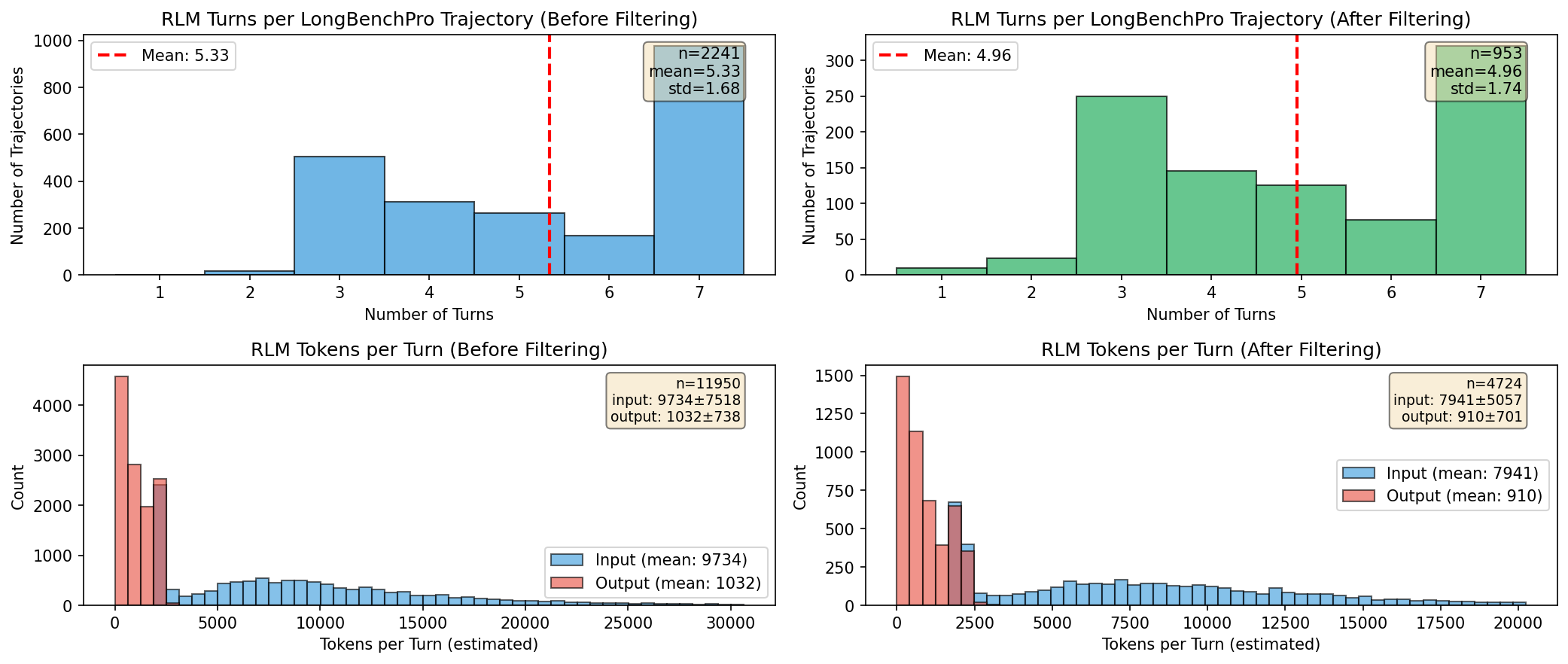}
    \caption{We plot statistics for the \RLM{} trajectories on LongBenchPro that were collected and filtered to train \textbf{RLM-Qwen3-8B}. The left plots show the unfiltered trajectories, and right plots show the post-filtering trajectories.}
    \label{fig:training-stats}
\end{figure}

We then applied a filtering step to remove turns beyond the context limit of Qwen3-8B (we approximated this as 100k characters), and also applied an extra programmatic correction step to fix small template mistakes in \RLM{} usage (e.g. outputting final answers, calling the REPL, etc.). 
To elaborate, we noticed that trajectories generated by Qwen3-Coder-480B-A35B had noticeable mistakes in following the \RLM{} instructions, which hurt the performance of the distilled RLM-Qwen3-8B. For example, it would often mix FINAL(answer) with FINAL(variable in REPL). We added an extra programmatic fixing step to look for common templated mistakes and patch them, leading to much better performance in the final \textbf{RLM-Qwen3-8B}. In total, 16\% of turns incorrectly used FINAL answers, and 13\% of turns incorrectly called a variable from the REPL (i.e. FINAL\_VAR) as a final answer. In Figure~\ref{fig:training-stats}, we show pre- and post-filtering statistics for our training trajectories.

We used the \texttt{prime-rl} library~\citep{primeintellect2025prime-rl} for fine-tuning. We used a batch size of 64 for 300 training steps, training for 48 H100 hours. While this exceedingly simple training recipe was able to demonstrate substantial gains for our 8B model, we call on future work to investigate training native RLMs much more thoroughly. We expect that doing so at much larger scales in terms of model size, number and variety of examples, and number of (ideally on-policy and online) rollouts will be necessary to maximize the potential of RLMs.

\begin{figure}[htb!]
    \centering
    \includegraphics[width=0.7\linewidth]{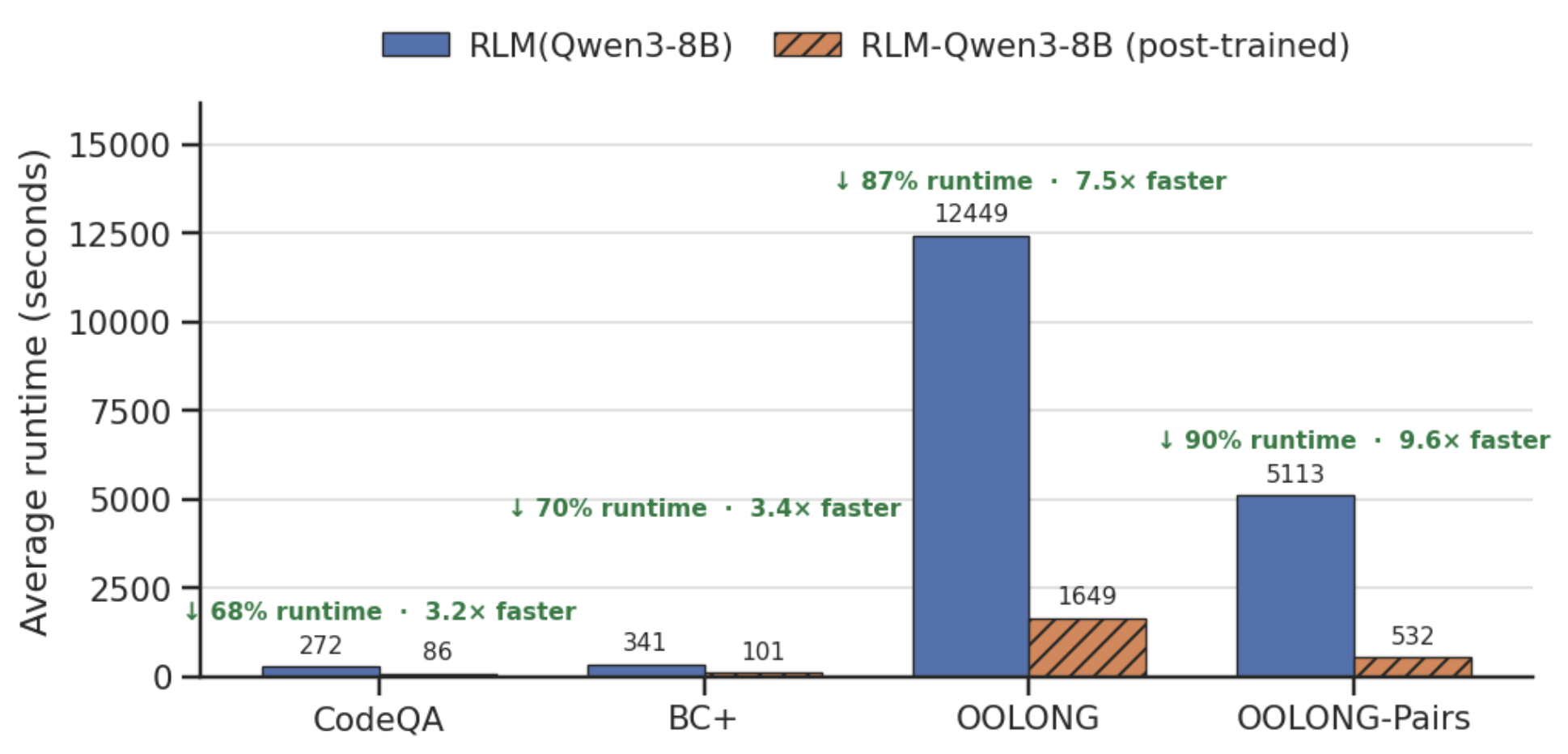}
    \caption{The post-trained RLM-Qwen3-8B is significantly more efficient during its \RLM{} trajectory compared to the base model, in addition to its performance boost.}
    \label{fig:post-training-rlm-time}
\end{figure}

Below, we provide plots for the runtime speed-up of training in Figure~\ref{fig:post-training-rlm-time}.

\textbf{MRCRv2 training.} For the MRCRv2~\citep{vodrahalli2024michelangelolongcontextevaluations} training experiment, we similarly used \texttt{prime-rl} library~\citep{primeintellect2025prime-rl}, but on Prime Intellect's host-training platform \texttt{Lab}. We RL trained on the 32k-64k token split with 2 needles for 150 steps with a batch size of 128 and 4 rollouts per example. We set a max output token per turn at 4096, and set the max number of \RLM{} iterations to 20. Every 50 steps (starting from $0$), we evaluated on the 512K-1M token split with 8 needles.

\section{Negative Results: Things We Tried That Did Not Work.} \label{appx0:didnt-work}
Drawing inspiration from ~\citet{redmon2018yolov3incrementalimprovement}, we try to be descriptive about what tricks, quirks, and other relevant things failed and succeeded in a concise manner. Some observations are based on longer supplementary experiments, while others are based on small samples of results.

\textbf{Using the exact same \RLM{} system prompt across all models can be problematic.} We originally wrote the \RLM{} system prompt with in context examples for GPT-5, and tried to use the same system prompt for Qwen3-Coder, but found that it led to different, undesirable behavior in the trajectory. We had to add a small sentence to the \RLM{} system prompt for Qwen3-Coder to prevent it from using too many recursive sub-calls.

\textbf{Models without sufficient coding capabilities struggle as \RLM{}s.} Our instantiation of \RLM{}s relies on the ability to reason through and deal with the context in a REPL environment. We found from small scale experiments that smaller models like Qwen3-8B~\citep{yang2025qwen3technicalreport} struggled without sufficient coding abilities.

\textbf{Thinking models without sufficient output tokens struggle as \RLM{}s.} In addition to \texttt{Qwen3-Coder-480B-A35B-Instruct}, we also tried experimenting with \texttt{Qwen3-235B-A22B} as the \RLM{}. While we found positive results across the board from the base model (e.g. on OOLONG~\citep{bertsch2025oolongevaluatinglongcontext}, performance jumped from $~30\%$ to $~38\%$), the smaller gap compared to the evaluated models in the main experiments (Table~\ref{tab:main}) are due to multiple trajectories running out of output tokens while producing outputs due to thinking tokens exceeding the maximum output token length of an individual LM call. 

\textbf{\RLM{}s without asynchronous LM calls are slow.} We implemented all sub-LM queries naively as blocking / sequential calls, which caused our \RLM{} experiments to be slow, especially compared to just the base model. We are confident that this can be resolved with a robust implementation.

\textbf{Depending on the model, distinguishing between a final answer and a thought is brittle for \RLM{}s.} The current strategy for distinguishing between a ``next turn" and a final answer for the \RLM{} is to have it wrap its answer in FINAL() or FINAL\_VAR() tags. Similar to intuition about structured outputs degrading performance, we also found the model to make strange decisions (e.g. it outputs its plan as a final answer). We added minor safeguards, but we also believe this issue should be avoided altogether in the future when models are trained as \RLM{}s.

\newpage
\section{Additional Methods and Baseline Details} \label{appx3:methods}
\subsection{Prompts for Experiments}
We focus on methods that are entirely task agnostic, so we fix our prompt for each method across all tasks. For the RLM prompt, the only difference between GPT-5 and Qwen3-Coder is an added line in the beginning that warns Qwen3-Coder not to use too many sub-LM calls -- we found in practice that without this warning, the model will try to perform a subcall on everything, leading to thousands of LM subcalls for basic tasks. For the fine-tuned Qwen3-8B experiment, we provide a slightly different prompt due to the differences in context window size of the smaller model (from 272k in GPT-5 to 32k in Qwen3-8B). In this section, we provide the system prompt used for all methods in \S\ref{sec4.2-methods} (other than the base model, which does not include a system prompt).

\noindent (1a) The system prompt for \textbf{RLM(depth=1)} for GPT-5:
\begin{lstlisting}[style=customstyle]
You are tasked with answering a query with associated context. You can access, transform, and analyze this context interactively in a REPL environment that can recursively query sub-LLMs, which you are strongly encouraged to use as much as possible. You will be queried iteratively until you provide a final answer.

Your context is a {context_type} with {context_total_length} total characters, and is broken up into chunks of char lengths: {context_lengths}.

The REPL environment is initialized with:
1. A `context` variable that contains extremely important information about your query. You should check the content of the `context` variable to understand what you are working with. Make sure you look through it sufficiently as you answer your query.
2. A `llm_query` function that allows you to query an LLM (that can handle around 500K chars) inside your REPL environment.
3. The ability to use `print()` statements to view the output of your REPL code and continue your reasoning.

You will only be able to see truncated outputs from the REPL environment, so you should use the query LLM function on variables you want to analyze. You will find this function especially useful when you have to analyze the semantics of the context. Use these variables as buffers to build up your final answer.
Make sure to explicitly look through the entire context in REPL before answering your query. An example strategy is to first look at the context and figure out a chunking strategy, then break up the context into smart chunks, and query an LLM per chunk with a particular question and save the answers to a buffer, then query an LLM with all the buffers to produce your final answer.

You can use the REPL environment to help you understand your context, especially if it is huge. Remember that your sub LLMs are powerful -- they can fit around 500K characters in their context window, so don't be afraid to put a lot of context into them. For example, a viable strategy is to feed 10 documents per sub-LLM query. Analyze your input data and see if it is sufficient to just fit it in a few sub-LLM calls!

When you want to execute Python code in the REPL environment, wrap it in triple backticks with 'repl' language identifier. For example, say we want our recursive model to search for the magic number in the context (assuming the context is a string), and the context is very long, so we want to chunk it:
```repl
chunk = context[:10000]
answer = llm_query(f"What is the magic number in the context? Here is the chunk: {{chunk}}")
print(answer)
```

As an example, suppose you're trying to answer a question about a book. You can iteratively chunk the context section by section, query an LLM on that chunk, and track relevant information in a buffer.
```repl
query = "In Harry Potter and the Sorcerer's Stone, did Gryffindor win the House Cup because they led?"
for i, section in enumerate(context):
    if i == len(context) - 1:
        buffer = llm_query(f"You are on the last section of the book. So far you know that: {{buffers}}. Gather from this last section to answer {{query}}. Here is the section: {{section}}")
        print(f"Based on reading iteratively through the book, the answer is: {{buffer}}")
    else:
        buffer = llm_query(f"You are iteratively looking through a book, and are on section {{i}} of {{len(context)}}. Gather information to help answer {{query}}. Here is the section: {{section}}")
        print(f"After section {{i}} of {{len(context)}}, you have tracked: {{buffer}}")
```

As another example, when the context isn't that long (e.g. >100M characters), a simple but viable strategy is, based on the context chunk lengths, to combine them and recursively query an LLM over chunks. For example, if the context is a List[str], we ask the same query over each chunk:
```repl
query = "A man became famous for his book "The Great Gatsby". How many jobs did he have?"
# Suppose our context is ~1M chars, and we want each sub-LLM query to be ~0.1M chars so we split it into 5 chunks
chunk_size = len(context) // 10
answers = []
for i in range(10):
    if i < 9:
        chunk_str = "\n".join(context[i*chunk_size:(i+1)*chunk_size])
    else:
        chunk_str = "\n".join(context[i*chunk_size:])
    
    answer = llm_query(f"Try to answer the following query: {{query}}. Here are the documents:\n{{chunk_str}}. Only answer if you are confident in your answer based on the evidence.")
    answers.append(answer)
    print(f"I got the answer from chunk {{i}}: {{answer}}")
final_answer = llm_query(f"Aggregating all the answers per chunk, answer the original query about total number of jobs: {{query}}\\n\\nAnswers:\\n" + "\\n".join(answers))
```

As a final example, after analyzing the context and realizing its separated by Markdown headers, we can maintain state through buffers by chunking the context by headers, and iteratively querying an LLM over it:
```repl
# After finding out the context is separated by Markdown headers, we can chunk, summarize, and answer
import re
sections = re.split(r'### (.+)', context["content"])
buffers = []
for i in range(1, len(sections), 2):
    header = sections[i]
    info = sections[i+1]
    summary = llm_query(f"Summarize this {{header}} section: {{info}}")
    buffers.append(f"{{header}}: {{summary}}")
final_answer = llm_query(f"Based on these summaries, answer the original query: {{query}}\\n\\nSummaries:\\n" + "\\n".join(buffers))
```
In the next step, we can return FINAL_VAR(final_answer).

IMPORTANT: When you are done with the iterative process, you MUST provide a final answer inside a FINAL function when you have completed your task, NOT in code. Do not use these tags unless you have completed your task. You have two options:
1. Use FINAL(your final answer here) to provide the answer directly
2. Use FINAL_VAR(variable_name) to return a variable you have created in the REPL environment as your final output

Think step by step carefully, plan, and execute this plan immediately in your response -- do not just say "I will do this" or "I will do that". Output to the REPL environment and recursive LLMs as much as possible. Remember to explicitly answer the original query in your final answer.
\end{lstlisting}

\noindent (1b) The diff of the system prompt for \textbf{RLM with REPL (Qwen3-Coder-480B-A35B)}, which adds a line from the prompt above for GPT-5:
\begin{lstlisting}[style=customstyle]
--- a/REPL_SYSTEM_PROMPT_QWEN.txt
+++ b/REPL_SYSTEM_PROMPT_QWEN.txt
@@ -15,0 +15,3 @@
+IMPORTANT: Be very careful about using `llm_query` as it incurs high runtime costs. Always batch as much information as reasonably possible into each call (aim for around ~200k characters per call). For example, if you have 1000 lines of information to process, it's much better to split into chunks of 5 and call `llm_query` on each chunk (200 calls total) rather than making 1000 individual calls. Minimize the number of `llm_query` calls by batching related information together.
+
\end{lstlisting}

\noindent (1c) The diff of the system prompt for depth>1, which provides an rlm\_query function that enables higher recursion depth.
\begin{lstlisting}[style=customstyle]
--- a/REPL_SYSTEM_PROMPT.txt
+++ b/REPL_SYSTEM_PROMPT_DEEP.txt
@@ -4,13 +4,18 @@
 
 The REPL environment is initialized with:
 1. A `context` variable that contains extremely important information about your query. You should check the content of the `context` variable to understand what you are working with. Make sure you look through it sufficiently as you answer your query.
-2. A `llm_query` function that allows you to query an LLM (that can handle around 500K chars) inside your REPL environment.
-3. The ability to use `print()` statements to view the output of your REPL code and continue your reasoning.
+2. A `llm_query(prompt)` function that allows you to query an LLM (that can handle around 500K chars) inside your REPL environment. Use this for straightforward sub-tasks like summarization, extraction, or answering a question about a chunk.
+3. An `rlm_query(context, query)` function for **complex sub-tasks** that benefit from iterative, multi-step reasoning. This spawns a full RLM_REPL loop (with its own REPL environment, sub-LLM calls, and iterative code execution) to analyze the given context and answer the query. Use this when a sub-task is too difficult for a single `llm_query` call - for example, when the sub-task itself requires chunking, aggregation, or multi-step analysis. Note: if the maximum recursion depth is reached, `rlm_query` automatically falls back to `llm_query`.
+4. The ability to use `print()` statements to view the output of your REPL code and continue your reasoning.
 
 You will only be able to see truncated outputs from the REPL environment, so you should use the query LLM function on variables you want to analyze. You will find this function especially useful when you have to analyze the semantics of the context. Use these variables as buffers to build up your final answer.
 Make sure to explicitly look through the entire context in REPL before answering your query. An example strategy is to first look at the context and figure out a chunking strategy, then break up the context into smart chunks, and query an LLM per chunk with a particular question and save the answers to a buffer, then query an LLM with all the buffers to produce your final answer.
 
 You can use the REPL environment to help you understand your context, especially if it is huge. Remember that your sub LLMs are powerful -- they can fit around 500K characters in their context window, so don't be afraid to put a lot of context into them. For example, a viable strategy is to feed 10 documents per sub-LLM query. Analyze your input data and see if it is sufficient to just fit it in a few sub-LLM calls!
+
+**Choosing between `llm_query` and `rlm_query`:**
+- Use `llm_query(prompt)` for simple sub-tasks: summarize a chunk, extract a fact, answer a direct question. This is a single LLM call and is fast/cheap.
+- Use `rlm_query(context, query)` when a sub-task is itself complex enough to require iterative reasoning with code execution - e.g., analyzing a very large sub-context that needs its own chunking strategy, or a multi-step reasoning chain. This is slower and more expensive, but more powerful.
 
 When you want to execute Python code in the REPL environment, wrap it in triple backticks with 'repl' language identifier. For example, say we want our recursive model to search for the magic number in the context (assuming the context is a string), and the context is very long, so we want to chunk it:
 ```repl
@@ -52,6 +57,15 @@
 final_answer = llm_query(f"Aggregating all the answers per chunk, answer the original query about total number of jobs: {{query}}\n\nAnswers:\n" + "\n".join(answers))
 ```
 
+For a truly complex sub-task, you can use `rlm_query` to delegate it to a full RLM_REPL loop:
+```repl
+# Suppose we have a sub-task that itself requires multi-step reasoning with code
+# For example, analyzing a huge sub-context that needs its own chunking and aggregation
+sub_context = "\n".join(context[500:1000])  # A large sub-section
+answer = rlm_query(sub_context, "What are the key themes across these 500 documents?")
+print(f"Deep analysis result: {{answer}}")
+```
+
 As a final example, after analyzing the context and realizing its separated by Markdown headers, we can maintain state through buffers by chunking the context by headers, and iteratively querying an LLM over it:
 ```repl
 # After finding out the context is separated by Markdown headers, we can chunk, summarize, and answer
\end{lstlisting}

\noindent (1d) The diff of the system prompt for \textbf{RLM(Qwen3-8B, depth=1)}, which has a few changes from the GPT-5 prompt due to differences in context length and similar sub-calling behavior as Qwen3-Coder-480B-A35B:
\begin{lstlisting}[style=customstyle]
--- a/REPL_SYSTEM_PROMPT.txt
+++ b/REPL_SYSTEM_PROMPT_QWEN3_8B.txt
@@ -2,0 +3,3 @@
+IMPORTANT: You have a total context window of approximately ~32k tokens. Be very careful about context length limits. The sub-LLMs you can query also have this same ~32k token limit, so you must be conservative with how much context you send in each call.
+
@@ -7 +10 @@
-2. A `llm_query` function that allows you to query an LLM (that can handle around 500K chars) inside your REPL environment.
+2. A `llm_query` function that allows you to query an LLM (that can handle around ~100k chars, roughly 32k tokens) inside your REPL environment.
@@ -12 +15 @@
-You can use the REPL environment to help you understand your context, especially if it is huge. Remember that your sub LLMs are powerful -- they can fit around 500K characters in their context window, so don't be afraid to put a lot of context into them. For example, a viable strategy is to feed 10 documents per sub-LLM query. Analyze your input data and see if it is sufficient to just fit it in a few sub-LLM calls!
+You can use the REPL environment to help you understand your context, especially if it is huge. Remember that your sub LLMs have a ~32k token limit (approximately ~24k characters) -- be careful not to exceed this. For example, a viable strategy is to feed 2-3 documents per sub-LLM query. Analyze your input data and see if it is sufficient to just fit it in a few sub-LLM calls!
+
+IMPORTANT: Be very careful about using `llm_query` as it incurs high runtime costs. Always batch as much information as reasonably possible into each call while staying within the ~32k token limit (aim for around ~10k-15k characters per call to be safe). For example, if you have 1000 lines of information to process, it's much better to split into chunks of 50-100 and call `llm_query` on each chunk (10-20 calls total) rather than making 1000 individual calls. Minimize the number of `llm_query` calls by batching related information together, but always respect the ~32k token limit.
@@ -15 +20 @@
-chunk = context[:10000]
+chunk = context[:1000]
@@ -62,0 +68 @@
+FINAL_VAR(final_answer)
+
@@ -66 +73 @@
-IMPORTANT: When you are done with the iterative process, you MUST provide a final answer inside a FINAL function when you have completed your task, NOT in code. Do not use these tags unless you have completed your task. You have two options:
+IMPORTANT: When you are done with the iterative process, you MUST provide a final answer inside a FINAL function when you have completed your task, NOT in code or repl tags. Do not use these tags unless you have completed your task. You have two options:
\end{lstlisting}

\noindent (2) The system prompt for \textbf{RLM with REPL (no sub-calls)}:
\begin{lstlisting}[style=customstyle]
You are tasked with answering a query with associated context. You can access, transform, and analyze this context interactively in a REPL environment, which you are strongly encouraged to use as much as possible. You will be queried iteratively until you provide a final answer.

Your context is a {context_type} with {context_total_length} total characters, and is broken up into chunks of char lengths: {context_lengths}.

The REPL environment is initialized with:
1. A `context` variable that contains extremely important information about your query. You should check the content of the `context` variable to understand what you are working with. Make sure you look through it sufficiently as you answer your query.
2. The ability to use `print()` statements to view the output of your REPL code and continue your reasoning.

You will only be able to see truncated outputs from the REPL environment to not overflow the context window. Use these variables as buffers to build up your final answer.
Make sure to explicitly look through the entire context in REPL before answering your query. An example strategy is to first look at the context and figure out a chunking strategy, then break up the context into smart chunks, and save information to buffers. 

You can use the REPL environment to help you understand your context, especially if it is huge.

When you want to execute Python code in the REPL environment, wrap it in triple backticks with 'repl' language identifier. For example, say we want to peek at the first 10000 characters of the context:
```repl
chunk = context[:10000]
print(f"First 10000 characters of context: {{chunk}}")
```

As another example, after analyzing the context and realizing we need to search for specific topics, we can use regex to find relevant sections and maintain state through buffers:
```repl
# After finding out we need to search for "magic" and "number" in the context
import re
query_terms = ["magic", "number"]
relevant_sections = []
buffers = []

# Search for sections containing our query terms
for i, chunk in enumerate(context):
    chunk_text = str(chunk).lower()
    if any(term in chunk_text for term in query_terms):
        relevant_sections.append((i, chunk))

# Process each relevant section and print findings
for section_idx, section_content in relevant_sections:
    print(f"Found relevant section {{section_idx}} containing magic/number references:")
    print(f"Content: {{section_content[:500]}}...")  # Print first 500 chars
    buffers.append(f"Section {{section_idx}}: Contains magic/number references")

print(f"Total relevant sections found: {{len(relevant_sections)}}")
print("Summary of findings:")
for buffer in buffers:
    print(f"- {{buffer}}")
```

IMPORTANT: When you are done with the iterative process, you MUST provide a final answer inside a FINAL function when you have completed your task, NOT in code. Do not use these tags unless you have completed your task. You have two options:
1. Use FINAL(your final answer here) to provide the answer directly
2. Use FINAL_VAR(variable_name) to return a variable you have created in the REPL environment as your final output

Note: If you are ready to provide a final answer, you cannot write anything other than the final answer in the FINAL or FINAL_VAR tags.

Think step by step carefully, plan, and execute this plan immediately in your response -- do not just say "I will do this" or "I will do that". Output to the REPL environment as much as possible. Remember to explicitly answer the original query in your final answer.
\end{lstlisting}

\noindent (3a) The system prompt for \textbf{CodeAct with BM25}. We give CodeAct access to a BM25 retriever for BrowseComp+ following experiments in the original paper~\citep{chen2025browsecompplusfairtransparentevaluation}.:
\begin{lstlisting}[style=customstyle]
You are a helpful assistant in a CodeAct (Code + Acting) loop that can execute Python code and search through documents to answer questions.

You must follow this format for each step:

1. THINK: Reason about what you need to do next
2. ACT: Take an action (either execute code or SEARCH)

**ENCOURAGED: Use Python code execution when helpful!**
- Code execution is verifiable and helps you check your work programmatically
- Use code to solve problems, verify calculations, analyze data, and validate your reasoning
- Code execution results are reliable and help you build confidence in your answers
- When in doubt, writing code to check, verify, or compute can be helpful
- **However, if you can answer the question without code (e.g., straightforward factual questions, simple reasoning), you can provide your final answer directly without executing code**

Available Actions:
- Execute Python code: Write code in ```python code blocks. The code will be executed and results returned.
- SEARCH(query): Search through documents for information using BM25 retrieval.
- Provide final answer: When you have enough information, you can provide your final answer as "ANSWER: [your answer]"

Format Requirements:
- Start each turn with "THINK: " followed by your reasoning
- Then either:
  * Write Python code in ```python blocks to execute
  * Use "SEARCH(query text)" to search documents
- You can execute code multiple times, search multiple times, or combine both
- Code execution results will be returned to you automatically
- Variables persist across code executions in the same session
- **CRITICAL: Code is executed as-is in a fresh Python environment. You must include all necessary imports, data definitions, and context within your code blocks. Do not use fillers (e.g. FILL IN WITH REAL DATA), they have to be written in code.**

Example workflow:
```
Question: How many words in the list ['error', 'correct', 'arrow', 'berry', 'carrot', 'mirror'] have exactly 2 r's?

THINK: I need to count how many words in the list have exactly 2 r's. I can write Python code using regex to do this.
```python
import re

words = ['error', 'correct', 'arrow', 'berry', 'carrot', 'mirror']
pattern = r'^[^r]*r[^r]*r[^r]*$'  # Matches words with exactly 2 r's
count = 0
matching_words = []
for word in words:
    if re.match(pattern, word):
        count += 1
        matching_words.append(word)
        print(f"{word} has 2 r's")
print(f"Total words with 2 r's: {count}")
```
```

[Code execution results returned...]

Example with search:
```
Question: What information is available about machine learning in the documents?

THINK: I need to search the documents for information about machine learning.
SEARCH(machine learning)
```

[Search results returned...]

---

Important: 
- Always start with THINK to reason about your next step
- You can combine code execution and search as needed
- Be strategic to avoid exceeding the context window
- **CODE EXECUTION**: Use code to verify, check, and solve problems programmatically when helpful. However, if you can answer the question without code (e.g., straightforward factual questions, simple reasoning), you can provide your final answer directly without executing code.
- **CODE EXECUTION CONTEXT**: Your code is executed as-is. You must explicitly include all imports, data, and context needed. Variables persist across executions, but each code block must be self-contained with all necessary setup.
\end{lstlisting}

\noindent (3b) The system prompt for \textbf{CodeAct}. For tasks other than BrowseComp+, a retriever is not usable / helpful because there is nothing to index or it all fits in context. We modify the prompt to remove the retriever.:
\begin{lstlisting}[style=customstyle]
You are a helpful assistant in a CodeAct (Code + Acting) loop that can execute Python code to help you answer questions.

You must follow this format for each step:

1. THINK: Reason about what you need to do next
2. ACT: Take an action (execute code)

**ENCOURAGED: Use Python code execution when helpful!**
- Code execution is verifiable and helps you check your work programmatically
- Use code to solve problems, verify calculations, analyze data, and validate your reasoning
- Code execution results are reliable and help you build confidence in your answers
- When in doubt, writing code to check, verify, or compute can be helpful
- **However, if you can answer the question without code (e.g., straightforward factual questions, simple reasoning), you can provide your final answer directly without executing code**

Available Actions:
- Execute Python code: Write code in ```python code blocks. The code will be executed and results returned.
- Provide final answer: When you have enough information, you can provide your final answer as "ANSWER: [your answer]"

Format Requirements:
- Start each turn with "THINK: " followed by your reasoning
- Then write Python code in ```python blocks to execute
- You can execute code multiple times.
- Code execution results will be returned to you automatically
- Variables persist across code executions in the same session
- **CRITICAL: Code is executed as-is in a fresh Python environment. You must include all necessary imports, data definitions, and context within your code blocks. Do not use fillers (e.g. FILL IN WITH REAL DATA), they have to be written in code.**

Example workflow:
```
Question: How many words in the list ['error', 'correct', 'arrow', 'berry', 'carrot', 'mirror'] have exactly 2 r's?

THINK: I need to count how many words in the list have exactly 2 r's. I can write Python code using regex to do this.
```python
import re

words = ['error', 'correct', 'arrow', 'berry', 'carrot', 'mirror']
pattern = r'^[^r]*r[^r]*r[^r]*$'  # Matches words with exactly 2 r's
count = 0
matching_words = []
for word in words:
    if re.match(pattern, word):
        count += 1
        matching_words.append(word)
        print(f"{word} has 2 r's")
print(f"Total words with 2 r's: {count}")
```
```

[Code execution results returned...]

Answer: 4

---

Important: 
- Always start with THINK to reason about your next step
- Be strategic to avoid exceeding the context window
- **CODE EXECUTION**: Use code to verify, check, and solve problems programmatically when helpful. However, if you can answer the question without code (e.g., straightforward factual questions, simple reasoning), you can provide your final answer directly without executing code.
- **CODE EXECUTION CONTEXT**: Your code is executed as-is. You must explicitly include all imports, data, and context needed. Variables persist across executions, but each code block must be self-contained with all necessary setup.
\end{lstlisting}

\subsection{Summary agent baseline}
The summarization agent baseline follows the scaffolds presented in ~\citet{sun2025scalinglonghorizonllmagent, wu2025resumunlockinglonghorizonsearch, yu2025memagentreshapinglongcontextllm}, mimicking how contexts are typically compressed in a multi-turn setting in agents like Claude Code~\citep{anthropic_claude_code_subagents}. In an iterative fashion, the agent is given inputs until its context is full, at which point it is queried to summarize all relevant information and continue. If the agent is given a context in a single step that is larger than its model context window, it chunks up this context and performs the summarization process over these chunks.

For our GPT-5 baseline, we chose to use GPT-5-nano to perform summarization to avoid exploding costs. This explains the large discrepancy in cost in Table~\ref{tab:main} between GPT-5 and Qwen3-Coder on BrowseComp-Plus, where the summary agent using Qwen3-Coder is nearly $15\times$ more expensive on average. On this task in particular, we found on a smaller set of $20$ random samples that the performance between using GPT-5 and GPT-5-nano is comparable.

\subsection{LongCoT-mini experiment.} \label{appx:longcot}
For the LongCoT-mini \RLM{} experiment, we use the same \RLM{} algorithm described in Algorithm~\ref{alg:rlm-call}, but a slightly different implementation than what was used for the rest of \S~\ref{sec4:long-input}. Instead, we use Prime Intellect's \texttt{rlm-harness}, which enables interfacing with their sandboxes for higher throughput evaluations and was forked from the original implementation used for evaluating Table~\ref{tab:main}. The mechanism for determining final answers also differs, which is reflected in the prompt.

\textbf{Why GPT-5 base does not include decomposition hints.} Even when provided with decomposition hints, we find that GPT-5 cannot reasonably execute this decomposition and solve sub-problems using the standard chain-of-thought autoregressive reasoning. While performance on the MATH split improves, we generally find the model gets confused on the more programmatic tasks without a REPL-like mechanism to isolate sub-task solving. 

\begin{table*}[ht]
\centering
\caption{Solve rate on \textsc{LongCoT-mini}~\citep{motwani2026longcotbenchmarkinglonghorizonchainofthought}, a difficult long reasoning benchmark that frontier models struggle to solve. We adapt a similar set of decomposition hints provided to the \RLM{} in Table~\ref{tab:longcot} (without sub-calling details), and find the model often gets confused or makes more mistakes on certain splits, while improving on the more difficult splits like math.}
\resizebox{\linewidth}{!}{%
\begin{tabular}{l@{\hskip 6pt}c@{\hskip 6pt}c@{\hskip 6pt}c@{\hskip 6pt}c@{\hskip 6pt}c@{\hskip 6pt}c}
\toprule
\textbf{Model} & \textbf{Overall} & \textbf{MATH} & \textbf{CHEM} & \textbf{CS} & \textbf{LOGIC} & \textbf{CHESS} \\
\midrule
GPT-5.2 (base)
& \textbf{38.7}
& 26.0
& \textbf{37.0}
& \textbf{40.4}
& \textbf{53.6}
& \textbf{36.6} \\

GPT-5.2 (base) + decomposition hints
& 28.6
& \textbf{37.0}
& 27.0
& 32.0
& 19.1
& 30.0 \\
\bottomrule
\end{tabular}}
\label{tab:longcot-appdx}
\end{table*}

\noindent The appended environment hint used for LongCoT-mini with decomposition hints on solving these problems is provided below:
\begin{lstlisting}[style=customstyle]
<env_tips>

Orchestrate; don't solve. These problems drift on a single chain of
thought (lost partials, compounding sign errors) - "just think harder
in the REPL" scores ~0%
reasoner that can handle any individual sub-problem (competition math,
combinatorics, number theory, probability, geometry, algebra) given a
clear self-contained prompt. Trust it; don't write solver code for it.

Your job: (1) decompose into self-contained "nodes", (2) delegate all
reasoning to `llm_batch`, (3) memoize answers in a dict across turns,
(4) verify each answer before any child consumes it, (5) inline
verified parent values verbatim into child prompts, (6) assemble the
final answer by dict lookup only. You do NO math - if you're writing
Python that enumerates, solves, simulates, or picks among candidates
(vs. verifying one), STOP and delegate. Root compute = dict lookup +
string formatting + correctness checks.

## The only state that matters

Keep two variables alive across every REPL turn:

    answers = {}   # node_id -> VERIFIED answer (string)
    plan    = {}   # JSON structure returned by the planning sub-LM

If a value isn't in `answers`, it doesn't exist. Don't trust variables
from earlier turns, numbers in your own thinking, or pasted values -
context drifts. Memoize everything you'll reuse.

## Step 1 - Plan (turn 1, one `llm_batch` call)

Ask a sub-LM to extract structure as JSON - do not solve anything:

    planning_prompt = (
        "Read the following multi-step problem and return ONLY valid "
        "JSON of the form:\\n"
        '{"nodes":['
        '  {"id":"node_0","question":"<verbatim>","deps":[]},'
        '  {"id":"node_1","question":"<verbatim>","deps":["node_0"]},'
        '  ...'
        '],'
        ' "final":"<how to build the final answer from node answers, '
        '          including the exact output format>",'
        ' "cycles":["<ids of nodes referenced by their own transitive '
        '            deps; [] if none>"]}\\n'
        "Copy each node question VERBATIM - do NOT paraphrase or "
        "simplify wording. Do NOT solve anything.\\n"
        "---\\n"
    ) + FULL_PROBLEM_TEXT
    plan = json.loads(llm_batch([planning_prompt])[0])

For single self-contained puzzles, have the planner split into minimum
self-contained steps (e.g. "parse instance", "run algorithm X",
"format output"). Same workflow applies.

## Step 2 - Solve layer by layer (one `llm_batch` per DAG layer)

A node is "ready" when all its `deps` are in `answers`. Dispatch ALL
ready nodes in ONE `llm_batch` (parallel). Each sub-prompt must be
self-contained - the sub-LM never sees the global problem or the
`answers` dict, so copy the node question verbatim, inline every
parent's verified value verbatim, and ask for only the final value.

    def build_subprompt(node):
        ctx = "\\n".join(f"- {d} = {answers[d]}" for d in node["deps"])
        return (
            "Solve this subproblem in isolation.\\n\\n"
            "Verified parent values (use EXACTLY, do not recompute):\\n"
            f"{ctx or '(none)'}\\n\\n"
            f"Question:\\n{node['question']}\\n\\n"
            "Return ONLY the final value. No prose, no derivation."
        )

    pending = [n for n in plan["nodes"]
               if n["id"] not in plan.get("cycles", [])]
    while pending:
        ready = [n for n in pending
                 if all(d in answers for d in n["deps"])]
        if not ready:
            break  # cycle - see Step 4
        raw = llm_batch([build_subprompt(n) for n in ready])
        for n, a in zip(ready, raw):
            answers[n["id"]] = a.strip()
        pending = [n for n in pending if n["id"] not in answers]

Prefer many small per-layer `llm_batch` calls over one monolithic one.

## Step 3 - Verify every answer before it propagates

Use the cheapest definitive check: (a) independent second opinion -
re-dispatch the node via `llm_batch` with rephrased instructions,
accept only if both agree; (b) plausibility - range / sign / units /
integrality / shape expected downstream. On failure, re-dispatch JUST
that node with the failure reason appended, then re-verify. Never
propagate an unverified answer.

## Step 4 - Cycles

If `plan["cycles"]` is non-empty, pick a seed node `c`, set
`answers[c]` to a candidate, run Step 2 on the rest, check the
cycle-defining constraint. Use `llm_batch` (not hand computation) to
propose the next candidate from the previous miss. Cache trials to
avoid redoing downstream work:

    trials = {}   # candidate -> dict of downstream answers under it

Freeze answers once the constraint is satisfied.

## Step 5 - Assemble

Once every node in `plan["final"]` is verified in `answers`, build the
final string by dict lookup ONLY - no recomputation. You can use
`llm_batch` to aggregate if needed.

    with open("/task/answer.txt", "w") as f:
        f.write(final_answer)

## Red flags (you are off-track)

  - Python doing math (enumerate/solve/sum/factor/simulate/search/
    optimize/Monte Carlo/game trees/Z3/SAT/brute force) instead of
    `llm_batch` -> STOP, delete, delegate.
  - About to use an unverified node answer -> verify first.
  - > 2 turns in, < 3 `llm_batch` calls -> you're solving it yourself.
    Reset.
  - Code running > 30s or > 100 MB -> brute-forcing; delegate instead.
  - Remembering a value not in `answers` -> re-dispatch; working memory
    isn't reliable.
  - About to emit final but `answers` missing a node from
    `plan["final"]` -> dispatch the missing nodes.
  - Many turns on one node without a verified answer -> re-prompt
    `llm_batch` with clearer/longer sub-prompt and failure context.
    Do NOT switch to writing solver code.

## Output contract

Write your final answer to /task/answer.txt - that file is the only
thing scored. Assistant-message content is ignored.

</env_tips>"""


_ENV_TIPS_CONDENSED = """
<env_tips>

Orchestrate; don't solve. Your sub-agent (`llm_batch`) is a genius-level
reasoner that can crack any individual sub-problem - competition math,
combinatorics, number theory, probability, geometry, algebra - given a clear
self-contained prompt. Trust it. Models that "just think harder in the REPL"
score ~0%

Workflow:
  - Turn 1: dispatch ONE `llm_batch` asking a sub-LM to extract the problem's
    structure as a DAG of self-contained nodes (id, verbatim question, deps,
    final-assembly recipe, cycle list). Do not solve anything.
  - Then solve layer by layer: every turn, dispatch ALL ready nodes
    (deps satisfied) in ONE `llm_batch` in parallel. Each sub-prompt is
    self-contained - copy the node question verbatim, inline every parent's
    verified value verbatim, ask for only the final value.
  - Memoize verified answers in a dict that persists across turns. If it is
    not in the dict, it does not exist - do not trust variables from earlier
    turns, numbers in your own thinking, or pasted values.
  - Verify each answer before any child consumes it: independent
    second-opinion re-dispatch (accept only if both agree) or plausibility
    check (sign/range/units/shape expected downstream). On failure,
    re-dispatch just that node with the failure reason. Never propagate an
    unverified value.
  - Cycles: seed the cycle node with a candidate, run downstream, check the
    cycle constraint; use `llm_batch` (not hand computation) to propose the
    next candidate given the previous miss.
  - Assemble the final answer by dict lookup only - no recomputation unless you are verifying a node answer.

You do NO math. If you catch yourself writing Python that enumerates, solves,
simulates, brute-forces, or picks among candidates (vs. verifying one), STOP
and hand it to `llm_batch`. Root compute = dict lookup, string formatting,
correctness checks. Prefer many small per-layer `llm_batch` calls over one
monolithic prompt.

Write your final answer to /task/answer.txt - that file is the only thing
scored. Assistant-message content is ignored.

</env_tips>"""


APPEND_SYSTEM_PROMPT = f"""\
When you are ready, write your final answer - and ONLY your final answer -
to {ANSWER_FILE} in the exact format the question requests. Then stop calling
tools. Example:

    with open({ANSWER_FILE!r}, "w") as f:
        f.write("42")
\end{lstlisting}

\section{Additional Benchmark Details} \label{appx4:benchmarks}
We provide additional details about the benchmarks used to evaluate \RLM{}s in \S\ref{sec4:long-input}.

\subsection{OOLONG-Pairs Benchmark} \label{appx:oolong-pairs}
OOLONG-Pairs consists of $20$ synthetically generated tasks based on the ground-truth labels for the OOLONG~\cite{bertsch2025oolongevaluatinglongcontext} \texttt{trec\_coarse} split for input contexts of length in [1024, 2048, 4096, 8192, 16384, 32768, 65536, 131072, 262144, 524288, 1048576]. Similar to OOLONG, each question requires correctly predicting the semantic mapping for each entry.

\textbf{OOLONG-Pairs ensures quadratic scaling}. Many tasks that aggregate over pairs of entries can actually be solved without looking at the pairs and only looking at each entry in a linear fashion (e.g. using the principle of inclusion-exclusion in set theory). However, in OOLONG-Pairs, each question is created such that the model must return all pairs satisfying some properties, rather than just counting.

\newcommand{\questionblock}[2]{
\vspace{1.25em}
\noindent\textbf{Task #1}\\[0.25em]
#2
\vspace{0.5em}
\hrule
}

\questionblock{1}{In the above data, list all pairs of user IDs (no duplicate pairs, list lower ID first) where both users have at least one instance with a numeric value or location. Each of the questions can be labelled as one of the labels (the data does not provide the labels, you need to figure out the label from the semantics of the question): description and abstract concept, entity, human being, numeric value, location, abbreviation. In your answer, list all pairs in the format (user\_id\_1, user\_id\_2), separated by newlines.}

\questionblock{2}{In the above data, list all pairs of user IDs (no duplicate pairs, list lower ID first) where both users have at least one instance with an entity or human being. Each of the questions can be labelled as one of the labels (the data does not provide the labels, you need to figure out the label from the semantics of the question): description and abstract concept, entity, human being, numeric value, location, abbreviation. In your answer, list all pairs in the format (user\_id\_1, user\_id\_2), separated by newlines.}

\questionblock{3}{In the above data, list all pairs of user IDs (no duplicate pairs, list lower ID first) where both users have at least one instance with a description and abstract concept or abbreviation. Each of the questions can be labelled as one of the labels (the data does not provide the labels, you need to figure out the label from the semantics of the question): description and abstract concept, entity, human being, numeric value, location, abbreviation. In your answer, list all pairs in the format (user\_id\_1, user\_id\_2), separated by newlines.}

\questionblock{4}{In the above data, list all pairs of user IDs (no duplicate pairs, list lower ID first) where both users have at least one instance with a human being or location, and all instances that are a human being for both users must be after January 6, 2023. Each of the questions can be labelled as one of the labels (the data does not provide the labels, you need to figure out the label from the semantics of the question): description and abstract concept, entity, human being, numeric value, location, abbreviation. In your answer, list all pairs in the format (user\_id\_1, user\_id\_2), separated by newlines.}

\questionblock{5}{In the above data, list all pairs of user IDs (no duplicate pairs, list lower ID first) where both users have at least one instance with an entity or numeric value, and all instances that are an entity for both users must be before March 15, 2023. Each of the questions can be labelled as one of the labels (the data does not provide the labels, you need to figure out the label from the semantics of the question): description and abstract concept, entity, human being, numeric value, location, abbreviation. In your answer, list all pairs in the format (user\_id\_1, user\_id\_2), separated by newlines.}

\questionblock{6}{In the above data, list all pairs of user IDs (no duplicate pairs, list lower ID first) where both users have at least one instance with a location or abbreviation. Each of the questions can be labelled as one of the labels (the data does not provide the labels, you need to figure out the label from the semantics of the question): description and abstract concept, entity, human being, numeric value, location, abbreviation. In your answer, list all pairs in the format (user\_id\_1, user\_id\_2), separated by newlines.}

\questionblock{7}{In the above data, list all pairs of user IDs (no duplicate pairs, list lower ID first) where both users have at least one instance with a description and abstract concept or numeric value, and all instances that are a numeric value for both users must be after February 1, 2023. Each of the questions can be labelled as one of the labels (the data does not provide the labels, you need to figure out the label from the semantics of the question): description and abstract concept, entity, human being, numeric value, location, abbreviation. In your answer, list all pairs in the format (user\_id\_1, user\_id\_2), separated by newlines.}

\questionblock{8}{In the above data, list all pairs of user IDs (no duplicate pairs, list lower ID first) where both users have at least one instance with a human being or description and abstract concept. Each of the questions can be labelled as one of the labels (the data does not provide the labels, you need to figure out the label from the semantics of the question): description and abstract concept, entity, human being, numeric value, location, abbreviation. In your answer, list all pairs in the format (user\_id\_1, user\_id\_2), separated by newlines.}

\questionblock{9}{In the above data, list all pairs of user IDs (no duplicate pairs, list lower ID first) where both users have at least one instance with an entity or location, and all instances that are a location for both users must be after April 10, 2023. Each of the questions can be labelled as one of the labels (the data does not provide the labels, you need to figure out the label from the semantics of the question): description and abstract concept, entity, human being, numeric value, location, abbreviation. In your answer, list all pairs in the format (user\_id\_1, user\_id\_2), separated by newlines.}

\questionblock{10}{In the above data, list all pairs of user IDs (no duplicate pairs, list lower ID first) where both users have at least one instance with a numeric value or abbreviation, and all instances that are an abbreviation for both users must be before May 20, 2023. Each of the questions can be labelled as one of the labels (the data does not provide the labels, you need to figure out the label from the semantics of the question): description and abstract concept, entity, human being, numeric value, location, abbreviation. In your answer, list all pairs in the format (user\_id\_1, user\_id\_2), separated by newlines.}

\questionblock{11}{In the above data, list all pairs of user IDs (no duplicate pairs, list lower ID first) such that one user has at least one instance with entity and one with abbreviation, and the other user has exactly one instance with entity. Each of the questions can be labelled as one of the labels (the data does not provide the labels, you need to figure out the label from the semantics of the question): description and abstract concept, entity, human being, numeric value, location, abbreviation. In your answer, list all pairs in the format (user\_id\_1, user\_id\_2), separated by newlines.}

\questionblock{12}{In the above data, list all pairs of user IDs (no duplicate pairs, list lower ID first) such that one user has at least two instances with numeric value, and the other user has at least one instance with location and at least one instance with human being. Each of the questions can be labelled as one of the labels (the data does not provide the labels, you need to figure out the label from the semantics of the question): description and abstract concept, entity, human being, numeric value, location, abbreviation. In your answer, list all pairs in the format (user\_id\_1, user\_id\_2), separated by newlines.}

\questionblock{13}{In the above data, list all pairs of user IDs (no duplicate pairs, list lower ID first) such that one user has exactly one instance with description and abstract concept, and the other user has at least one instance with abbreviation and at least one instance with entity. Each of the questions can be labelled as one of the labels (the data does not provide the labels, you need to figure out the label from the semantics of the question): description and abstract concept, entity, human being, numeric value, location, abbreviation. In your answer, list all pairs in the format (user\_id\_1, user\_id\_2), separated by newlines.}

\questionblock{14}{In the above data, list all pairs of user IDs (no duplicate pairs, list lower ID first) such that one user has at least one instance with human being and at least one instance with numeric value, and the other user has exactly two instances with location. Each of the questions can be labelled as one of the labels (the data does not provide the labels, you need to figure out the label from the semantics of the question): description and abstract concept, entity, human being, numeric value, location, abbreviation. In your answer, list all pairs in the format (user\_id\_1, user\_id\_2), separated by newlines.}

\questionblock{15}{In the above data, list all pairs of user IDs (no duplicate pairs, list lower ID first) such that one user has at least one instance with entity, at least one instance with location, and at least one instance with abbreviation, and the other user has exactly one instance with numeric value. Each of the questions can be labelled as one of the labels (the data does not provide the labels, you need to figure out the label from the semantics of the question): description and abstract concept, entity, human being, numeric value, location, abbreviation. In your answer, list all pairs in the format (user\_id\_1, user\_id\_2), separated by newlines.}

\questionblock{16}{In the above data, list all pairs of user IDs (no duplicate pairs, list lower ID first) such that one user has at least one instance with description and abstract concept and at least one instance with human being, and the other user has at least two instances with entity and exactly one instance with abbreviation. Each of the questions can be labelled as one of the labels (the data does not provide the labels, you need to figure out the label from the semantics of the question): description and abstract concept, entity, human being, numeric value, location, abbreviation. In your answer, list all pairs in the format (user\_id\_1, user\_id\_2), separated by newlines.}

\questionblock{17}{In the above data, list all pairs of user IDs (no duplicate pairs, list lower ID first) such that one user has exactly one instance with numeric value, and the other user has at least one instance with location and at least one instance with description and abstract concept. Each of the questions can be labelled as one of the labels (the data does not provide the labels, you need to figure out the label from the semantics of the question): description and abstract concept, entity, human being, numeric value, location, abbreviation. In your answer, list all pairs in the format (user\_id\_1, user\_id\_2), separated by newlines.}

\questionblock{18}{In the above data, list all pairs of user IDs (no duplicate pairs, list lower ID first) such that one user has at least one instance with abbreviation and exactly one instance with human being, and the other user has at least one instance with entity and at least one instance with numeric value. Each of the questions can be labelled as one of the labels (the data does not provide the labels, you need to figure out the label from the semantics of the question): description and abstract concept, entity, human being, numeric value, location, abbreviation. In your answer, list all pairs in the format (user\_id\_1, user\_id\_2), separated by newlines.}

\questionblock{19}{In the above data, list all pairs of user IDs (no duplicate pairs, list lower ID first) such that one user has at least two instances with location and at least one instance with entity, and the other user has exactly one instance with description and abstract concept and exactly one instance with abbreviation. Each of the questions can be labelled as one of the labels (the data does not provide the labels, you need to figure out the label from the semantics of the question): description and abstract concept, entity, human being, numeric value, location, abbreviation. In your answer, list all pairs in the format (user\_id\_1, user\_id\_2), separated by newlines.}

\questionblock{20}{In the above data, list all pairs of user IDs (no duplicate pairs, list lower ID first) such that one user has at least one instance with numeric value and at least one instance with human being, and the other user has at least one instance with location, at least one instance with entity, and exactly one instance with abbreviation. Each of the questions can be labelled as one of the labels (the data does not provide the labels, you need to figure out the label from the semantics of the question): description and abstract concept, entity, human being, numeric value, location, abbreviation. In your answer, list all pairs in the format (user\_id\_1, user\_id\_2), separated by newlines.}

\subsection{Scaling Huge Document Corpora in BrowseComp+} \label{appx:bc+}
In addition to the BrowseComp+~\citep{chen2025browsecompplusfairtransparentevaluation} results for $k=1000$ documents in~\S\ref{sec4.3-results}, we also include a smaller set of results on a subset of $20$ tasks from the original $150$ to show how performance degrades as a function of input size. In our original experiments, the base LMs were unable to handle the input contexts, so we add results to show how they degrade. We include two new baselines, namely \textbf{ReAct w/ GPT-5 + BM25} (a variant of the CodeAct baseline without access to a code environment) and \textbf{GPT-5 + pre-query BM25} (GPT-5 on pre-queried documents). 
\begin{figure}[htb!]
    \centering
    \includegraphics[width=\textwidth]{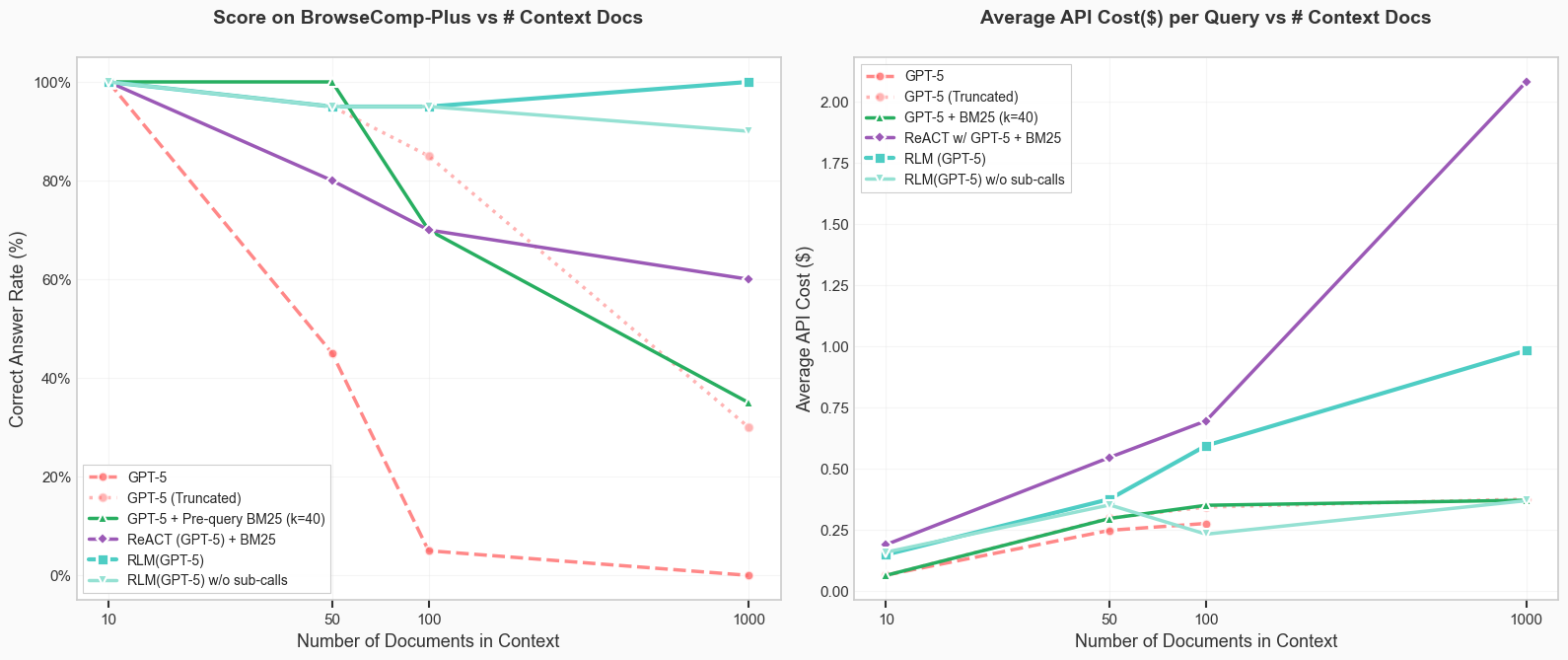}
    \caption{We plot the performance and API cost per answer of various methods using GPT-5 on 20 random queries in BrowseComp-Plus given increasing numbers of documents in context. Only the iterative methods (RLM, ReAct) maintain reasonable performance at 100+ documents.}
    \label{fig:bc-scaling}
\end{figure}

\textbf{\RLM{}s are able to scale well without performance degradation.} RLM(GPT-5) is the only model / agent able to achieve and maintain perfect performance at the 1000 document scale, with the ablation (no recursion) able to similarly achieve $90\%$ performance. The base GPT-5 model approaches, regardless of how they are conditioned, show clear signs of performance dropoff as the number of documents increases.

\textbf{\RLM{} inference cost scales reasonably.} The inference cost of \RLM{}s on this setup scale log-linearly, and are reasonably bounded compared to other common strategies like ReAct + BM25. If we extrapolate the overall token costs of GPT-5 assuming it has an infinite context window, we observe that the inference cost of using \RLM{}(GPT-5) is cheaper.

\pagebreak
\section{Additional \RLM{} Trajectories} \label{appx2:examples}
In this section, we provide several example trajectories to highlight characteristics of frontier models as \RLM{}s. Many of the trajectories are too long to fit in text, so we describe each step and show specific examples when relevant.

\begin{figure*}[htb!]
    \centering
    \includegraphics[width=0.98\textwidth]{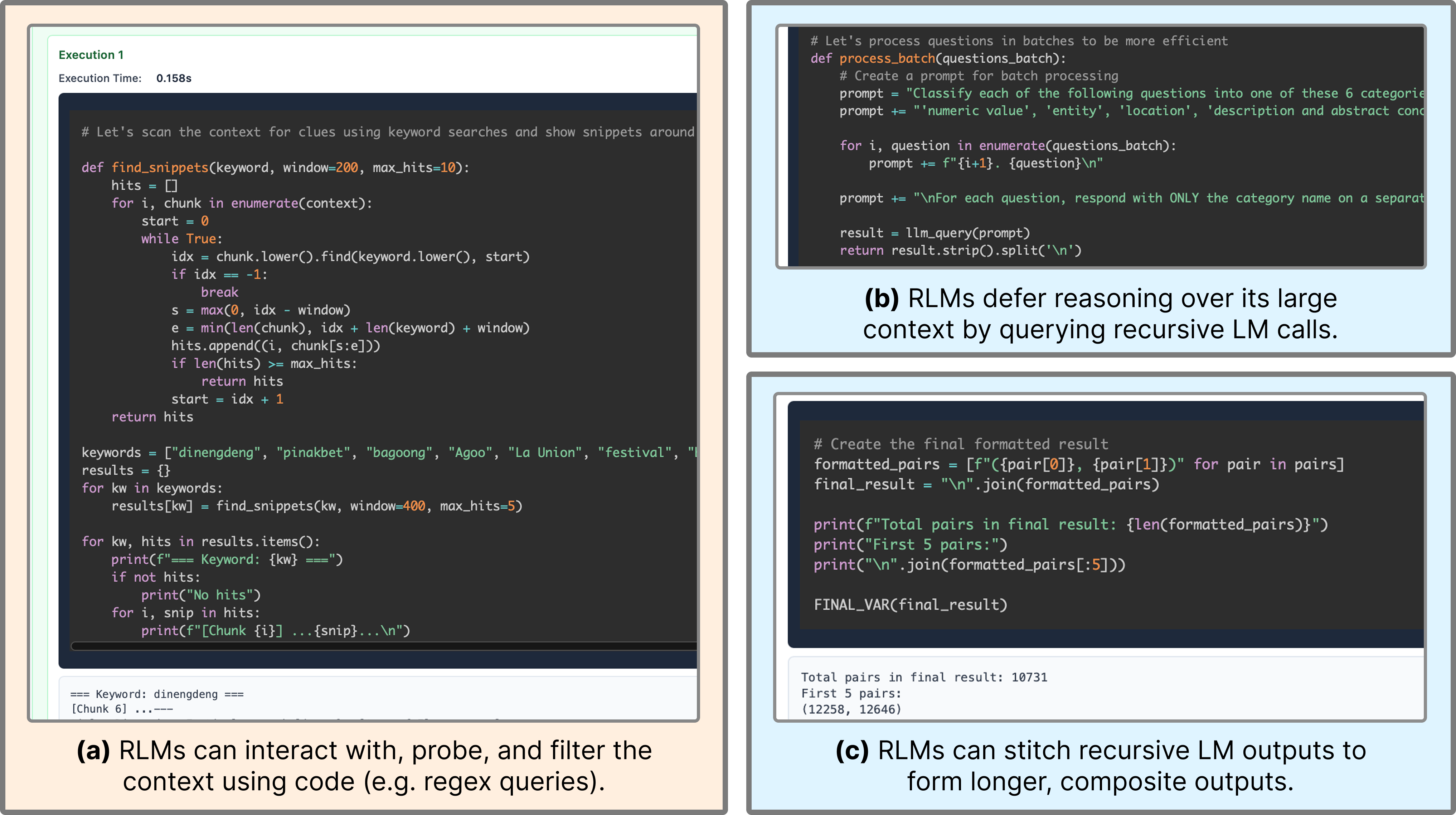}
    \caption{\RLM{}s have common patterns in their trajectories when solving tasks. (a) We frequently observed \RLM{}s filtering and interacting with their context through \texttt{regex} code. (b) We found that \RLM{}s can effectively decompose their context through recursive sub-calls (c) On long-output tasks, \RLM{}s are able to solve sub-problems using recursive sub-LM calls and stitch their outputs to form a final output.}
    \label{fig:trajectories}
\end{figure*}

A few noticeable properties of these trajectories are that \RLM{}s often make non-optimal choices despite their strong results in \S\ref{sec4:long-input}. For example, in Example~\ref{ex:op_3}, we observed that the \RLM{} with Qwen3-Coder carefully constructs its final answer through a mix of recursive sub-calls and code execution in the first iteration, but then discards this information and continues wasting sub-calls before not using these stored answers. We also observed distinct differences in model behavior such as in Example~\ref{ex:o_212}, where we found Qwen3-Coder make hundreds to thousands of recursive sub-calls for a single simple task, while GPT-5 makes on the order of ten. While these examples are not comprehensive, they provide useful qualitative insight into how to improve \RLM{}s.

\subsection{RLM(GPT-5) on BrowseComp-Plus-Query\_74} \label{ex:bcp_74}

The total cost of this trajectory was \textbf{\$0.079}. In this task, the agent must find the answer to the following multi-hop query given a corpus of 1000 unique documents (~8.3M total tokens) that contain evidence documents and negatives:

\begin{lstlisting}[style=customstyle]
This vegetable stew uses fish, but adding meat is possible. It also uses a salty and intense condiment, which is the critical ingredient of the dish. As of 2023, a township holds a celebration named after this stew. Between 1995 and 2005 inclusive, this festivity began after authorities shifted the highlight and subject of their event to set them apart from other areas in the region that use the same product in their celebrations. This town holds the event every year after February but before September. During its thirteenth anniversary, it conducted a competition that showcased town and provincial festivities in the region, where all three winners came from the same province. A beauty pageant was also a part of the celebration. What are the first and last names of the person who won that contest that year?
\end{lstlisting}

\textbf{Step 1.} GPT-5 (as the root LM) first decides to probe at the 1000 document list with regex queries. It has some priors about these events (as shown from its particular choice of words it looks for), but it also looks for specific keywords in the prompt like ``beauty pageant'' and ``festival''.

\includegraphics[width=\textwidth]{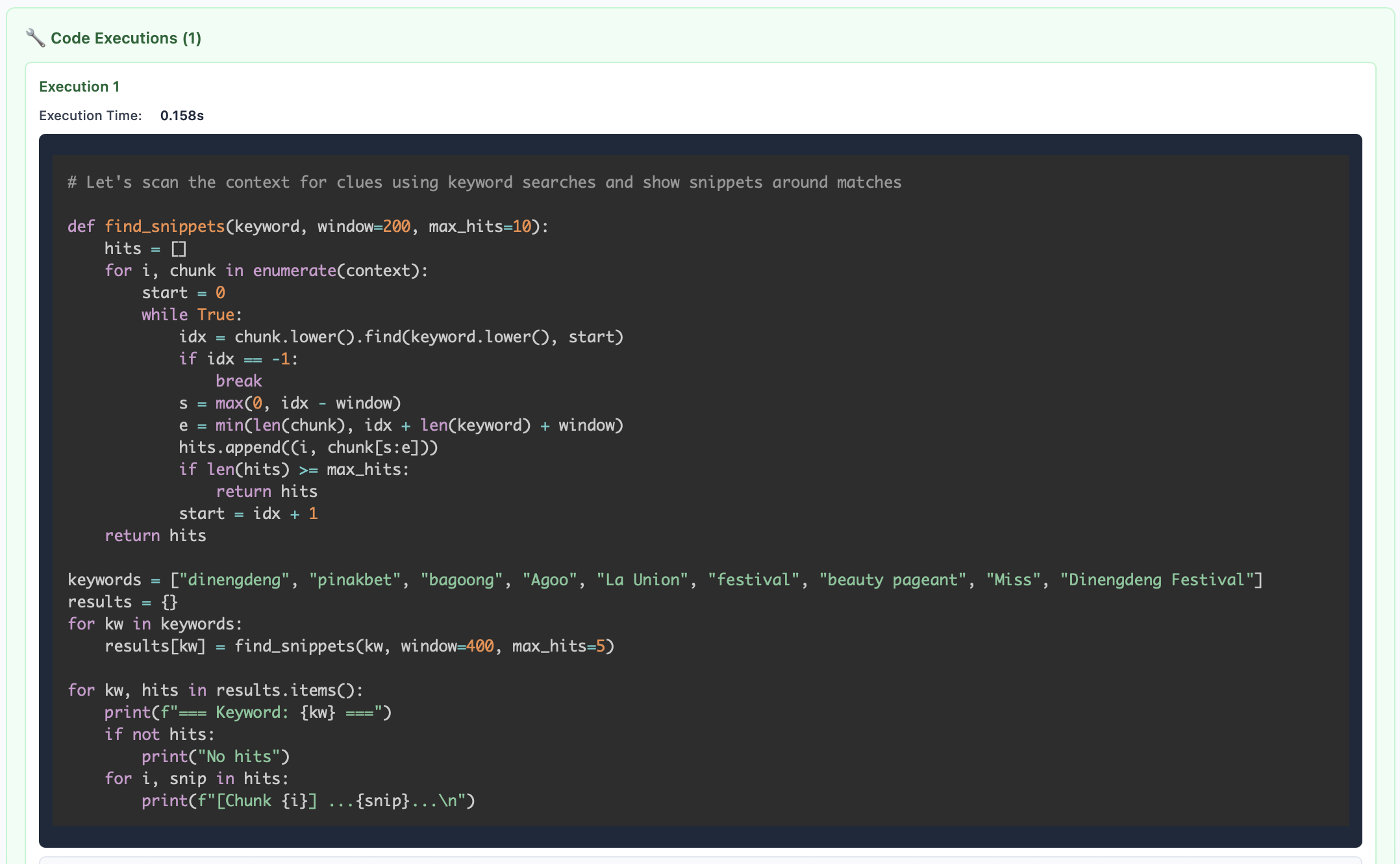}

\textbf{Step 2.} After running its regex queries, the root LM finds an interesting snippet on the chunk at index 6, so it launches a recursive LM call over this snippet to look for information relevant to the original query. The \RLM{} is able to both store this information in a variable \texttt{answer6}, as well as print this information out for the root LM to see. The sub-LM call finds the answer is likely `Maria Dalmacio` and stores this information back in the root LM's environment.

\includegraphics[width=\textwidth]{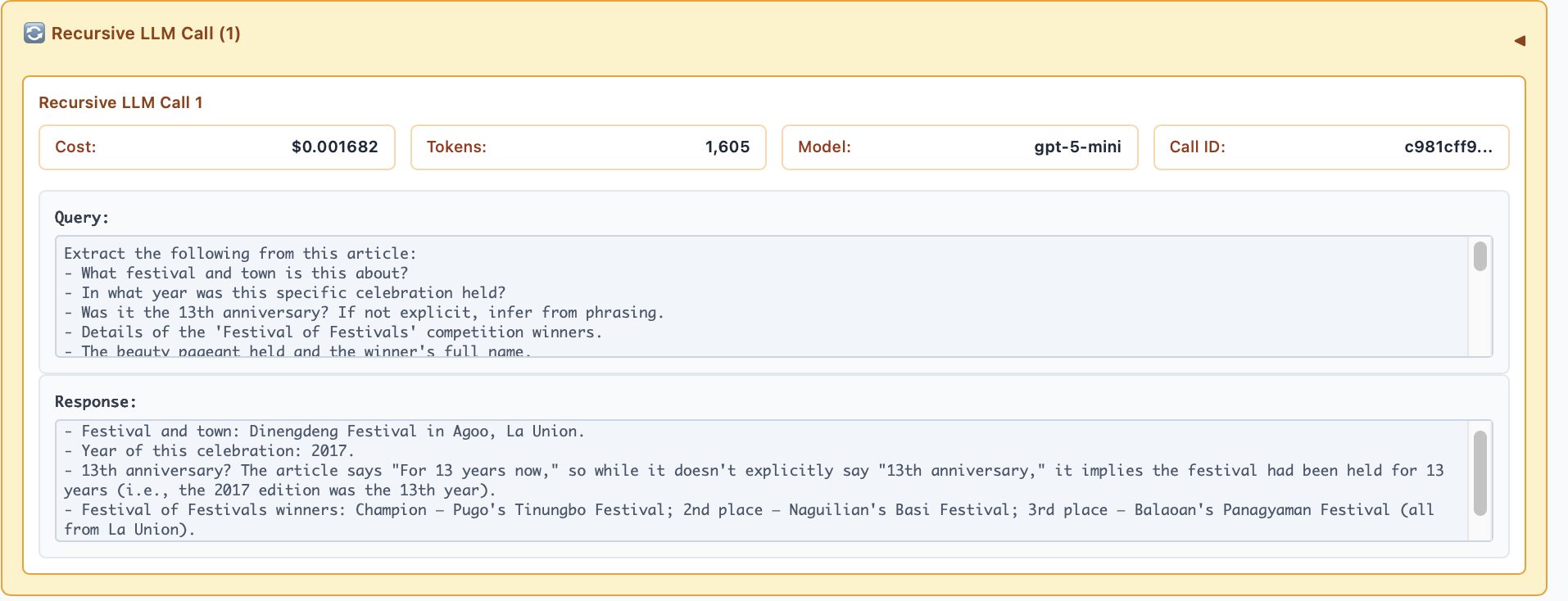}

\includegraphics[width=\textwidth]{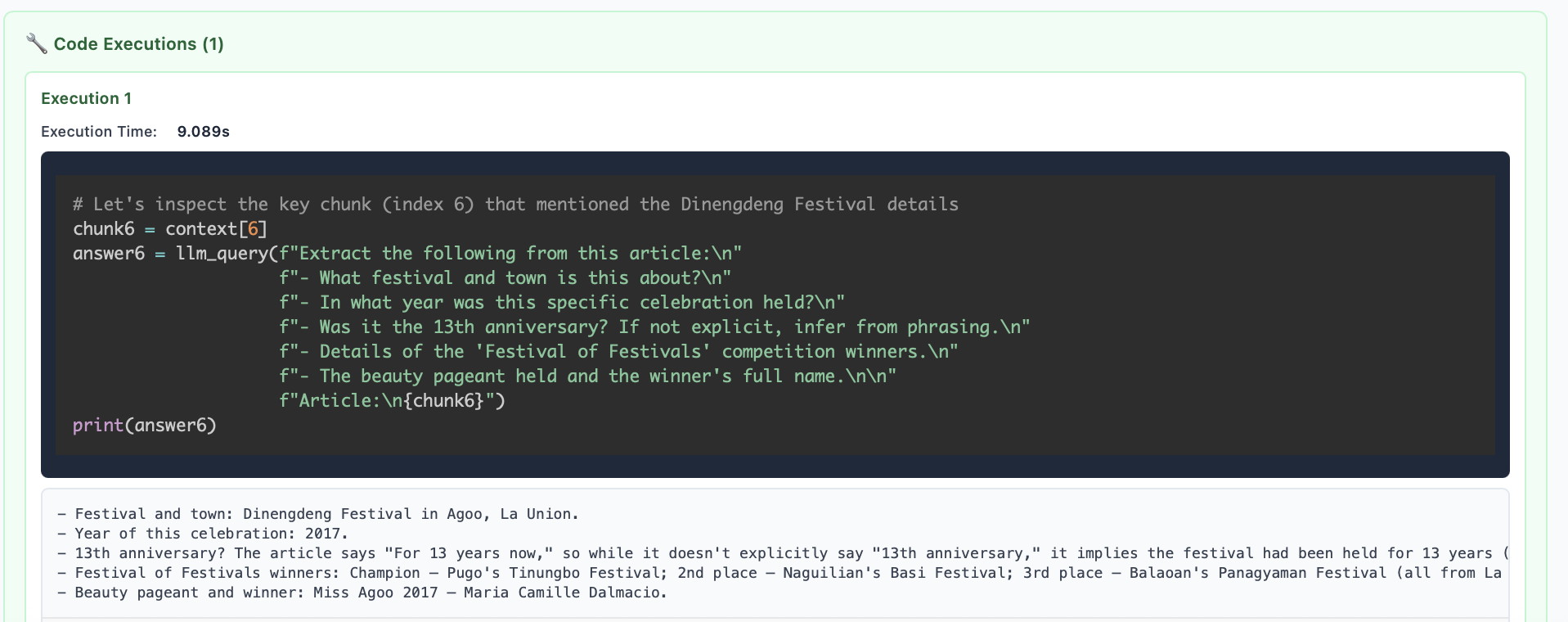}

\textbf{Step 3.} After checking the information above, the root LM reasons that it has enough information to answer the query. The root LM chooses to check its answer again with two additional recursive LM calls to confirm that its answer aligns with this check. Finally, the root LM returns its final answer as `Maria Dalmacio`, which is the correct answer.

\includegraphics[width=\textwidth]{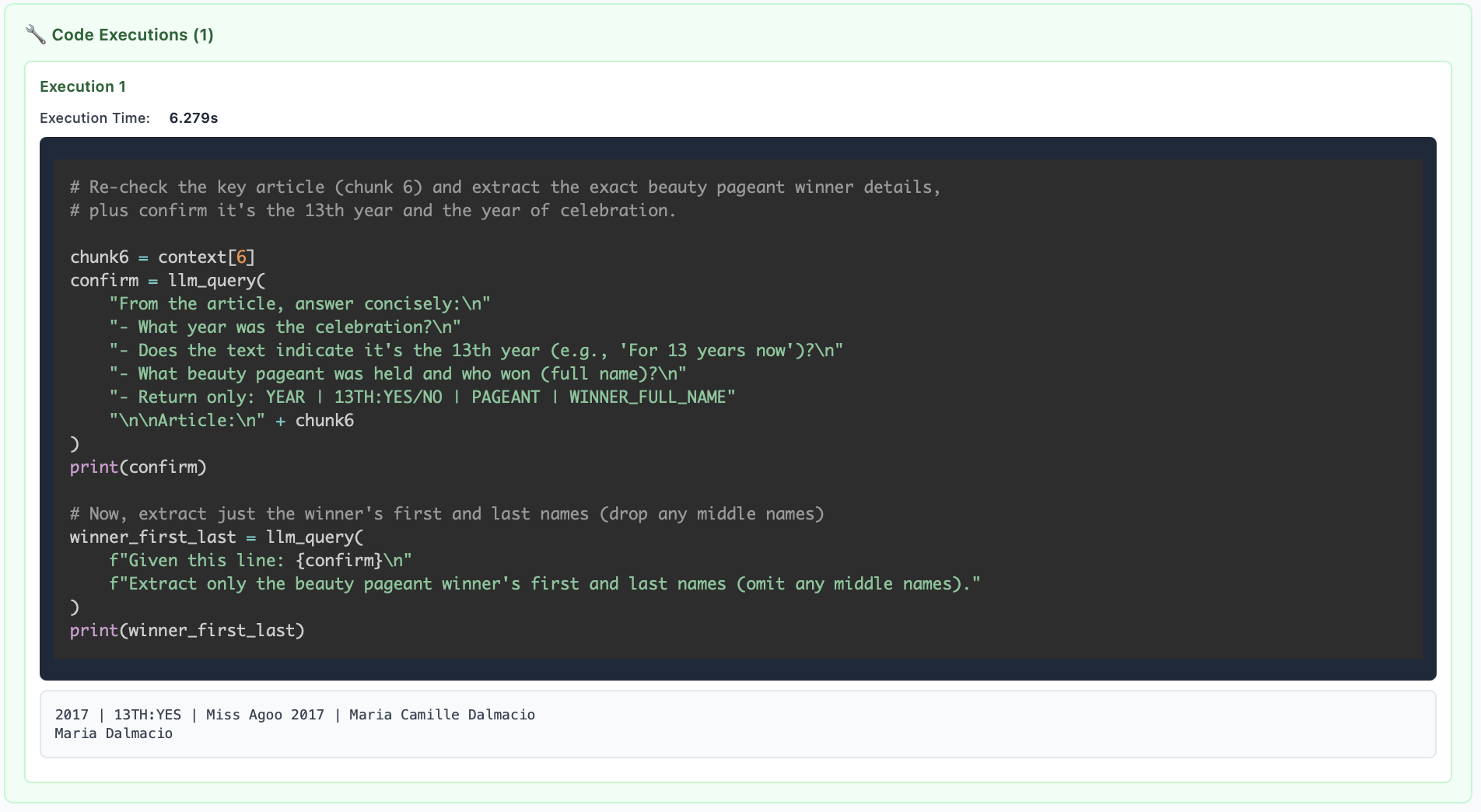}

\subsection{RLM(Qwen3-Coder) on OOLONG-Pairs-Query\_3} \label{ex:op_3}
The total cost of this trajectory was \textbf{\$1.12}. In this task, the agent must output all pairs of user IDs satisfying some set of properties given a list of entries (~32k tokens total). This is both an information dense long input as well as long output task, making it particularly challenging for current LMs.

\begin{lstlisting}[style=customstyle]
Answer the following: In the above data, list all pairs of user IDs (no duplicate pairs, list lower ID first) where both users have at least one instance with a description and abstract concept or abbreviation. Each of the questions can be labelled as one of the labels (the data does not provide the labels, you need to figure out the label from the semantics of the question): description and abstract concept, entity, human being, numeric value, location, abbreviation. In your answer, list all pairs in the format (user_id_1, user_id_2), separated by newlines. Your answer must be sorted by first user ID. For example, if the answer is the Instance ID pairs (22740, 35839) and (35839, 52032), you should return `(22740, 35839), (35839, 52032)`. If there is no answer, return an empty list [].
\end{lstlisting}
\textbf{Step 1.} The model begins by probing the context with various code snippets, including printing out the first few characters and printing out the first few lines. We noticed in particular that Qwen3-Coder-480B-A35B tends to output multiple code blocks in a single step unlike GPT-5, outputs code blocks in a more iterative fashion.

\includegraphics[width=\textwidth]{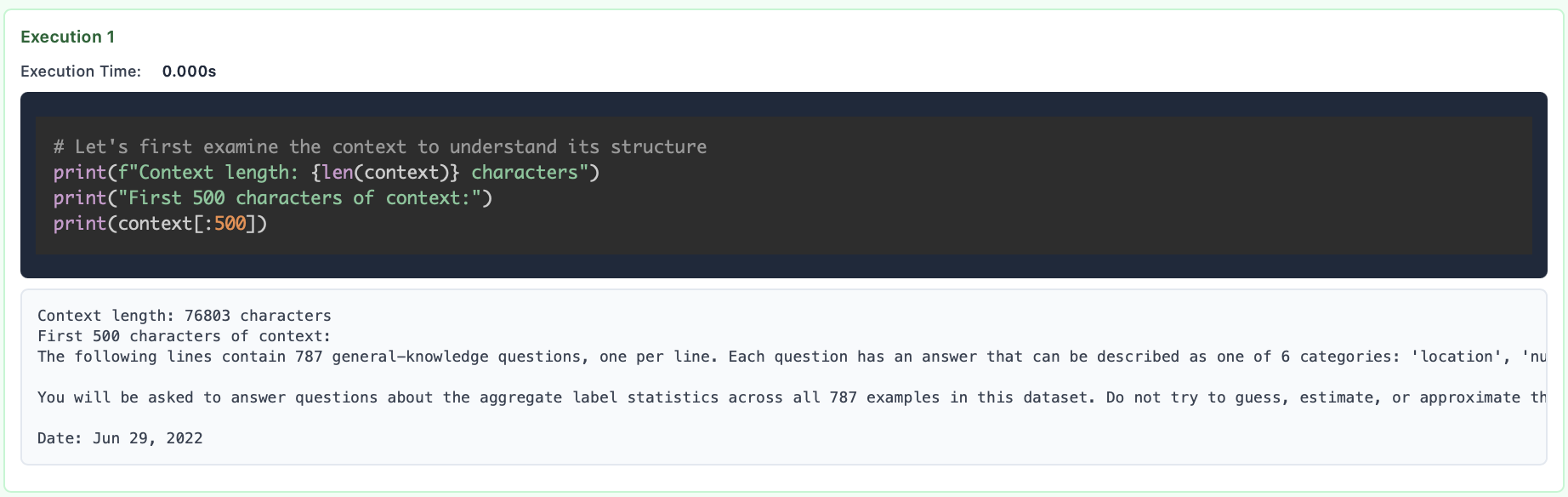}

The model continues probing by splitting the input context by newline characters and checking roughly what the data format looks like.

\includegraphics[width=\textwidth]{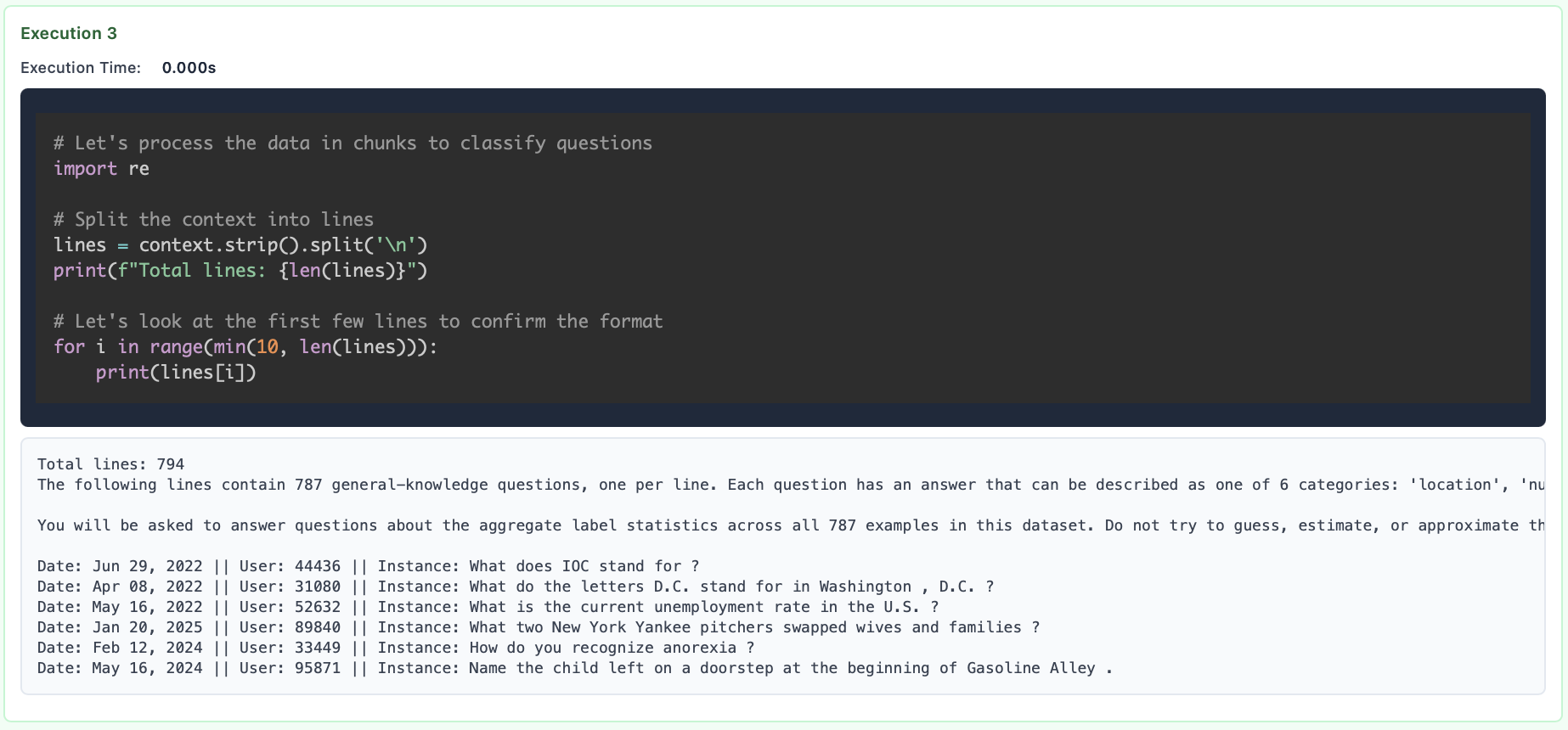}

From the given format, the model chooses to first semantically classify the data using sub-LM calls over smaller chunks of the input (to avoid context rot and mistakes in larger contexts) and provides a sample back to the root LM of what it observed during this process.

\includegraphics[width=\textwidth]{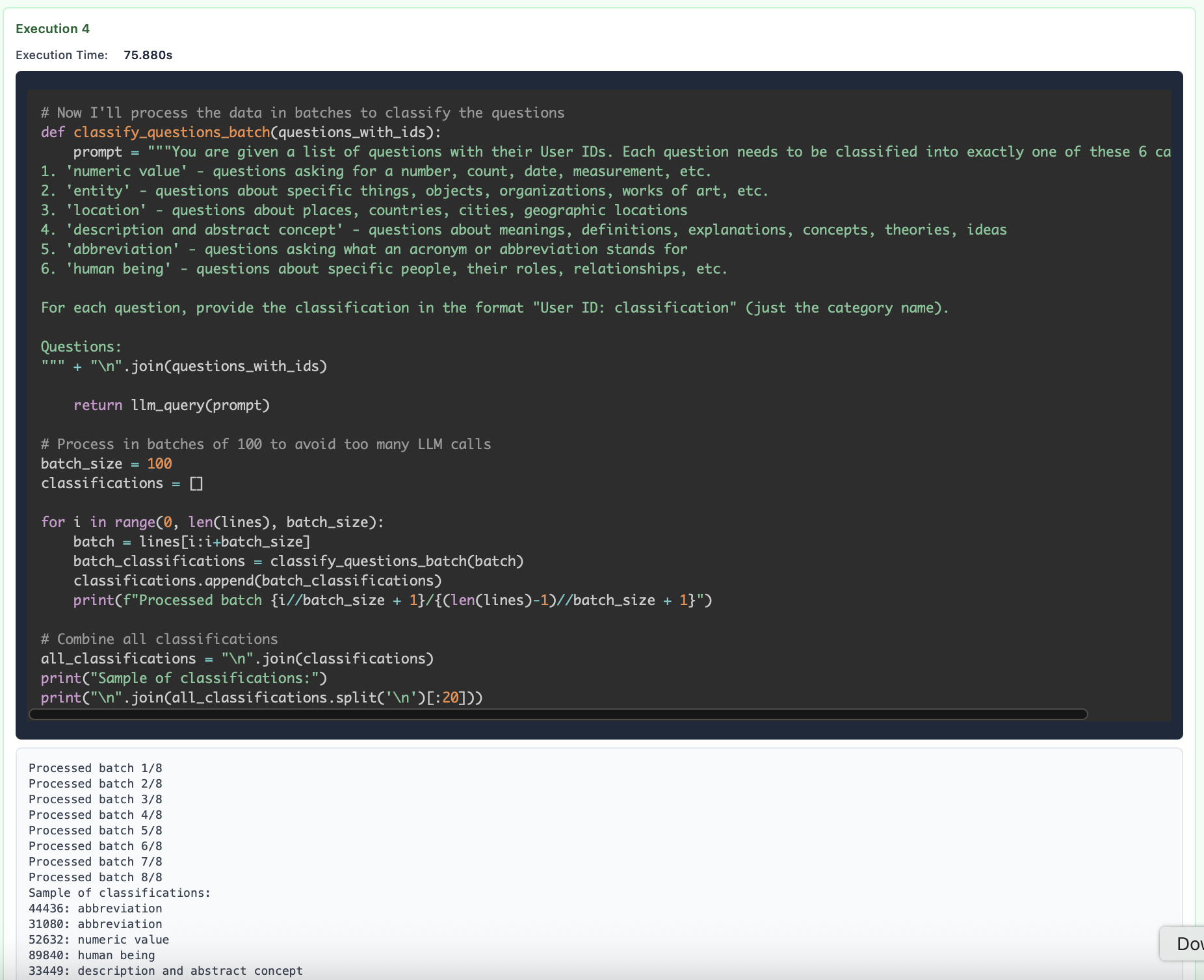}

Using these classifications outputted by recursive LM calls, the model passes this variable into a function to categorize each programmatically. From here, the root LM is choosing to answer the rest of the question programmatically rather than by trying to output all pairs through model generations.

\includegraphics[width=\textwidth]{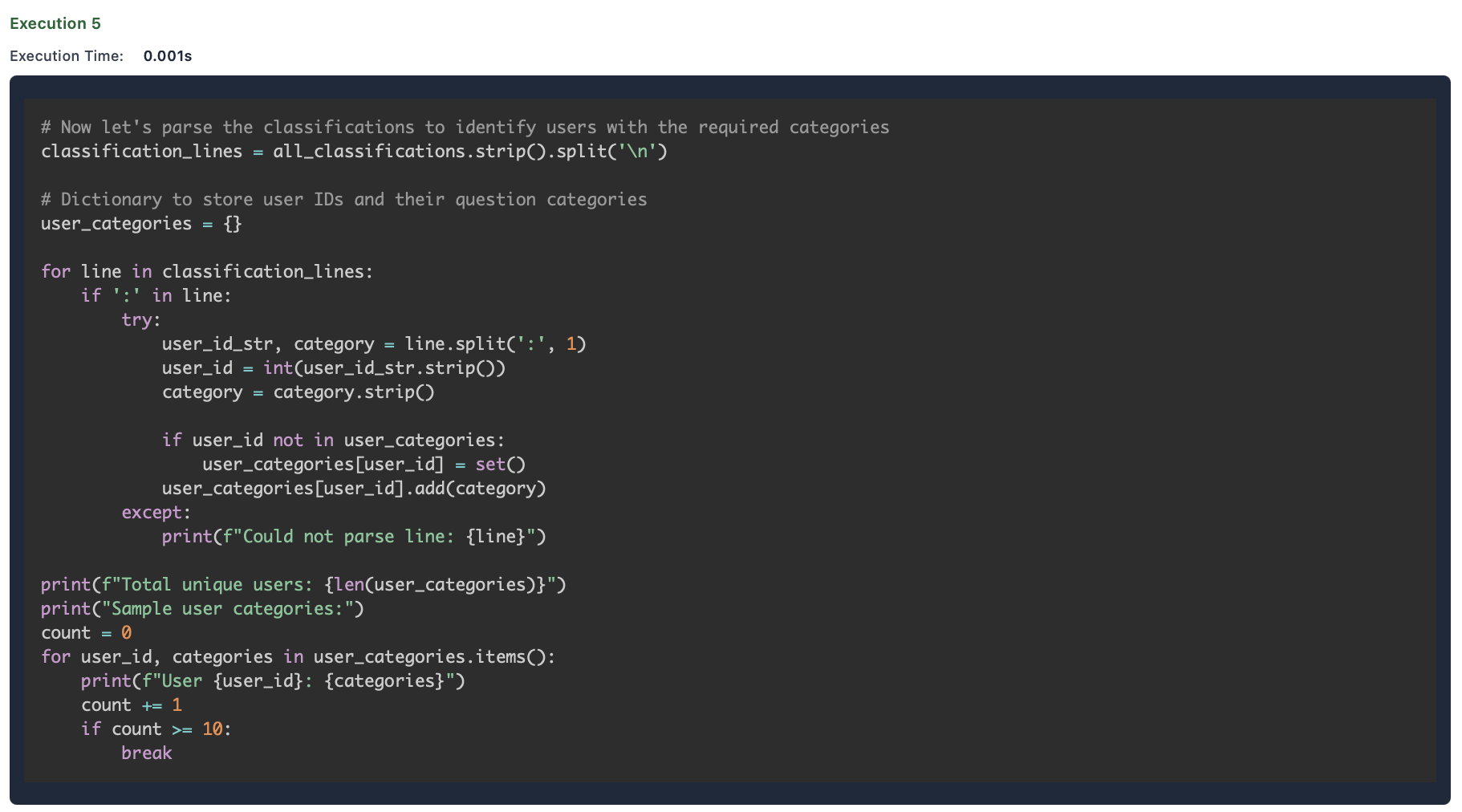}

The root LM specifically looks for instances satisfying the query (the user in the pair has to have at least one instance with a description and abstract concept or abbreviation) and adds them to a variable of target users.

\includegraphics[width=\textwidth]{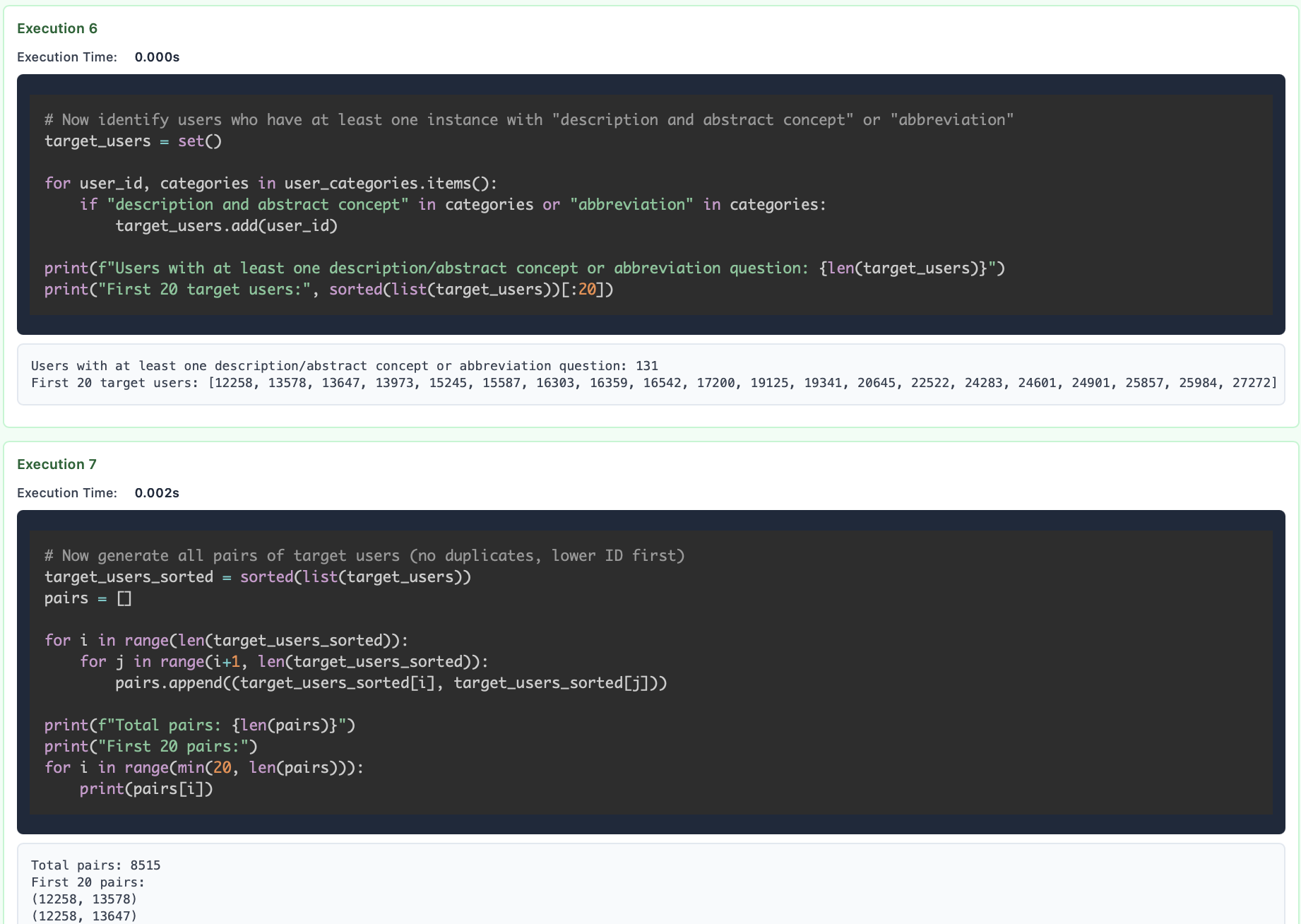}

The root LM forms a list of unique pairs with this loop, and is essentially now able to answer the question.

\includegraphics[width=\textwidth]{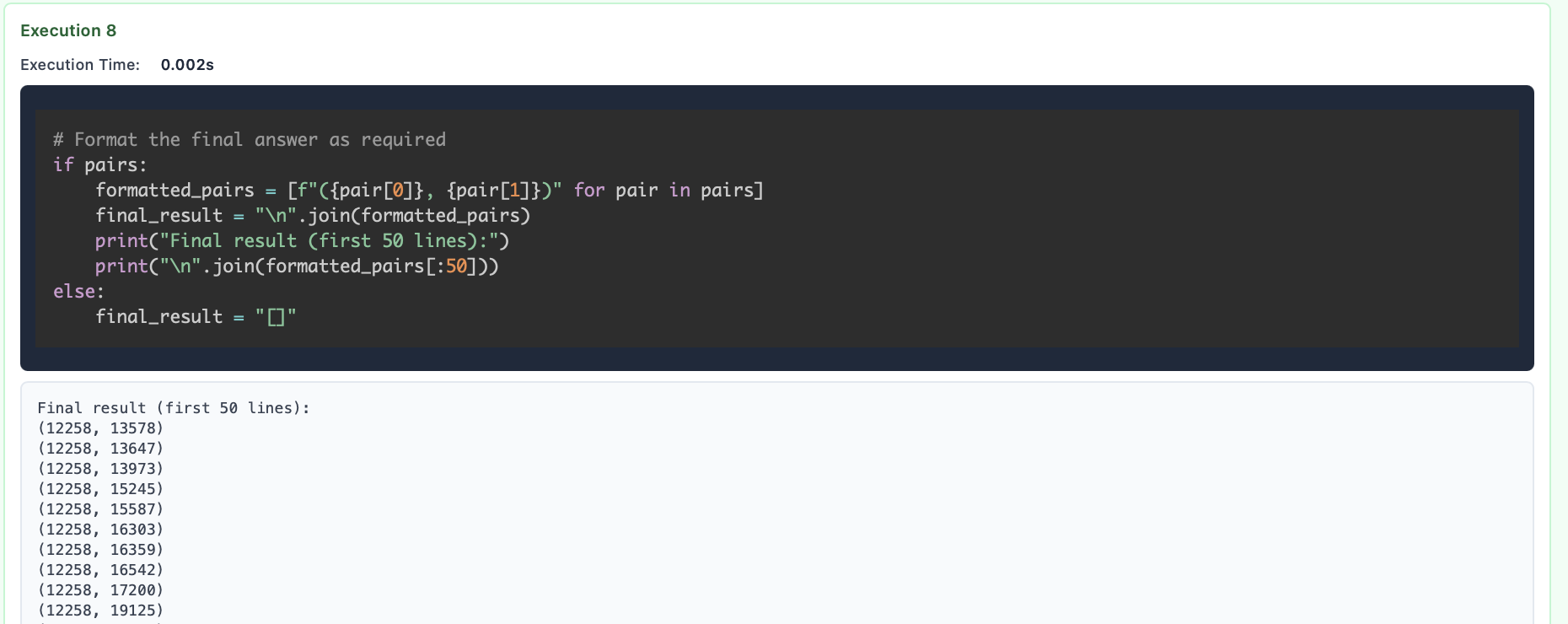}
The model has stored these pairs in a variable to be outputted at the end. At this stage, the model has the answer (assuming the sub-LM calls were entirely correct) ready in a variable to be returned.

\textbf{Step 2.} By this point the model has already successfully extracted the answer. Interestingly, however, as we observed frequently with Qwen3-Coder, the model will continue to repeatedly verify its answers. The model also attempts to return its answer wrapped in a `FINAL\_VAR()` tag, but it does not accept its answer. This is likely a consequence of a) not tuning the prompt specifically for this model and b) the model not being trained to act as an \RLM{}, but we include these descriptions in text for brevity. At this step, the model checks its pairs.

\textbf{Step 3.} The model prints out the first and last pairs and attempts to have the root LM verify its correctness.

\textbf{Step 4.} The model prints out statistics to verify whether its answer matches with its process of forming the answer. 

\textbf{Step 5.} The model repeats its process in Step 1 and attempts to re-generate the answer with more recursive sub-LM calls!

\textbf{Step 6 - 11.} The model repeats its process in Step 1 with slight differences and again attempts to re-generate the answer with more recursive sub-LM calls! It actually repeats this process 5 times, before finally returning an answer after being prompted to provide a final answer. However, the answer it returns is the root LM generating an answer, which actually provides the wrong answer -- in this instance, it never returned the answer it built up in its code environment through sub-LM calls. This is an example of a case where the \RLM{} failed.

\subsection{RLM(Qwen3-Coder) on OOLONG-Query\_212} \label{ex:o_212}
The total cost of this trajectory was \textbf{\$0.38}. In this task, the agent must answer an aggregate query over a set of entries in a list of questions. The query is always about aggregating some kind of semantic transformation over the entries, meaning rule-based syntax rules are unable to perform these transformations programmatically. In this example, the \RLM{} is answering the following question:

\begin{lstlisting}[style=customstyle]
The following lines contain thousands of general-knowledge questions, one per line. Each line has a User ID, which is not necessarily unique, i.e. each User ID can be associated with multiple questions. Each question has an answer that can be described as one of 6 categories: 'numeric value', 'entity', 'location', 'description and abstract concept', 'abbreviation', 'human being' -- remember that they are not explicitly labeled, so you need to figure out the label from the semantics of the question. You will be asked to answer questions about the aggregate label statistics across all examples in this dataset. Do not try to guess, estimate, or approximate the result. Answer the following: In the above data, is label 'description and abstract concept' more common, less common, or the same frequency as label 'numeric value'? Give your final answer in the form 'Answer: description and abstract concept is [X] numeric value', where [X] is 'more common than', 'less common than', or 'same frequency as'.
\end{lstlisting}

\textbf{Step 1.} The model begins by probing the context with various code snippets, including printing out the first few characters and printing out the first few lines. Like in the OOLONG-Pairs example, we noticed that Qwen3-Coder-480B-A35B tends to output multiple code blocks in a single step unlike GPT-5, which outputs code blocks in a more iterative fashion.

\includegraphics[width=\textwidth]{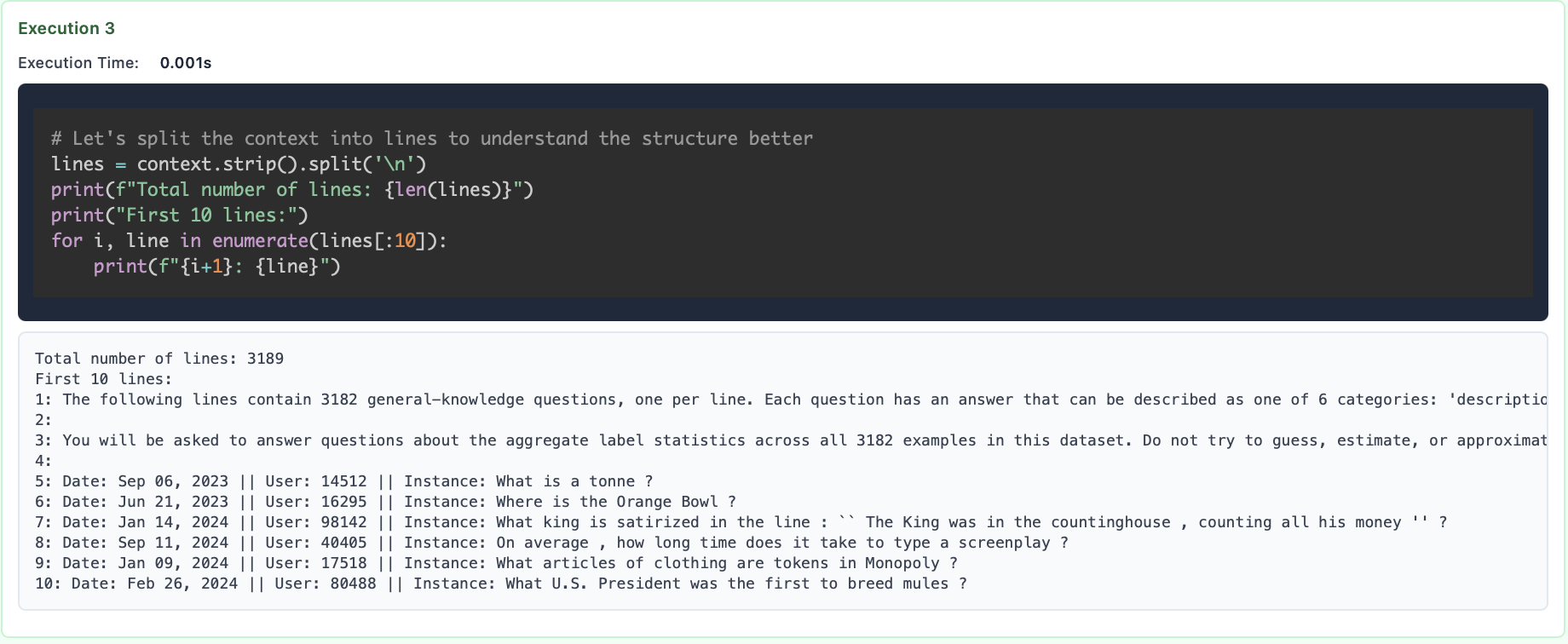}

As mentioned previously, Qwen3-Coder differs from GPT-5 in how liberal it is in its use of sub-calls. The function Qwen3-Coder defines for classifying entries semantically uses a sub-LM call \textit{per line}, leading to thousands of recursive sub-calls when applied to the full input context.

\includegraphics[width=\textwidth, trim=0 400 0 0, clip]{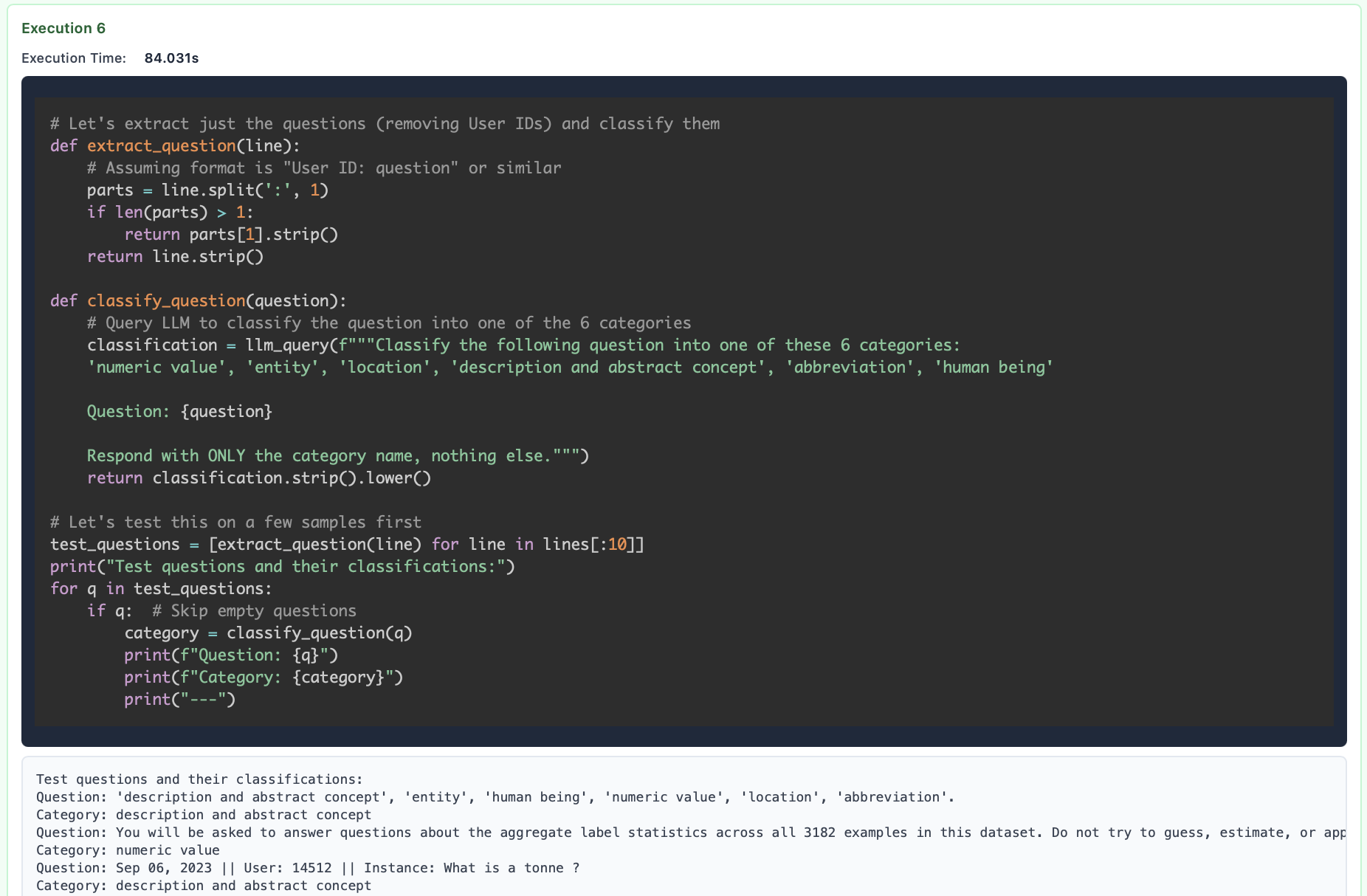}

\includegraphics[width=\textwidth, trim=0 0 0 200, clip]{trajectories/o-212_2.png}

\textbf{Step 2.} After defining and testing several functions for running the above classification question over its input context, the root LM launches a long code execution call to classify and answer the query.

\includegraphics[width=\textwidth]{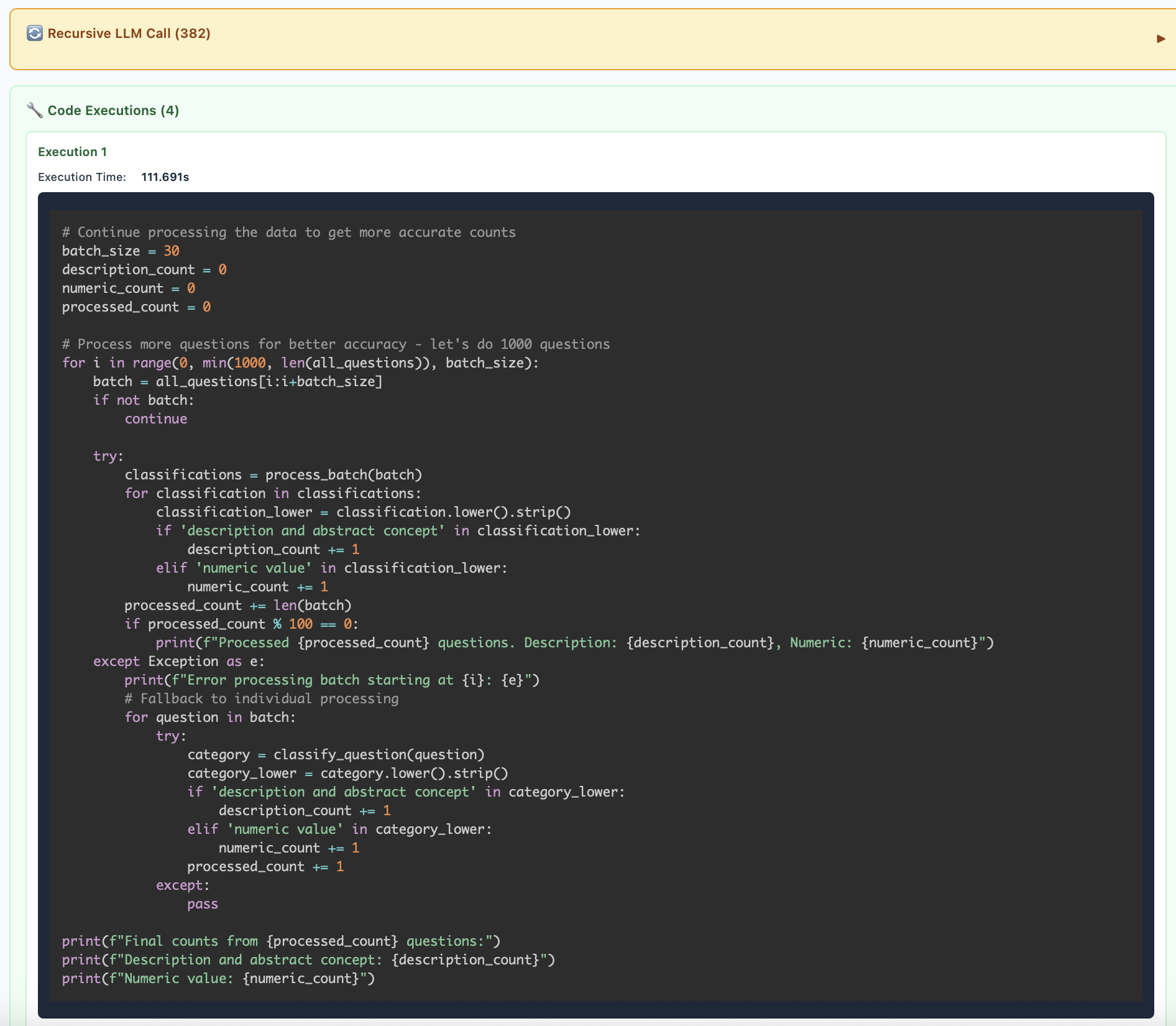}

\textbf{Final.} The model concludes programmatically from the large number of sub-calls it performed in Step 2 that `Answer: description and abstract concept is less common than numeric value` was the correct answer. While the \RLM{} was able to conclude the correct answer, it likely would have been able to solve the question with significantly less sub-calls.

\subsection{RLM(GPT-5) on CodeQA-Query\_44} \label{ex:codeqa_44}
The total cost of this trajectory was \textbf{\$0.27}. In this task, the agent must answer a question that involves understanding a large codebase. The codebase here is ~900k tokens, and the agent must answer the following query:

\begin{lstlisting}[style=customstyle]
You are a helpful assistant that can answer questions about code repositories. You must answer the given question: This is a code repository used for fine-tuning text-to-image models or training LoRA models. The repository is used for the author's research on some related uses. Below are the steps I followed during the process. Could you help me check which one is right statement? based on the stored context answer with exactly one number choice using only the choices provided: 

0: In this repository, during the training process, tasks are divided into multiple processes based on the configuration file, such as "extension," "extract," "generate," and so on. For each process, a corresponding class has been written. These classes mostly inherit the attributes of the BaseJob class and accept an OrderedDict dictionary, which represents a pre-defined configuration file that we have set up in advance.Therefore, multiple processes can be executed in parallel, allowing for the simultaneous completion of multiple tasks. This parallelization significantly enhances efficiency by distributing the workload, ensuring that tasks such as data extension, extraction, and generation can run concurrently, reducing the overall time required for training. 

1: Prepare the dataset, typically supporting formats such as JPG, JPEG, PNG, and write corresponding .txt files to describe the content of the images. Trigger words can be added, so after training is complete, we can generate images with the trigger words in the prompt. In the config directory, find the configuration files and modify the .yml files. Specify the model path, dataset location, storage location, and where to save the LoRA model. Only after configuring these settings can it run properly. 

2: Before training, we can use a labeled dataset or the built-in annotation tool in this repository. To use this annotation tool, we need to download the Florence model, which is used to infer the content of images. Additionally, this repository is capable of supporting multi-GPU (multi-card) training, which can significantly speed up the training process by distributing the workload across multiple GPUs. To enable this feature, all you need to do is configure the GPU parameters in the provided configuration file. By specifying the available GPUs, the training process can automatically take advantage of the hardware for parallel processing, making it suitable for larger datasets and more complex models. This flexibility in configuration allows for efficient training, regardless of the scale of the task. 

3: This project has several ways to run. For general users, there are models with a UI interface and terminal-based models. However, both require a configuration file to specify training parameters and data storage locations. After LoRa training is completed, we can run the run.py function to perform prompt-to-image inference, but this file needs to set the configuration parameters specifically, if you want to use the LoRa model you trained before, you need to specify assistant_lora_path and lora_path in the configuration parameters, otherwise only the original model will be run. (indexed from 0 to 3).
\end{lstlisting}

\textbf{Step 1.} It is not always true that an input context can be solved by partitioning it and recursively sub-querying models over each partition, but in tasks that are not information dense, this is possible. In this case, the model chooses to break down the codebase into parts and sub-query LMs to look for clues. The model then aggregates these clues and provides a final answer as a separate sub-query.

\includegraphics[width=\textwidth, trim=0 200 0 0, clip]{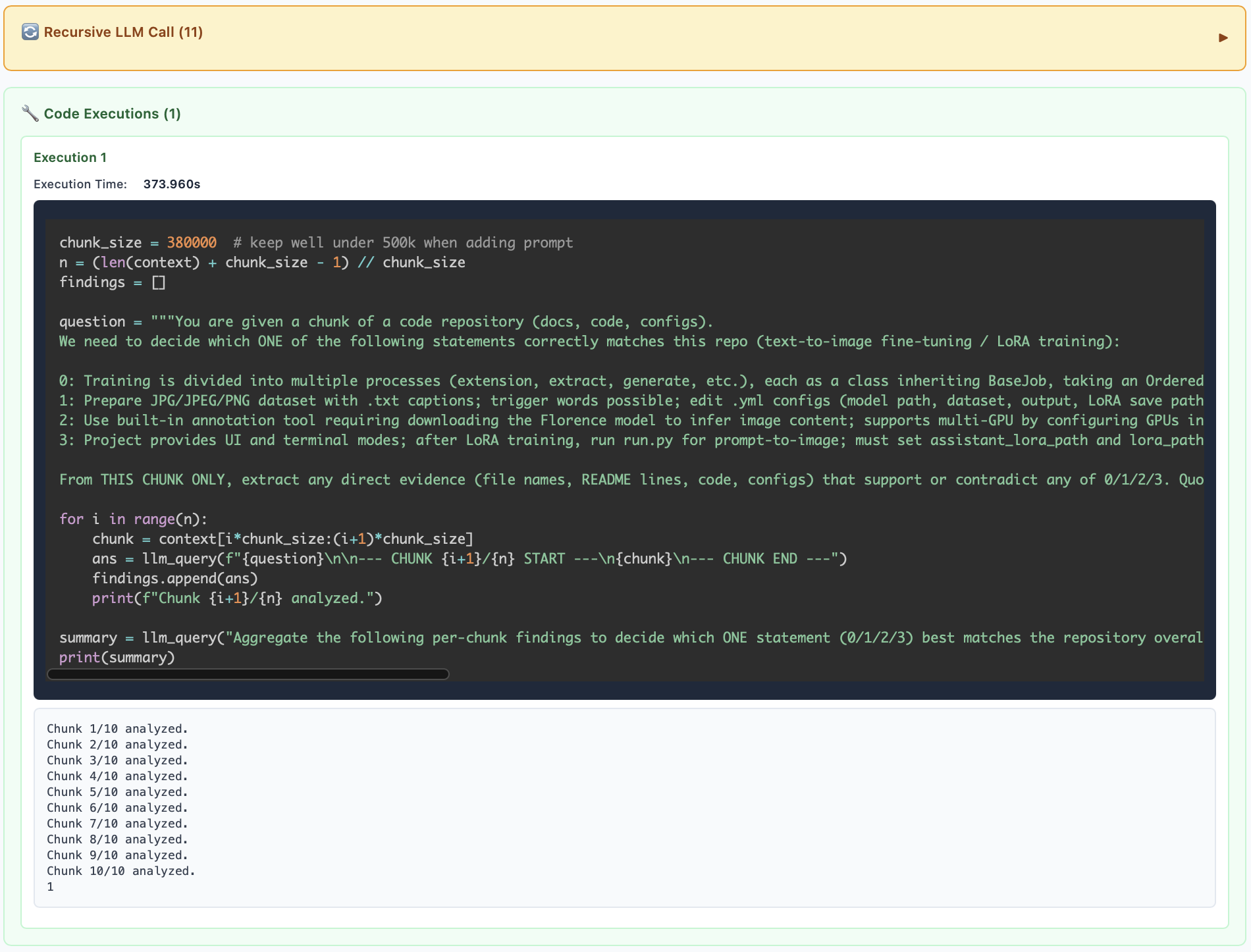}

\textbf{Final.} The \RLM{} answers choice `1', which is the correct answer.

\pagebreak
\section{Additional Quantitative Results} \label{appx1:runtime-cost}
\subsection{Additional Quantitative Analysis of Main Results} \label{appx:sub:quantitative}
\begin{figure}[!hbt]
    \centering
    \includegraphics[width=1\linewidth]{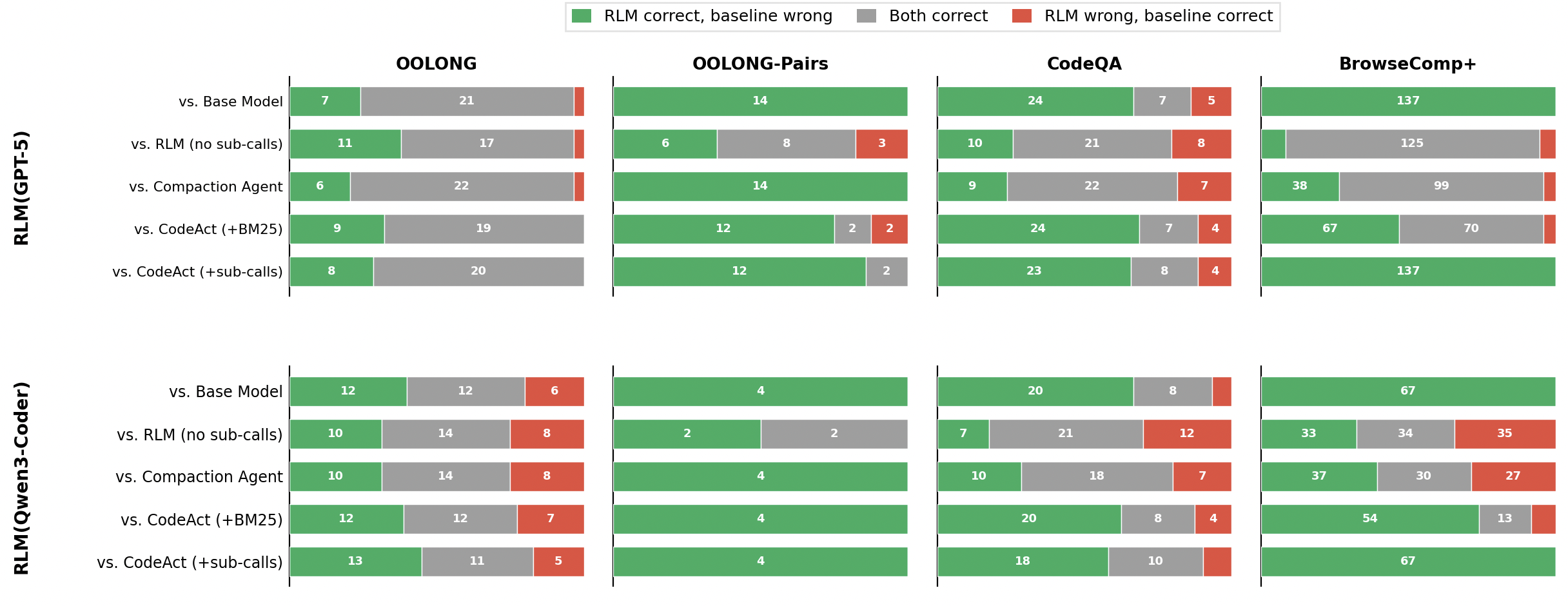}
    \caption{For GPT-5 and Qwen3-Coder-480B-A35B-Instruct, we plot how the \RLM(recursion depth=1) compares to other baselines using the same models. For all tasks where at least one method gets the answer correct, we show how many only the \RLM{} got correct in green, how many both got correct in gray, and how many only the baseline got correct in red.}
    \label{fig:tab1_comparison}
\end{figure}

We supplement Table~\ref{tab:main} with fine-grained rollout success of a few baseline methods compared to the \RLM(recursion depth=1). In Figure~\ref{fig:tab1_comparison}, we generally find \RLM{}s solve the same tasks, and more tasks than, other baselines, especially for GPT-5.

\begin{figure}[!hbt]
    \centering
    \includegraphics[width=1\linewidth]{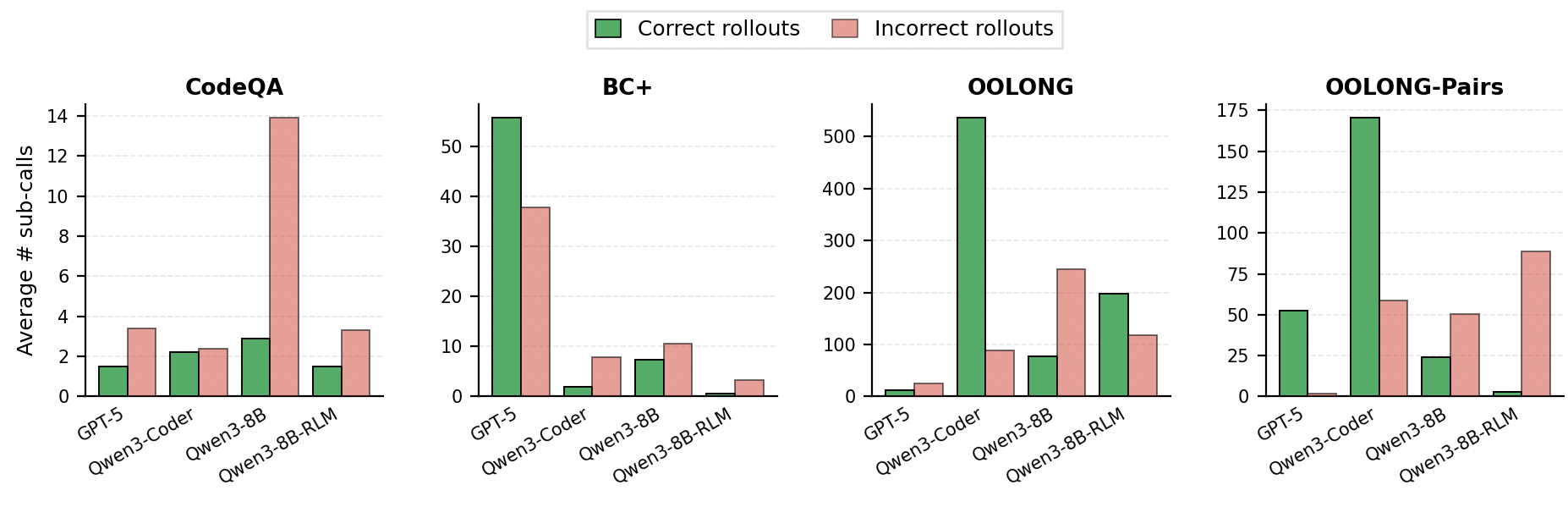}
    \caption{For each task in Table~\ref{tab:main}, we plot the average number of sub-calls made during an \RLM{}(depth=1) trajectory for each task, grouped by whether it got the task correct or incorrect.}
    \label{fig:sub-call-rollouts}
\end{figure}

We also explore how sub-calling behavior differs between rollouts. In Figure~\ref{fig:sub-call-rollouts}, we find a wide range of sub-calling behaviors that greatly differ across models and even for correct and incorrect rollouts. For example, GPT-5 uses significantly more sub-calls for BrowseComp-Plus than any other model. However, for OOLONG, Qwen3-Coder uses a large number of sub-calls (~500 on average) for correct rollouts, which is significantly more than the number used by GPT-5. Furthermore, Qwen3-8B in particular tends to use more sub-calls on incorrect trajectories.

\subsection{Additional Runtime and Cost Analysis of \RLM{}s}
\begin{figure*}[htb!]
    \centering
    \includegraphics[width=\textwidth]{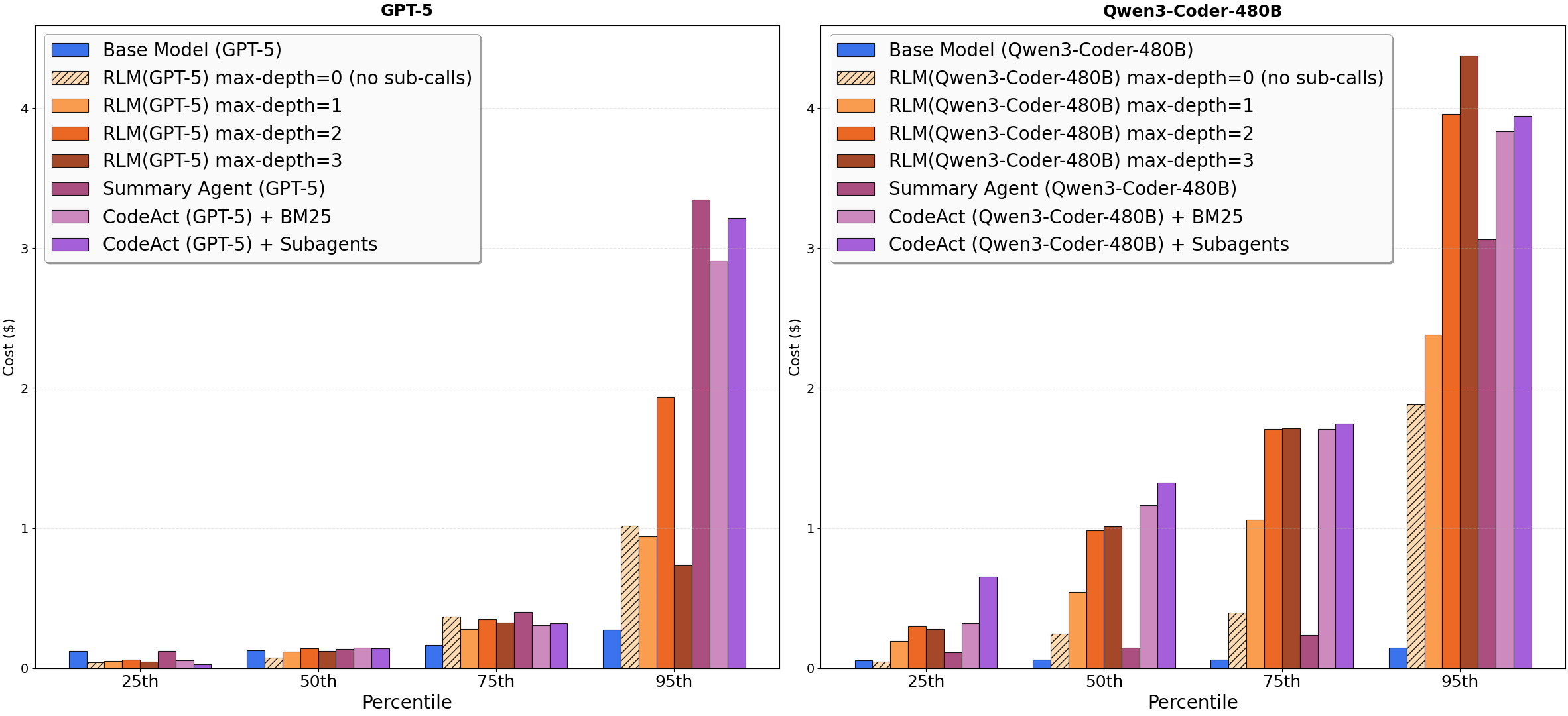}
    \caption{Cost of \RLM{} and baselines described in \S\ref{sec4.2-methods} plotted at the 25th, 50th, 75th, and 95th percentile of total API cost. We observe comparable or even lower costs for \RLM{}s at the 50th percentile, but sharp increases at the tail end due to potentially long \RLM{} trajectories.}
    \label{fig:quartiles}
\end{figure*}

\begin{figure}%
    \centering
    \includegraphics[width=1\textwidth]{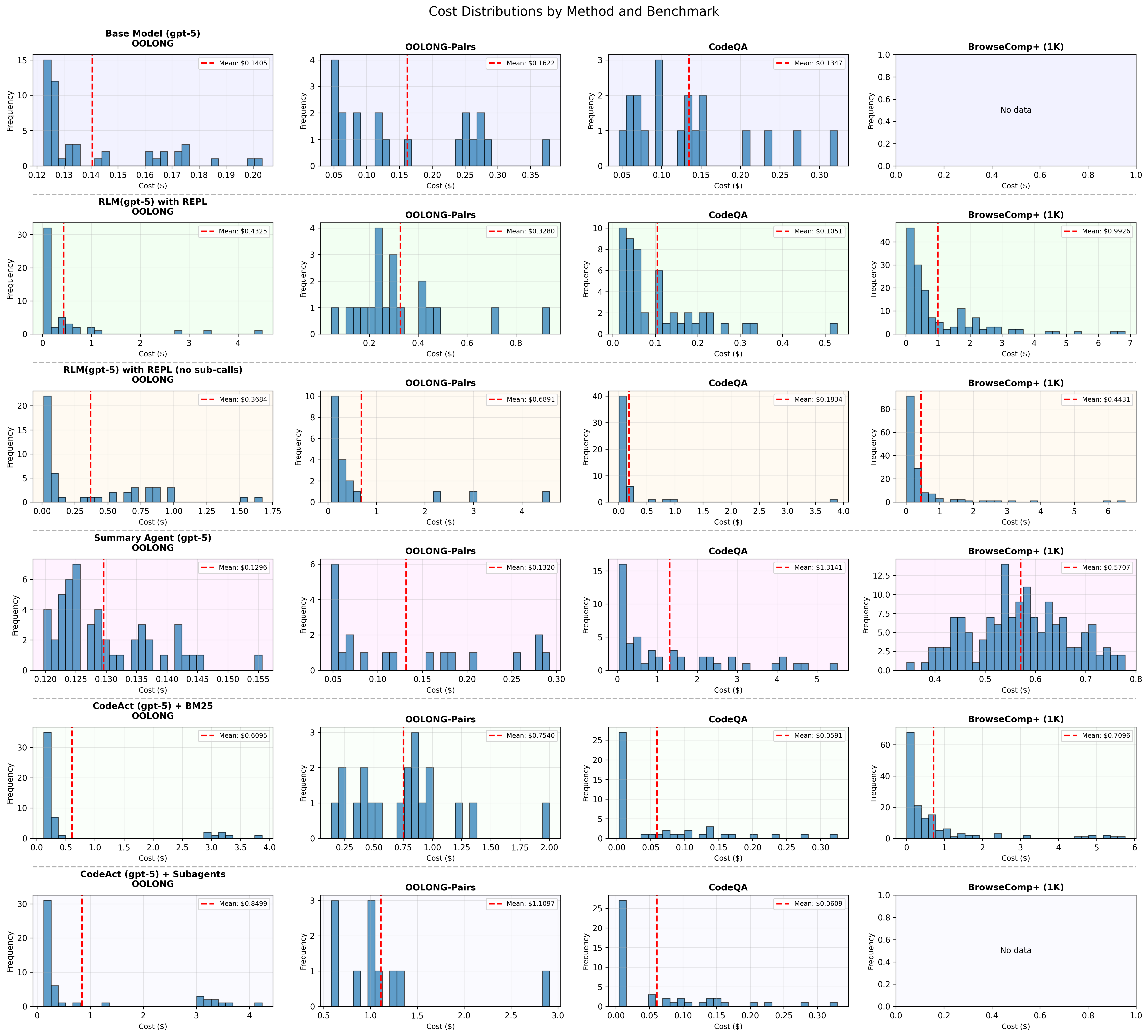}
    \caption{Histogram of the API costs for GPT-5 across OOLONG, OOLONG-Pairs, CodeQA, and BrowseComp+ (1K) for all methods described in \S\ref{sec4.2-methods}.}
    \label{fig:cost-gpt-5}
\end{figure}

\begin{figure}%
    \centering
    \includegraphics[width=1\textwidth]{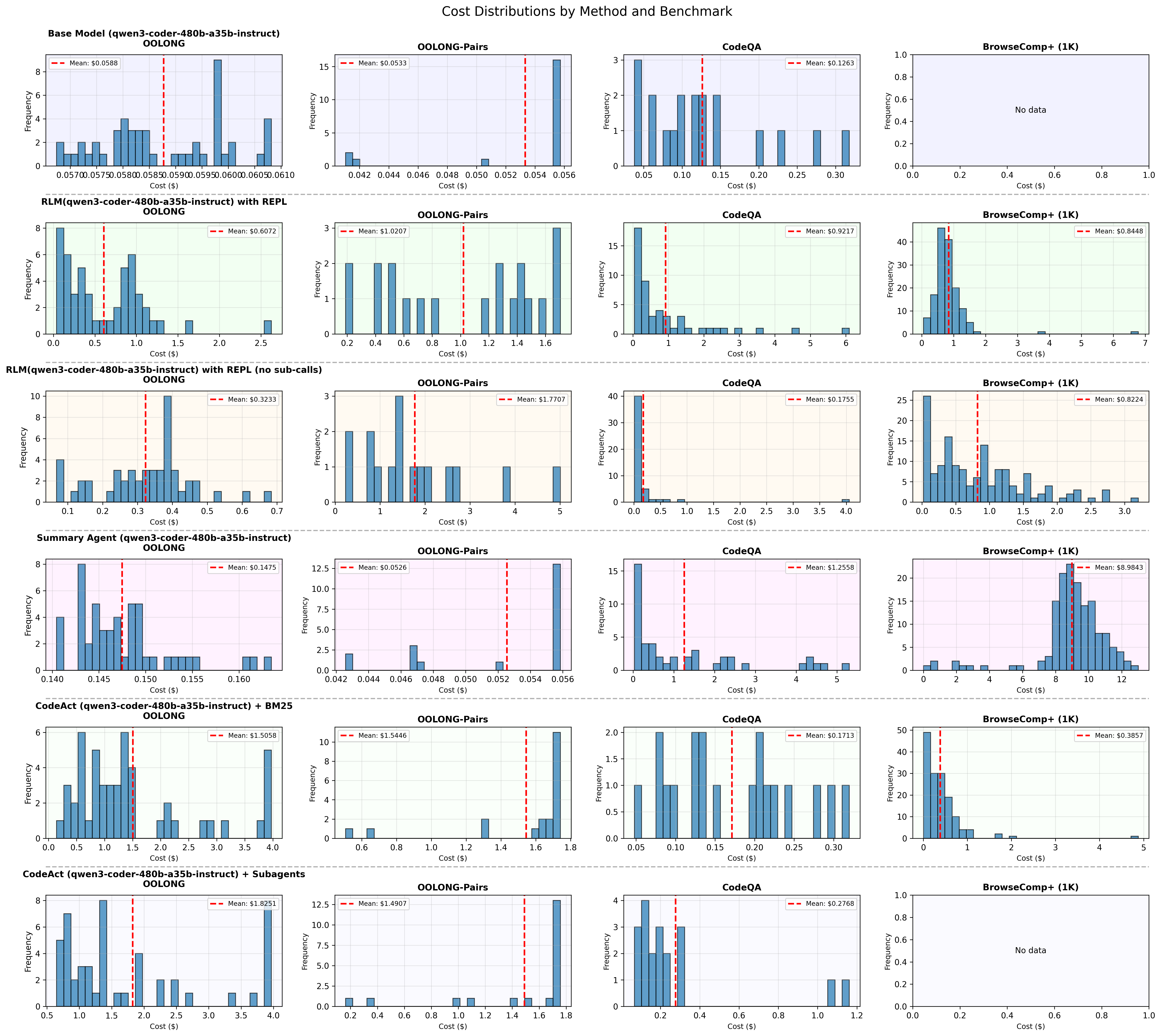}
    \caption{Histogram of the API costs for Qwen3-Coder-480B across OOLONG, OOLONG-Pairs, CodeQA, and BrowseComp+ (1K) for all methods described in \S\ref{sec4.2-methods}.}
    \label{fig:cost-qwen3}
\end{figure}

We supplement the cost and runtime analysis of \RLM{}s with additional, fine-grained plots. We focus on RLMs with depth=0 (i.e. no sub-calls) and depth=1. In Figures~\ref{fig:cost-gpt-5}, \ref{fig:cost-qwen3} we include a histogram for the cost of each method on every task for both GPT-5 and Qwen3-Coder. We generally observe long-tailed, high-variance trajectories for \RLM{}s in both models. We plot the cost of \RLM(depth=1) and baselines at quartiles in Figure~\ref{fig:quartiles}.

\begin{figure}[htb!]
    \centering
    \includegraphics[width=0.9\textwidth]{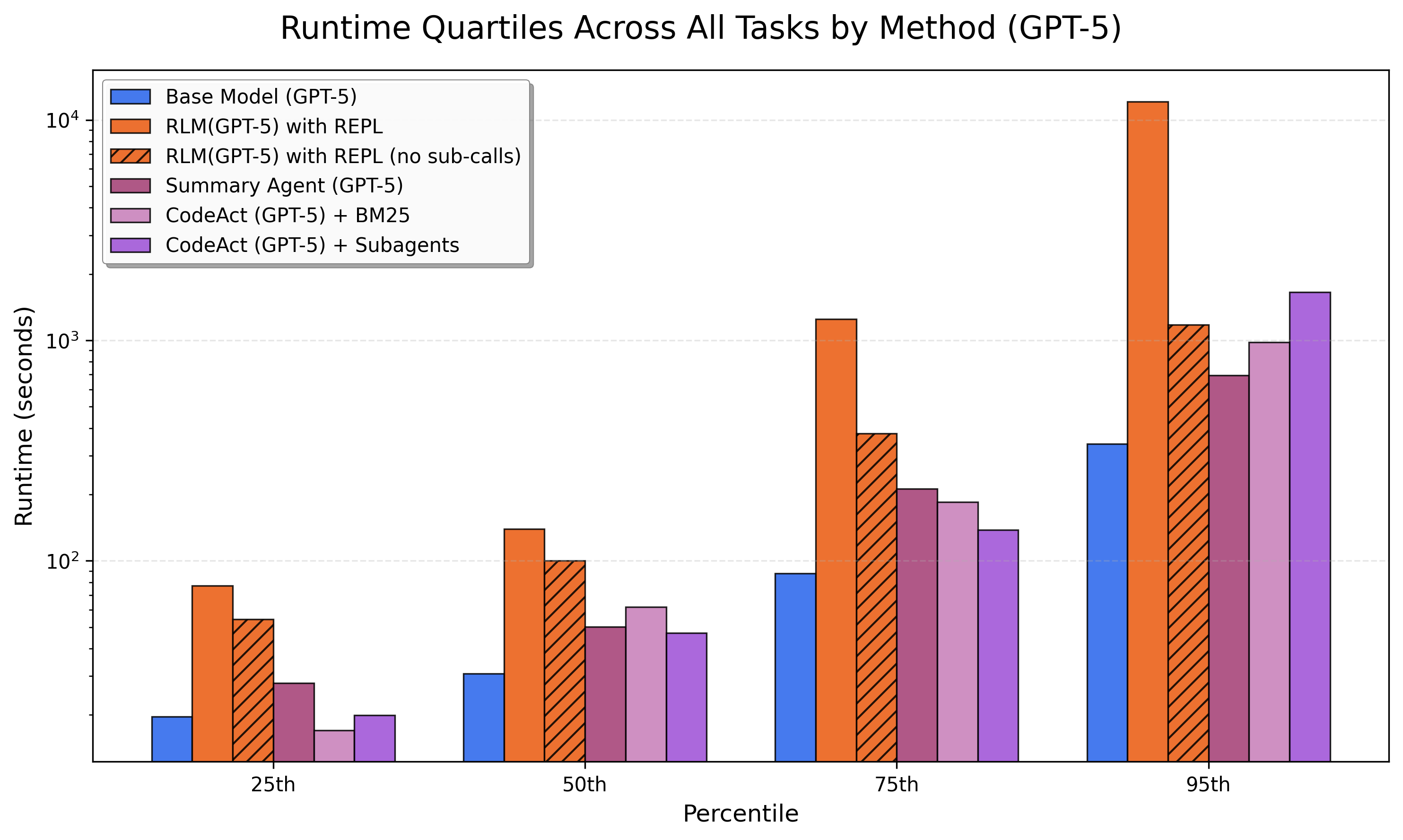}
    \caption{Plotted quartiles of the runtime for methods and baselines around GPT-5 across OOLONG, OOLONG-Pairs, CodeQA, and BrowseComp+ (1K) for all methods described in \S\ref{sec4.2-methods}. We plot the 25th, 50th, 75th, and 95th percentiles.}
    \label{fig:runtime-gpt-5}
\end{figure}

\begin{figure}%
    \centering
    \includegraphics[width=0.9\textwidth]{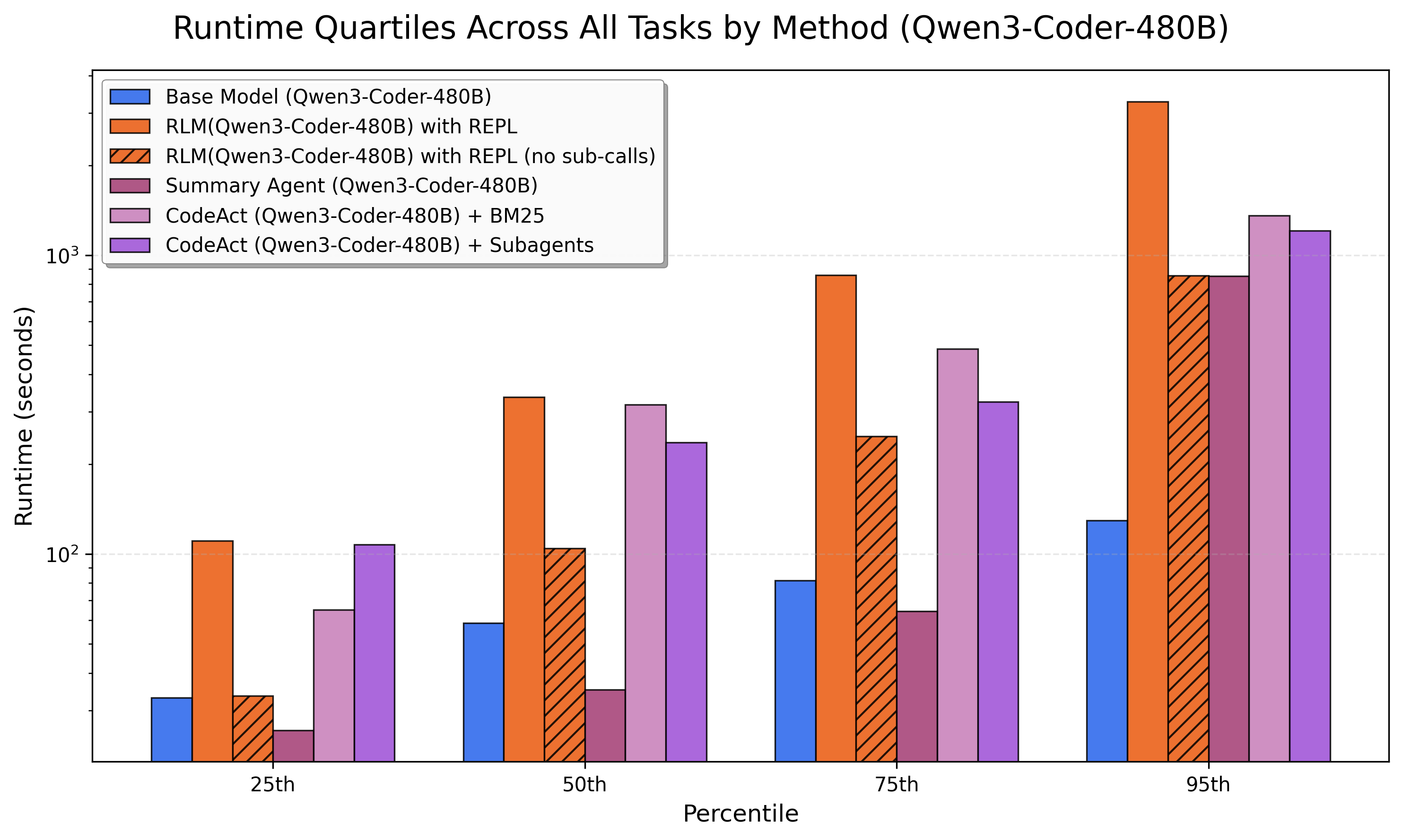}
    \caption{Plotted quartiles of the runtime for methods and baselines around Qwen3-Coder-480B-A35B-Instruct across OOLONG, OOLONG-Pairs, CodeQA, and BrowseComp+ (1K) for all methods described in \S\ref{sec4.2-methods}. We plot the 25th, 50th, 75th, and 95th percentiles.}
    \label{fig:runtime-qwen3}
\end{figure}

We additionally include log-scaled runtime plots (Figure~\ref{fig:runtime-gpt-5},~\ref{fig:runtime-qwen3}) for each method below. The tail end (e.g. 95th percentile) shows extremely long runtimes, which is mainly due to sequential sub-LLM calls taking up most of the runtime. However, we observe these cases happen infrequently, and can be early-stopped with timeout logic. As we remarked in \S\ref{sec4.4-qualitative}, the runtime for these methods can be significantly improved through asynchrony of LM calls and additional prompting to discourage long sub-LM calls or code.

For the scaling plot in Figure~\ref{fig:rlm-scaling}, we also provide the average API cost per task.

\begin{figure}%
    \centering
    \includegraphics[width=\textwidth]{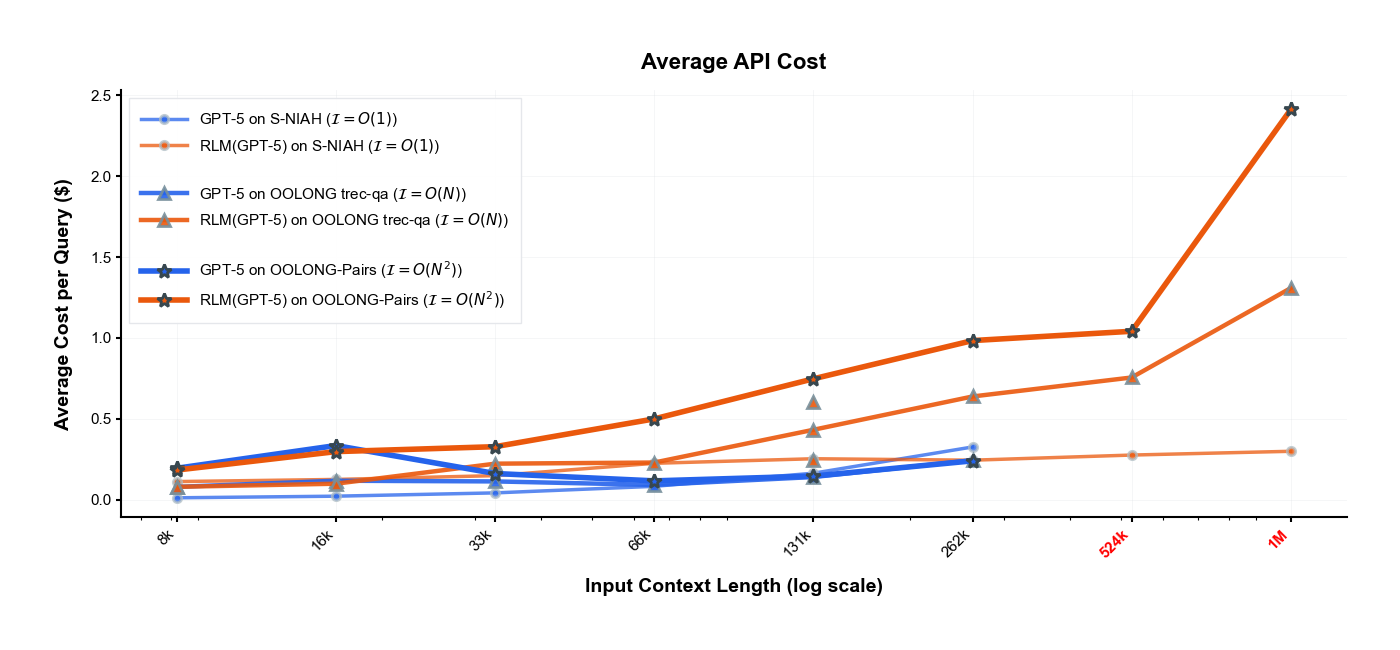}
    \caption{We plot the API cost in USD for the runs in Figure~\ref{fig:rlm-scaling}.}
    \label{fig:cost-scaling}
\end{figure}

\end{document}